\documentclass[12pt,singlespace]{mitthesis}

\usepackage{lgrind}
\usepackage{cmap}
\usepackage[T1]{fontenc}
\usepackage{hyperref}
\hypersetup{ colorlinks, citecolor=blue, linkcolor=blue!80!black!80,urlcolor=blue}
\usepackage{graphicx}

\usepackage{multibib}
\usepackage{xcolor}
\newcites{pers}{Personal References}
\usepackage{graphicx}
\usepackage{subfig}
\usepackage{booktabs} 

\usepackage{amssymb,amsmath,amsfonts,amscd,amstext,bbm, bm, enumerate, dsfont, mathtools}
\usepackage{enumitem}
\usepackage{microtype}
\usepackage{epstopdf}
\usepackage{wrapfig}
\usepackage{verbatim}
\usepackage{etoc}
\usepackage{adjustbox}
\usepackage{color,colortbl}
\usepackage{multibib}

\usepackage{algorithm,algpseudocode}
\usepackage{verbatim}
\usepackage{tcolorbox}
\usepackage{import}
\usepackage{tikz,pgfplots}
\pgfplotsset{compat=1.17}
\usetikzlibrary{tikzmark,shapes}
\usetikzlibrary{matrix}
\usepgfplotslibrary{groupplots}
\usepackage{tabularx}
\usepackage{array}
\usepackage{csquotes}
\usepgfplotslibrary{statistics}
\usepackage{multicol}
\usepackage{multirow}
\usepackage{url}
\usepackage{svg}

\usepackage{setspace}
\usepackage{xcolor}
\usepackage{microtype}
\usepackage{cmap}
\usepackage[final]{pdfpages}
\usepackage{etoolbox}
\usepackage{breakcites}
\usepackage{nomencl}
\makenomenclature

\usepackage{fancyhdr}
\AtBeginEnvironment{subappendices}{%
\counterwithin{figure}{section}
\counterwithin{table}{section}
}

\begin{document}

\begin{titlepage}
   \begin{center}
       
       \includegraphics[width=0.4\textwidth]{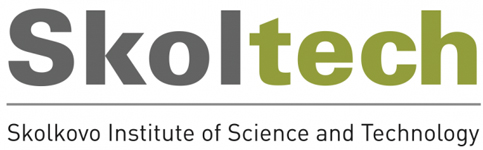}\\
       
       \textbf{Skolkovo Institute of Science and Technology}
       
       \vspace*{2cm}
       
       \textbf{Data-Driven Stochastic AC-OPF using Gaussian Processes}
 
       \vspace{0.5cm}
 
       \vspace{1.5cm}
 
       \textbf{\textit{Doctoral Thesis}}
       
       \textbf{by}
       
       \textbf{Mile Mitrovic}
       
       \vspace{1.5cm}
       
      \textbf{DOCTORAL PROGRAM IN ENGINEERING SYSTEMS}

       \vspace{1.5cm}
       
       \textbf{Supervisor:}
       
       \textbf{Assistant Professor Elena Gryazina}

       \vspace{1.5cm}
       
       \textbf{Co-advisor:}
       
       \textbf{Assistant Professor Petr Vorobev}

       \vspace{4.5cm}
       
       \textbf{Moscow - 2023}
       
       \vspace{1cm}
       \textbf{\copyright \;Mile Mitrovic, 2023}
       
       \vfill
   \end{center}
\end{titlepage}
\begin{minipage}{0.7\linewidth}
\textit{I hereby declare that the work presented in this thesis was carried out by myself at Skolkovo Institute of Science and Technology (Moscow) and has not been submitted for any other degree.}
\end{minipage}\\

\textit{Candidate: Mile Mitrovic}

\textit{Supervisors: Assistant Professor Elena Gryazina}

 \begingroup
 \let\cleardoublepage\clearpage
 \chapter*{Abstract}
 The thesis focuses on developing a data-driven algorithm, based on machine learning, to solve the stochastic alternating current (AC) chance-constrained (CC) Optimal Power Flow (OPF) problem. Although the AC CC-OPF problem has been successful in academic circles, it is highly nonlinear and computationally demanding, which limits its practical impact. The proposed approach aims to address this limitation and demonstrate its empirical efficiency through applications to multiple IEEE test cases.

To solve the non-convex and computationally challenging CC AC-OPF problem, the proposed approach relies on a machine learning Gaussian process regression (GPR) model. The full Gaussian process (GP) approach is capable of learning a simple yet non-convex data-driven approximation to the AC power flow equations that can incorporate uncertain inputs. The proposed approach uses various approximations for GP-uncertainty propagation. The full GP CC-OPF approach exhibits highly competitive and promising results, outperforming the state-of-the-art sample-based chance constraint approaches.

To further improve the robustness and complexity/accuracy trade-off of the full GP CC-OPF, a fast data-driven setup is proposed. This setup relies on the sparse and hybrid Gaussian processes (GP) framework to model the power flow equations with input uncertainty.

 \chapter*{Personal References}
 [1] M. Mitrovic, A. Lukashevich, P. Vorobev, V. Terzija, Y. Maximov, and D. Deka. Fast data-driven chance constrained ac-opf using hybrid sparse gaussian processes. In 2023 IEEE Belgrade PowerTech, pages 1–7. IEEE, 2023. \newline\newline
[2] M. Mitrovic, A. Lukashevich, P. Vorobev, V. Terzija, S. Budennyy, Y. Maximov, and D. Deka. Data-driven stochastic ac-opf using gaussian process regression. International Journal of Electrical Power \& Energy Systems, 152:109249, 2023. \newline\newline
[3] M. Mitrovic, O. Kundacina, A. Lukashevich, S. Budennyy, P. Vorobev, V. Terzija, Y. Maximov, and D. Deka. Gp cc-opf: Gaussian process based optimization tool for chance-constrained optimal power flow. Software Impacts, 16:100489, 2023. \newline\newline
[4] S. Asefi, M. Mitrovic, D. Cetenovic, V. Levi, E. Gryazina, V. Terzija. Anomaly detection and classification in power system state estimation: Combining model-based and data-driven methods. \textit{Sustainable Energy, Grids and Networks}, 2023.
\newline\newline
[5] S. Asefi, M. Mitrovic, D. Cetenovic, V. Levi, E. Gryazina, V. Terzija. Anomaly Detection, Classification and Identification Tool (ADCIT). \textit{Software Impact}, 2023. \newline\newline
[6] M. Mitrovic, D. Titov, K. Volkhov, I. Lukicheva, A. Kudryavzev, P. Vorobev, V. Terzija.
ML-based Method for Diagnostics of Overhead
Transmission Line Insulator State by Leakage
Current. \textit{Engineering Applications of Artificial Intelligence}, 2023 - \textbf{Submitted}. \newline\newline
[7] R. Vorobyev, M. Mitrovic, I. Kremnev. Computer Vision \& Machine Learning Techniques for Non-Destructive Testing of Composites. ECCM 2022 - Proceedings of the 20th European Conference on Composite Materials: Composites Meet Sustainability
4, pp. 520-525.
\newline\newline
[8] O. Kundacina, G. Gojic, D. Miskovic, M. Mitrovic, and D. Vukobratovic. Supporting future electrical utilities: Using deep learning methods in EMS and DMS algorithms. 2023, 22nd International Symposium INFOTEH-JAHORINA (INFOTEH). IEEE, 2023, pp. 1–6.

 \chapter*{Acknowledgments}
 I would like to convey my heartfelt appreciation to the members of my thesis jury, supervisors, colleagues, as well as my friends and family, for their invaluable support, for sharing ideas, and for unwavering encouragement throughout my PhD journey.

I would like to express my deepest gratitude to the members of my thesis jury for their invaluable time and dedication in reviewing and enhancing this dissertation: Henni Ouerdane, Andrei Osiptsov, Federico Martin Ibanez, Haoran Zhao, Ashok Kumar Pradhan and Alexander Nazin. Their commitment and insightful feedback have greatly contributed to the improvement of this work and have provided me with valuable ideas for future research.

I am especially thankful to my advisors, Elena Gryazina and Petr Vorobev, for their guidance and support throughout my PhD studies. Their expertise and mentorship have been instrumental in shaping the direction of my research. 

Special thanks to Deepjyoty Deka and Yury Maximov for introducing me to the thesis topic and for their continuous encouragement to explore new fields and develop new skills.

I would like to express my profound appreciation to Prof. Vladimir Terzija for his support of my ideas and research directions. Our engaging discussions on research, science, and life have been both enlightening and inspiring. Additionally, I am grateful to Prof. Terzija for enabling me to connect with scholars from various universities worldwide.

I cannot thank Dmitry Titov and Dragan Cetenovic enough for their collaboration on a separate project during my PhD. Their assistance in writing papers and their unwavering technical support throughout my studies have been invaluable.

I would like to extend my heartfelt appreciation to my friends and colleagues from Skoltech and around the world who have been an integral part of my PhD journey. Their unwavering support has been instrumental in making this academic endeavor a fulfilling and memorable experience. I am immensely grateful to Strahinja and Kruna Markovic, Zeljko and Anja Tekic, Julijana Cvjetinovic, Dejan Dzunja, Ognjen Kundacina, Aleksandr Lukashevich, Sajjad Asefi, Nikolay Ivanov, Oleg Khamisov, Irina Lukicheva, Aleksandra Burashnikova, Sahar Moghimian, Anastasia Cumik, Artem Zabolotnyi, Roman Yusupov, Akshay Vishwanathan, Vito Michele Leli and many others who have provided their invaluable support and encouragement throughout my journey. Although I may not have listed everyone individually, I am deeply thankful to each and every person who has played a role in my PhD journey. Their presence, friendship, and shared experiences have created a nurturing and inspiring environment, allowing me to overcome challenges and grow both personally and professionally. I am truly grateful for their friendship and the memories we have created together.

My sincere thanks go to my parents, Cvjetko and Radmila, and my sisters Tanja and Nina for their love, support, and understanding. You inspired me to pursue happiness and become a better person.

 \pagestyle{plain}
\tableofcontents

\nomenclature[A]{}{\textbf{Abbreviations}} 
\nomenclature[A]{TSOs}{Transmission System Operators}
\nomenclature[A]{PF}{Power Flow}
\nomenclature[A]{OPF}{Optimal Power Flow}
\nomenclature[A]{RES}{Renewable energy sources}
\nomenclature[A]{RO}{Robust optimization}
\nomenclature[A]{P-OPF}{Probabilistic
Optimal Power Flow}
\nomenclature[A]{CC}{Chance-constrained}
\nomenclature[A]{AC}{Alternating Current}
\nomenclature[A]{DC}{Direct Current}
\nomenclature[A]{GP}{Gaussian Process}
\nomenclature[A]{GPR}{Gaussian Process Regression}
\nomenclature[A]{TA}{Taylor
Approximation}
\nomenclature[A]{EM}{Exact Moment Matching}
\nomenclature[A]{DS}{Distribution System}
\nomenclature[A]{DSOs}{Distribution System Operators}
\nomenclature[A]{KKT}{Karush-Kuhn-Tucker conditions}
\nomenclature[A]{SDP}{Semidefinite programming problem}
\nomenclature[A]{QP}{Quadratic programming}
\nomenclature[A]{AGC}{Automatic generation control}
\nomenclature[A]{SOCP}{Second-order cone programming}
\nomenclature[A]{ERM}{Empirical Risk Minimization}
\nomenclature[A]{SRM}{Structural Risk Minimization}
\nomenclature[A]{MAE}{Mean Absolute Error}
\nomenclature[A]{MSE}{Mean Squared Error}
\nomenclature[A]{RMSE}{Root Mean Squared Error}
\nomenclature[A]{MSLE}{Mean Squared Logarithmic Error}
\nomenclature[A]{ERM}{Empirical Risk Minimization}
\nomenclature[A]{SRM}{Structural Risk Minimization}
\nomenclature[A]{SE}{Squared Exponential}
\nomenclature[A]{SEard}{Squared Exponential Automatic Relevance Determination}
\nomenclature[A]{SLSQP}{Sequential Least
Squares Programming}
\nomenclature[A]{MLE}{Maximum Likelihood Estimate}

\nomenclature[B]{}{\textbf{Notation}}
\nomenclature[B]{$\bar{i}$}{Complex current}
\nomenclature[B]{$\bar{v}$}{Complex voltage}
\nomenclature[B]{$v$}{Voltage magnitude}
\nomenclature[B]{$\theta$}{Voltage angle}
\nomenclature[B]{$\mathcal{G}$}{Set of conventional generator buses}
\nomenclature[B]{$\mathcal{L}$}{Set of load and RES buses}
\nomenclature[B]{$\mathcal{E}$}{Set of transmission lines}
\nomenclature[B]{$\mathcal{B}$}{Set of all buses in power grid}
\nomenclature[B]{$m$}{Overall number of buses}
\nomenclature[B]{$n$}{Overall number of transmission lines}
\nomenclature[B]{$y$}{Line admittance}
\nomenclature[B]{$z$}{Line impendance}
\nomenclature[B]{$p$}{Active power}
\nomenclature[B]{$q$}{Reactive power}
\nomenclature[B]{$s$}{Apparent power}
\nomenclature[B]{$Y$}{Admittance matrix}
\nomenclature[B]{$B$}{Susceptance matrix}
\nomenclature[B]{$G$}{Conductance matrix}
\nomenclature[B]{$R$}{Resistance matrix}
\nomenclature[B]{$X$}{Reactance matrix}
\nomenclature[B]{$\Omega$}{Total power mismatch}
\nomenclature[B]{$\omega$}{Uncertanty realizations}
\nomenclature[B]{$\alpha$}{Participation factor}
\nomenclature[B]{$\epsilon$}{Violation probability}
\nomenclature[B]{$\gamma$}{Power ratio}
\nomenclature[B]{$L$}{Lagrangian system equations}
\nomenclature[B]{$L$}{Lagrangian system equations}
\nomenclature[B]{$\bigtriangledown_{xx}L$}{Hessian of Lagrangian system equations}
\nomenclature[B]{$J_{KKT}$}{Jacobian matrix of the KKT system}
\nomenclature[B]{$s_{sl}$}{Slack variables}
\nomenclature[B]{$x$}{Decision variables}
\nomenclature[B]{$c$}{Scalar cost coefficients}
\nomenclature[B]{$tr(\cdot)$}{Trace of a matrix}
\nomenclature[B]{$rank(\cdot)$}{Rank of a matrix}
\nomenclature[B]{$\mathbb{E}$}{Expectations}
\nomenclature[B]{$\mathbb{P}$}{Probabilities}
\nomenclature[B]{$\mathcal{D}$}{Population distribution}
\nomenclature[B]{$\mathcal{S}$}{Sample distribution}
\nomenclature[B]{$x^{in}$}{Input sample vector of dataset}
\nomenclature[B]{$y^{out}$}{Output sample vector of dataset - label}
\nomenclature[B]{$\ell_c$}{Loss function}
\nomenclature[B]{$\mathcal{F}$}{Set of learning functions}
\nomenclature[B]{$m_s$}{Number of samples in training set $\mathcal{S}$}
\nomenclature[B]{$\mathcal{R}$}{Set of real numbers $\mathcal{S}$}
\nomenclature[B]{$\mathbb{D}$}{Vector dimension $\mathcal{S}$}
\nomenclature[B]{$\mathcal{N}(\cdot)$}{Normal distribution}
\nomenclature[B]{$I$}{Identity matrix}

\nomenclature[B]{$\frac{\partial \cdot (x)}{\partial x}$}{ the partial derivative of function $\cdot$ evaluated at~$x$}

\printnomenclature
 \listoftables
 \listoffigures

 \clearpage
 \chapter{General Introduction}\label{ch:intro}
\chaptermark{Introduction}

The power grid is widely regarded as one of the most remarkable engineering accomplishments of the 20th century. It has played a key role in enabling economic prosperity and advancing social progress for billions of people worldwide. However, the task of managing and controlling the power grid is becoming increasingly complex for Transmission System Operators (TSOs). Despite the stabilization of the average annual total demand, new patterns of energy consumption and generation have emerged, posing new challenges for TSOs~\cite{liu2012challenges}. 

TSOs rely on Optimal Power Flow (OPF) as a fundamental tool to ensure secure and cost-effective power system operation, commonly used in electricity markets~\cite{ng2018statistical} and system security assessments~\cite{capitanescu2011state}. OPF is a mathematical optimization problem that determines the optimal settings for power generators, transformers, and other devices in the power system~\cite{cain2012history}. In this context, OPF can be seen as an economic dispatch problem, with the goal of determining how the grid should set generator outputs in real-time. This involves dispatching generators at regular intervals, usually every fifteen minutes to an hour (depending on the power grid), to balance demand and generator output at the lowest possible cost while respecting the operational limitations of the generators and transmission lines. By using OPF to optimize the generator output, TSOs can ensure a reliable and efficient power supply while minimizing costs.

In recent years, we have witnessed that conventional power grids have been undergoing a transformation into modern smart grids. This transition has resulted in the integration of various new facilities, including renewable energy sources (RES), electric cars (EC), and Internet of Things (IoT) devices. The use of RES like wind and solar plants introduces large uncertainty into the power grids, which can have unexpected consequences. Neglecting or underestimating the impact of these uncertainties can result in conservative operations, driving up operating costs. Moreover, aggressive operations will probably lead to constraint violations and thus jeopardize the security of the grid. For example, renewable outputs have the potential to produce power flows (PFs) that significantly exceed the ratings of power lines. Exceeded line ratings can cause grid instability and cascading failures that may ultimately result in a blackout. Moreover, there is a regulation that RES should cover $20\%$ of all generations in the US and Europe by $2030$~\cite{cigre1, eere1, dena1, gonzalez2006experience}. As a result, TSOs must strike a balance between operating costs and security. This fact leads to solving the OPF problem in a stochastic context. 

The thesis aims to tackle the challenges outlined in the problem background above by formulating a more faster and robust solution method for addressing stochastic OPF problems. The primary research objectives of this thesis include:
\begin{enumerate}
\item developing an approach that strikes a trade-off between computational complexity and optimal solutions in solving stochastic OPF problems; 
\item enhancing the overall efficiency of the solutions derived from the stochastic OPF approach. 
\end{enumerate}
Accordingly, the main research questions of this thesis are:
\begin{enumerate}
\item how can a trade-off between computational complexity and optimal solutions be achieved in the stochastic OPF problem?
\item how proposed approach can improve the efficiency of the stochastic OPF solution?
\end{enumerate}

Stochastic OPF involves solving the OPF problem while taking uncertainties into account. Various approaches have been proposed in the literature to address this issue. Among them, the most popular are robust optimization (RO), probabilistic OPF (P-OPF), and chance-constrained (CC) approach. While RO ensures secure operations against all possible uncertainty realizations within a given set, it often leads to conservative solutions~\cite{ben2009robust, warrington2013policy}. P-OPF, on the other hand, is challenging to put into practice as it results in probabilistic distributions of control variables that can not lead to deterministic scheduling strategies~\cite{ullah2022advanced, schellenberg2005cumulant, zhang2010probabilistic}. The chance-constrained approach, however, ensures that chance constraints are satisfied within an acceptable violation probability~\cite{du2021chance, xiao2001hybrid, zhang2011chance, sjodin2012risk, vrakopoulou2013probabilistic, roald2015security, vrakopoulou2013probabilistic_, bienstock2014chance, roald2015security, morillo2022distribution, wu2019chance}. In general, the chance-constrained approach is a mathematical optimization technique used in decision-making under uncertainty. It means that this approach helps in making decisions that account for uncertain parameters or variables while maintaining control over the allowable level of risk or the probability of constraint violations. These uncertainties often stem from external factors like market fluctuations, varying weather conditions, or unpredictable changes in demand. Since the chance-constrained problem is typically applied to constrained optimization problems, decision-making is specified with the risk tolerance or acceptable level of constraint violation. Thus, the chance-constrained (CC) OPF approach enables TSOs to balance security and operating costs in an intuitive and transparent manner. Therefore, this work focuses on the CC-OPF approach.

While chance constraints offer a way to address uncertainty in a quantitative manner, solving the Alternating Current Chance-Constrained Optimal Power Flow (AC CC-OPF) is notoriously challenging \cite{roald2017chance, muhlpfordt2019chance}. Alternating current (AC) is a type of electric current that periodically reverses direction within a circuit. Unlike direct current (DC), which flows steadily in one direction, AC changes direction periodically, typically in a sinusoidal waveform. To overcome the challenge of solving AC CC-OPF, many studies convert the stochastic optimization problem to a deterministic one. However, this reduces the confidence region of AC CC-OPF compared to the feasible region of AC OPF. Additionally, both regions are non-convex and difficult to deal with. To make the problem more manageable, researchers try to properly eliminate the nonlinear aspects and reformulate the stochastic optimization problem as a deterministic one, using convexification techniques such as linear approximation or convex relaxation. Thus, they provide the resulting convex optimization problem numerically tractable.

Although convexification techniques have been used to make the AC CC-OPF problem tractable to solve, the resulting solutions may not be optimal in terms of achieving the lowest possible cost. To address these challenges, we decided to investigate a new approach. Therefore, this study proposes a data-driven approach to replace the AC power flow (AC-PF) balance equations with a probabilistic approximation based on a supervised machine learning (ML) model. The proposed approach employs a Gaussian Process Regression (GPR) as the probabilistic supervised ML model. The GPR model is directly integrated into the CC-OPF formulation to create a new data-driven approach called Gaussian Process Chance-Constrained Optimal Power Flow (GP CC-OPF). This approach offers a novel way to model and solve stochastic OPF problems using Gaussian Processes (GP) and has the potential to provide more accurate solutions with reduced computational effort.

\section{Contributions}

This thesis aims to develop a robust and effective approach for solving stochastic CC-OPF problems that strikes a trade-off between solution accuracy and computational complexity. By achieving this, the proposed approach can enable the optimization of operating costs in the presence of uncertainty while maintaining the security of the power system. Therefore, we propose a novel GP-based data-driven approximation of the AC-PF balance equations, which is integrated into the CC-OPF formulation. Moreover, we consider two cases: 
\begin{enumerate}
\item \textbf{full} GP CC-OPF: where GPR fully replaces AC-PF balance equations;
\item \textbf{hybrid} GP CC-OPF: that combines linear direct current (DC) PF balance equations with the data-driven estimation of the residuals between DC-PF and AC-PF based on GPR. 
\end{enumerate}

We account for both fluctuating loads and RES as input uncertainty variables. To additionally propagate input uncertainty to output variables we consider and compared different approximation techniques such as first and second-order Taylor Approximation (TA) and Exact Moment Matching (EM). 

To ensure the scalability of the proposed GP-based approach, we use sparse GPR, which employs a few selective data samples for estimation. The practical efficiency of the proposed approach is validated and illustrated using a number of standard IEEE test cases. Additionally, we compare the proposed data-driven GP CC-OPF approach with state-of-the-art sample-based CC-OPF approaches. Results show that the proposed GP-based reformulation of the CC-OPF is competitive and outperforms conventional sample-based formulations.

One of the key advantages of the GP-based approach is reflected in the fact that the proposed approach does not require knowledge of the grid configuration and parameters. This thesis can have significant importance for TSOs who seek to make informed decisions regarding cost-effective and secure system operation under uncertain conditions.

\section{Thesis Structure}

This thesis is divided into two main parts. The first part focuses on presenting the theory of optimal power flow and the Gaussian processes relevant to our research topic. In this section, we also provide an overview of the current state-of-the-art in the field.

The second part of the thesis will be dedicated to presenting our contributions to the field. Here, we will showcase our unique perspective and original research findings.

\noindent The first part of this thesis consists of two chapters. 
\begin{itemize}
    \item[$\ast$] In \textbf{chapter \ref{ch:2}}, we introduce the fundamental concept of power systems and provide a detailed explanation of the OPF framework. Furthermore, we delve into recent approaches and methods that have been developed to solve OPF problems. Additionally, in this chapter, we describe how a synthetic dataset is generated using a simulated power system model. 
    \item[$\ast$] In \textbf{chapter \ref{ch:3}}, we introduce the fundamental concepts of machine learning focusing on supervised learning and present the theoretical background of the Gaussian process regression model.
\end{itemize}

\noindent The second part of this thesis consists of three chapters. 
\begin{itemize}
    \item[$\ast$] In \textbf{chapter \ref{ch:4}}, we present a novel data-driven CC-OPF approach based on GPR. The GPR model is trained on a synthetic dataset and used to replace the full AC-PF balance equations in the CC-OPF approach. This approach allows for the propagation of input uncertainties to output variables, and we compare different approximation techniques for this. Additionally, we compare the results of the full GP CC-OPF approach with sample-based stochastic approaches.  
    \item[$\ast$] In \textbf{chapter \ref{ch:5}}, we propose a more accurate, scalable and robust approach than in chapter~\ref{ch:4}. Specifically, we utilize a hybrid approach that combines the linear DC-PF and additive GP part. The GP part is learned on residuals between AC-PF and DC-PF, allowing us to better capture the complex relationships between the input and output variables of the power system. To further improve the computational efficiency of our approach, we incorporate a sparse GP. This involves using a small set of points to better approximate the marginal likelihood and reduce the computational complexity, while still preserving important information from the original sample span.
    \item[$\ast$] In \textbf{chapter \ref{ch:6}}, we provide a brief explanation and documentation of the software design for this research.
\end{itemize}

\noindent Finally, in \textbf{chapter \ref{ch:conclusion}} we conclude our work and present directions for future work.

\section{Corresponding papers}
The contributions presented in this manuscript are built upon the following papers that were developed as part of the research conducted during this Ph.D. It is important to acknowledge that these papers were collaborative efforts involving numerous co-authors. However, it should be noted that the personal contribution to these papers encompasses all experimental aspects, with the exception of the scenario-based chance-constrained method for comparison that was performed by Aleksandr Lukashevich. Additionally, there was a partial contribution made to the theoretical parts, guided by the supervision of Elena Gryazina, Petr Vorobev, Yury Maximov and Deepjyoti Deka.

\begin{itemize}
\item[] \textbf{Chapter \ref{ch:4}} is based on the paper \cite{mile1} published at the \textit{International Journal of Electrical Power \&
Energy Systems} (IJEPES 2023).

\item[] \textbf{Chapter \ref{ch:5}} is based on the paper \cite{mile2} published at the \textit{IEEE Belgrade PowerTech 2023 Conference} (PowerTech 2023).

\item[] \textbf{Chapter \ref{ch:6}} is based on the paper \cite{mitrovic2023gp} published at the \textit{Software Impact} journal (SIMPA 2023).
\end{itemize}

 ~\\~\\~\\~\\~\\~\\~\\~\\~\\~\\~\\~\\~\\
\addcontentsline{toc}{chapter}{Part I: Background Theory \& State-of-the-Art}
\begin{center}{\Huge \textsc{\underline{Part I}}\\~\\ \textsc{Background Theory \\ ~ \\ \& \\~ \\  State-of-the-Art}}\end{center}\normalsize
 \chapter{Power System Description and Optimal Power Flow}
\label{ch:2}
\section{Introduction} \label{ch2_intro}

Chapter \ref{ch:2} aims to provide readers with a comprehensive understanding of power systems and optimal power flow (OPF). 

The first part introduces the fundamental concepts of power grids and the equations used to describe their behavior. The goal is to help readers develop intuition about these objects and concepts that are critical for modeling and simulating power grids.

The second part focuses on OPF and its mathematical concepts, highlighting its importance in power systems. Additionally, we discuss state-of-the-art solutions for solving OPF problems, along with some of the challenges involved.

In the third part, we provide readers with a brief introduction to sampling and generating a synthetic dataset used in the contribution part.

In summary, this chapter is structured into five sections. Section~\ref{ch2_ps} introduces the main concepts of power systems, while section~\ref{ch2_opf} presents an overview of OPF and its significance. Section~\ref{ch2_dataset} talks about generating a synthetic dataset from a simulated power system. Finally, section~\ref{ch2_conclusion} provides a summary of the key points covered in the chapter. Overall, this chapter will provide readers with a solid foundation in power systems and OPF, setting the stage for more advanced discussions later in the contribution part (part II).
\section{Power System Description} \label{ch2_ps}

Generally, power systems can be divided into three main components: production (generation, supply), transmission, and distribution (load, consumption). A visual representation of this concept is presented in Fig.~\ref{fig:power_system_1}. 
\begin{figure}[H]
\centering
\vspace*{-10pt}
\includegraphics[width=0.95\textwidth]{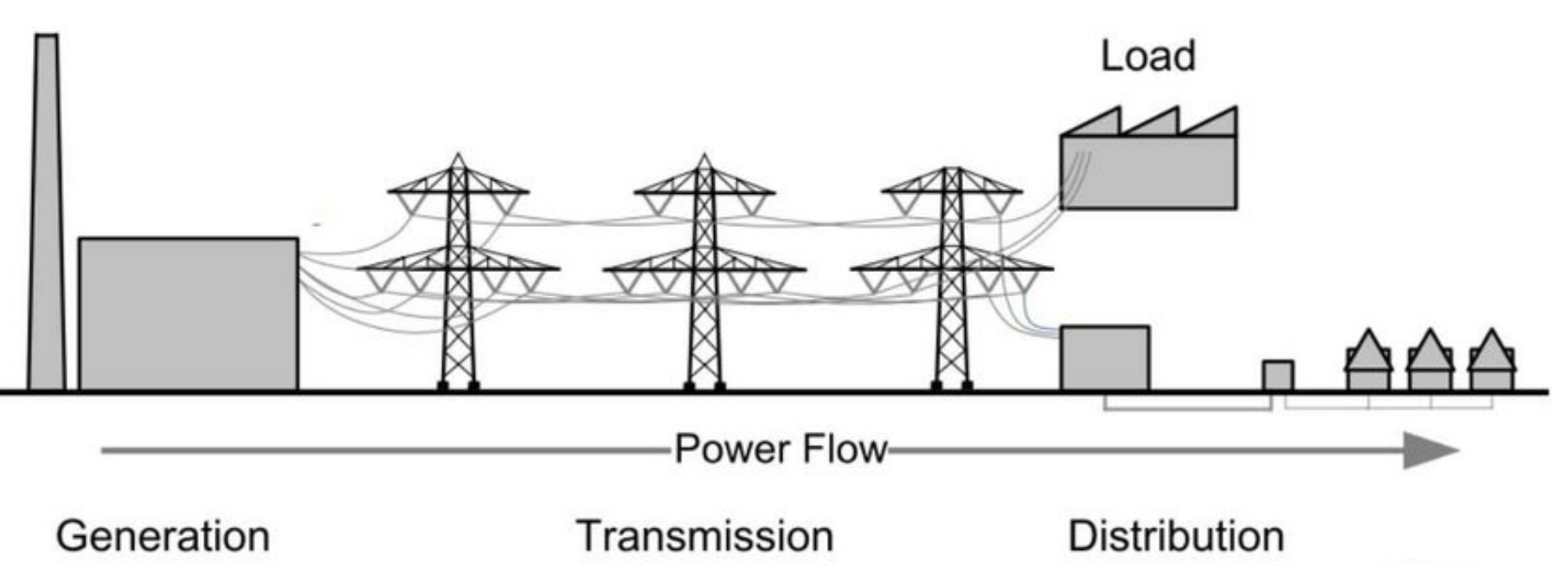}
\caption[Fundamental concept of the power systems; the figure is taken from \href{https://engineeringnotesonline.com/power-system-structure/}{https://engineeringnotesonline.com}]{Fundamental concept of the power systems; the figure is taken from \href{https://engineeringnotesonline.com/power-system-structure/}{https://engineeringnotesonline.com}.}
\label{fig:power_system_1}
\end{figure}

\textbf{Generation} involves facilities that produce electricity and sends it to the power system. There are various sources of electricity production, including thermal power plants (coal, fuel, gas, or nuclear), hydropower plants, and renewable sources (wind or solar). In our discussion, we will consider thermal and hydropower plants as controlled conventional generation, while renewables will be treated as uncontrolled, uncertain generation.

\textbf{Load} refers to all facilities that consume electrical power. It is essential to note that the load is not just a single household, but rather a group of consumers, such as a small town or a large industrial firm. This assumption is made from the perspective of Transmission System Operators (TSOs), as this group of consumers is directly connected to the high-voltage transmission system. Typically, this group of consumers works on a low-voltage called the Distribution System (DS) operated by the Distribution System Operators (DSOs).  

\textbf{Transmission} system contains high-voltage lines that serve as a connection between power generation and consumption. It is operated by TSOs whose responsibility is to ensure that consumers can access the required amount of power at any time and from any location. Additionally, TSOs are charged with maintaining the system's reliability and ensuring that consumers have a secure supply of electricity. This thesis will primarily focus on the high-voltage transmission system, which will be further described in this section.

In most developed countries, the energy market is structured so that there are three distinct entities: producers, TSOs, and DSOs. However, the transmission system is usually a state monopoly. This means that TSOs are responsible for ensuring that producers and consumers with their specific behavior have efficient access to the grid, while also controlling the entire power grid to maintain its reliability and security.

Power grids are often represented as graphs to help explain the terminology and notation used in the field. To provide an example of this, we examine a graph of the IEEE 5-bus high-voltage power system in Fig.~\ref{fig:power_system_2}. 
\begin{figure}[H]
\centering
\vspace*{-10pt}
\includegraphics[width=1.\textwidth]{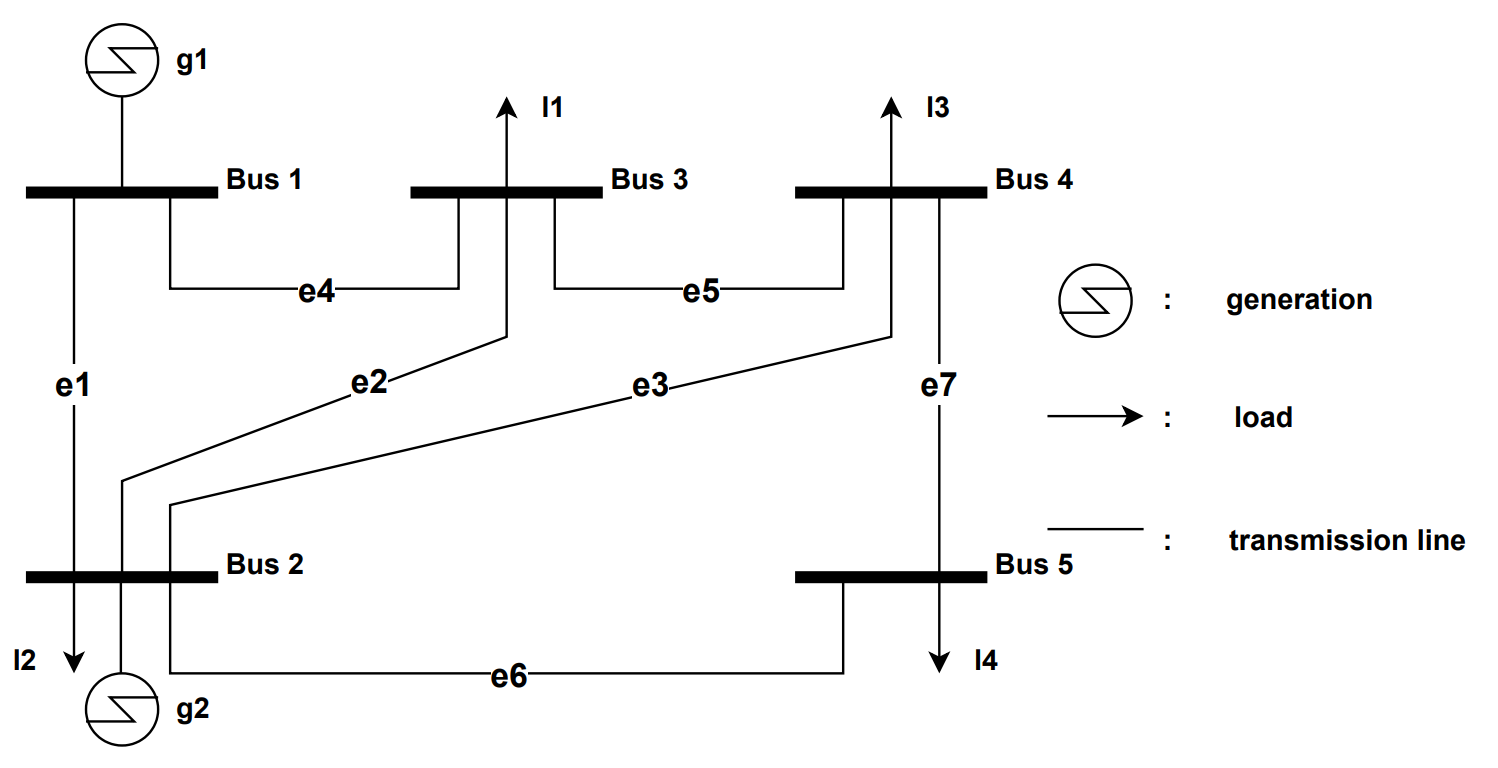}
\caption[Power System Graph]{Graph representation of the power system (IEEE5 bus example).}
\label{fig:power_system_2}
\end{figure}

The power system graph shown in the figure comprises of four essential components: buses, generators, loads, and transmission lines. A bus refers to a physical substation that connects various elements via switching devices, allowing for changes in connectivity between components. Generators are represented by $g$ and form a set $\mathcal{G}$ ($g \in \mathcal{G}$), whereas loads are represented by $l$ and belong to set $\mathcal{L}$ ($l \in \mathcal{L}$). In this work, for brevity, we will use the common word \textbf{injection} which refers to both generation and loads. Transmission lines, denoted by $e$ ($e \in \mathcal{E}$), connect two or more buses and enable power flow (PF) in the power grid. All buses in the power grid, we will denote with set $\mathcal{B}$

While the power grid also includes other elements such as transformers, phase shifter transformers, HVDC, AC/DC converters, and capacitor banks, this work focuses solely on the four primary components mentioned above. Therefore, we will not delve into the description of these additional elements.

\subsection{Power Flow}
In this subsection, we will discuss the power flow analysis used in power systems. This analysis is crucial for determining the steady-state operating conditions of the power grid. It involves calculating the voltage magnitudes, phase angles, and power flows in all components of the power system, including generators, transmission lines, and loads.

During power flow analysis, the electrical characteristics of the power system components, such as their impedances, and the real and reactive power demands of the loads are considered. By solving the first principle equations that describe the power system's behavior, the power flow analysis calculates the steady-state voltages and power flows throughout the grid.

The results obtained from the power flow analysis are essential for power system planning, design, and operation. They enable us to determine if the system is operating within its capacity limits, identify potential bottlenecks, and assess the impact of new loads or generators on the power system. Moreover, the power flow analysis is vital for developing control and protection schemes that ensure the stability and reliability of the power system.

To provide a better understanding of the power flow analysis, we will discuss the first principal equation. Since the majority of the grid operates on alternating current (AC), we will focus on AC power flow analysis. However, we will also briefly explain direct current (DC) power flow analysis for the reader's benefit.

In AC power transmission systems, voltage, and current intensity are the key variables that describe the system~\cite{kundur94power}. These variables are represented by complex numbers consisting of a magnitude and a phase (angle). The objective of power flow analysis is to determine these magnitudes and angles at every bus in the grid using the available data and to utilize them to calculate power line flows. 

This thesis focuses on power flow equations that apply to a fixed voltage frequency and a quasi-stationary operating condition, where power generation and loads are balanced. This condition ensures that the total power generated is equal to the total power consumed, including losses. Although this work does not delve into transient phenomena that occur over shorter periods, it is worth noting that such phenomena can have a significant impact on power grid operations in certain situations.

Based on Fig.~\ref{fig:power_system_2}, we can infer that the grid under consideration comprises $m=5$ buses and $n=7$ lines. Each bus $k \in (1,...m)$  is characterized by voltage magnitude $v_k$ and voltage angle $\theta_k$, while each line $e$ connecting two buses ($j$ and $k$) is described by an impedance ($z_{j,k}$) which represents the physical properties of the line~\cite{kundur94power}. This impedance is a complex number that includes both resistance and reactance. In power flow equations, impedance is often expressed as admittance, where $y_{j,k}=\frac{1}{z_{j,k}}$. A power line with admittance $0$ indicates that there is no connection between the respective buses.

Following~\cite{kundur12power}, the current in lines between busses $k$ and $j$ is denoted with $\bar{i}_{k,j}$, and represents: 
\begin{equation}
 \label{eq:ch2_ps_1}
 \bar{i}_{k,j} = y_{k,j}(\bar{v}_j - \bar{v}_k)
\end{equation}
where $\bar{v}_j$ and $\bar{v}_k$ are voltages (complex number) in buses and $y_{k,j}$ is an admittance of the considered line. The power system is based on  Kirchhoff’s law where the sum of the currents entering and leaving a bus must be equal. Therefore, any current injected at bus $k$ must also leave bus $k$ and can be expressed as:
\begin{equation}
 \label{eq:ch2_ps_2}
 \bar{i}_{k} + \sum_{j=1, j\neq k}^{m} \bar{i}_{k,j} = 0
\end{equation}
The equation~\ref{eq:ch2_ps_2} can easily be derived from Kirchhoff’s bus law presented in Fig.~\ref{fig:power_system_3}.
\begin{figure}[H]
\centering
\vspace*{-10pt}
\includegraphics[width=1.\textwidth]{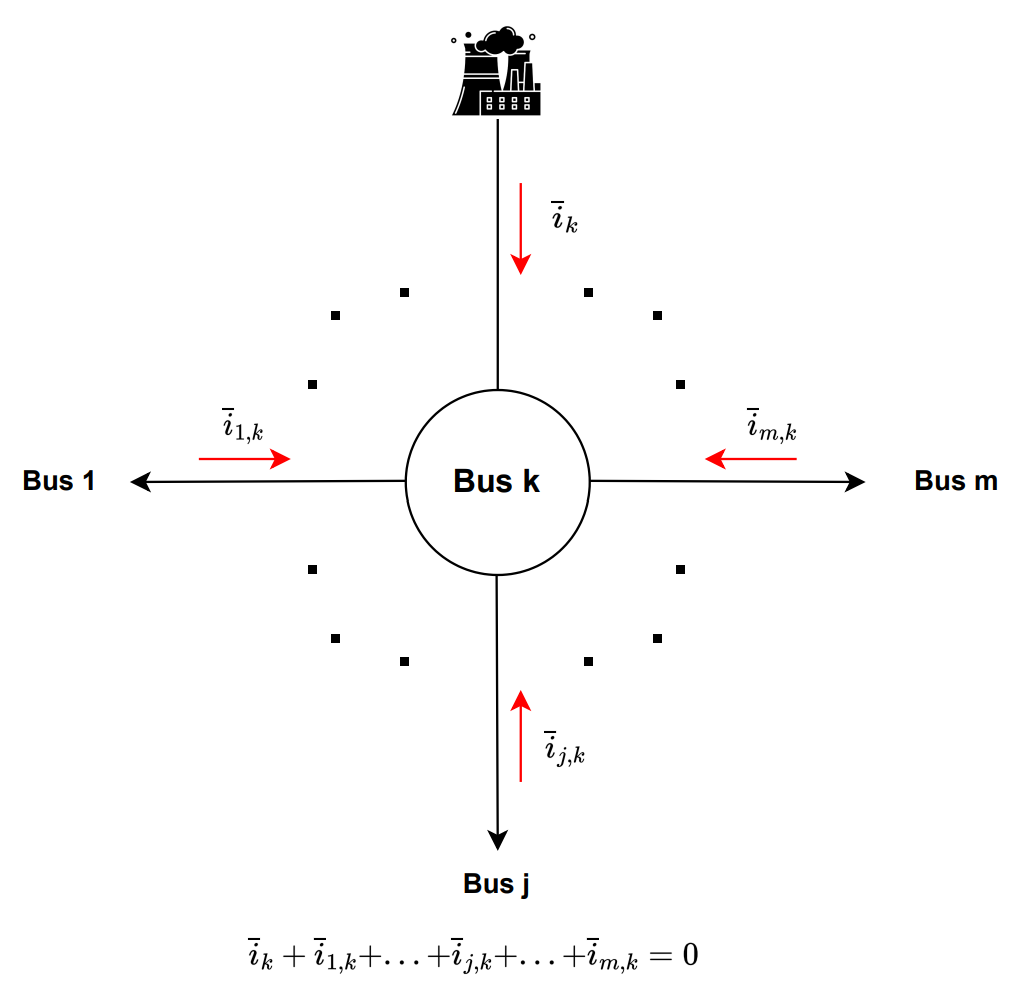}
\caption[Kirchhoff’s law]{Illustration of Kirchoff’s bus law.}
\label{fig:power_system_3}
\end{figure}

Further, combining equations~\ref{eq:ch2_ps_1} and~\ref{eq:ch2_ps_2} we get:
\begin{equation}
 \label{eq:ch2_ps_3}
 \bar{i}_{k} + \sum_{j=1, j\neq k}^{m} \bar{i}_{k,j} = 0
\end{equation}
By vectorizing equation~\ref{eq:ch2_ps_3}, the full power grid can be described in matrix form as follows:
\begin{equation}
 \label{eq:ch2_ps_4}
 \begin{bmatrix} \bar{i}_{1}  \\ \cdot \cdot \\ \bar{i}_{j} \\ \cdot \cdot \\ \bar{i}_{m}\end{bmatrix}
 =
 Y
\begin{bmatrix} \bar{v}_1  \\ \cdot \cdot \\ \bar{v}_j \\ \cdot  \cdot \\ \bar{v}_m \end{bmatrix}
\end{equation}
where $Y$ is the $m$ x $m$ admittance matrix, defined as:
\begin{equation}
 \label{eq:ch2_ps_5}
 Y =
\begin{bmatrix} -\sum_{j\neq 1} y_{1,j} & y_{1,2} & ... & y_{1,j} & ... & y_{1,m} \\ y_{2,1} & -\sum_{j\neq 2} y_{2,j} & ... & y_{2,j} & ... & y_{2,m} \\ ... & ... & ... & ... & ... & ...\\ y_{m,1} & ... & ... & y_{m,j} & ... & -\sum_{j\neq m} y_{m,j} 
\end{bmatrix}
\end{equation}
where the non-diagonal elements are equal as e.g $y_{1,2} = y_{2,1}$.

The injected current can be represented as the injected complex power $\bar{s}_k$:
\begin{equation}
 \label{eq:ch2_ps_6}
 \bar{s}_k = \bar{v}_k \bar{i}_{k}^{*}
\end{equation}
or as a complex number of injected active power ($p_k$) and reactive power ($q_k$):
\begin{equation}
 \label{eq:ch2_ps_7}
 \bar{s}_k = p_k + \mathbf{i} q_k
\end{equation}
where in this context $\mathbf{i}$ is the complex number ($\mathbf{i}^2 = -1$), and $p_k$ and $q_k$ are a real and imaginary part of complex power.

Similarly, the active power ($p_{k,j}$) and reactive power ($q_{k,j}$) in transmission lines can be expressed as power flow as follows:
\begin{equation}
 \label{eq:ch2_ps_8}
 \bar{s}_{k,j} = p_{k,j} + \mathbf{i} q_{k,j}
\end{equation}

By replacing Eq.~\ref{eq:ch2_ps_4} with equations~\ref{eq:ch2_ps_6} and~\ref{eq:ch2_ps_7} we get the AC power flow equations as:
\begin{subequations}\label{eq:ch2_ps_9}
\begin{align}
 p_{k,j} = v_k v_j \left(G_{k,j}\cos(\theta_{k}- \theta_{j}) + B_{k,j}\sin(\theta_{k} - \theta_{j})\right) \label{eq:ch2_ps_9.1} \\
 q_{k,j} = v_k v_j \left(G_{k,j}\sin(\theta_{k}- \theta_{j}) - B_{k,j}\cos(\theta_{k} - \theta_{j})\right) \label{eq:ch2_ps_9.2}
\end{align}
\end{subequations}
where Eq.~\ref{eq:ch2_ps_9.1} describes the relationship between active power flow and voltage magnitude and angle ($v$ and $\theta$), while Eq.~\ref{eq:ch2_ps_9.2} expresses reactive power flow. The admittance $y_{k,j}$ between buses $k$ and $j$ in a power grid is a complex parameter, which can be expressed as a sum of its real and imaginary parts. Therefore, the real part is denoted as $G_{k,j}$ and represents the conductance of the line between buses $k$ and $j$, while the imaginary part is denoted as $B_{k,j}$ and represents the susceptance of the line.

The system is balanced when the power flows leaving each bus are equal to the grid injection at that bus:
\begin{subequations}\label{eq:ch2_ps_10}
\begin{align}
 p_k = v^2_k G_{kk} + \sum_{j=1}^{m} p_{k, j}  \label{eq:ch2_ps_10.1}\\
 q_k = -v^2_k B_{kk} + \sum_{j=1}^{m} q_{k, j} \label{eq:ch2_ps_10.2},
\end{align}
\end{subequations}
where $p_k$ and $q_k$ are active and reactive power injections at bus $k$, while the first term ($v^2_k G_{kk}$ and $v^2_k B_{kk}$) represents bus shunt elements. 

The power flow equation serves to determine the unknown variables in the power grid. These variables can be either inputs or outputs of the power flow equations, depending on the type of bus they are associated with. Known variables represent inputs in the power flow equation system, while unknown ones (outputs) should be calculated. There are three types of buses: Slack, PV bus and PQ bus (Table.~\ref{tab:ch2_ps_1}).

\begin{table}[h!]
    \centering
    \begin{tabular}{lccccc}
    \hline
    Bus Type & Bus properties &
    $\theta_k$ & $v_k$ & $p_k$ & $q_k$\\
    \hline
    \textbf{Slack} & at least one per grid & \textbf{known} & \textbf{known} & unknown & unknown\\
    \textbf{PV} & connected generation & unknown & \textbf{known} & \textbf{known} & unknown \\
    \textbf{PQ} & no connected generation & unknown & unknown & \textbf{known} & \textbf{known} \\
        \hline
    \end{tabular}
    \caption[Summary of the variables computed by a power flow]{Bus description and variables that are calculated in the power flow calculation process depending on the type of bus.}
    \label{tab:ch2_ps_1}
\end{table}

A \textbf{Slack} bus is a bus where the voltage angle $\theta_k$ and magnitude $v_k$ are known and considered as inputs. At least one bus in the network must be a slack bus, where the voltage angle is set to 0 degrees ($\theta_k = 0$). The remaining buses in the network can be either PV or PQ buses.

A \textbf{PV} bus is a bus where the active power $p_k$ injection and voltage magnitude $v_k$ are known and considered as inputs. This type of bus is usually connected to a conventional generator unit such as thermal or hydropower, which can control the bus voltage.

A \textbf{PQ} bus is a bus where the active $p_k$ and reactive power $q_k$ injections are known and considered as inputs. This type of bus is usually connected to a renewable energy source (solar or wind) or has no generation unit connected to it. These buses are not voltage controlled and, therefore, the voltage magnitude is an output of the power flow equation.

In summary, the power flow equation is a crucial tool in determining the unknown variables of an electrical power system. These variables include voltage magnitudes $v_k$ and angles $\theta_k$, as well as power flow ($p_{k,j}$ and $q_{k,j}$) in the transmission lines. To obtain these values, a process called power flow computation is required. Power flow computation involves solving a sequence of nonlinear equations using a simulation engine. This engine takes in the known inputs, such as generator and load information, and uses them to calculate the unknown variables. The process is iterative, and the equations are solved until convergence is achieved, meaning that the computed values satisfy the set of equations.

\subsection{DC Approximation} \label{dc-approx}

In this section, we will provide a brief overview of the direct current (DC) approximation, which is a common approximation technique in power flow modeling. Understanding this approximation will be helpful for readers to grasp the concepts discussed in the next section~\ref{ch2_opf} and chapter~\ref{ch:5}.

The DC approximation involves linearizing the power flow equations to make the problem convex. Although it has certain limitations, particularly in voltage-prone power grids, it offers two primary benefits. Firstly, the DC approximation always provides a solution and cannot diverge, ensuring its reliability. Secondly, it is computationally fast, making it an attractive option for real-time operation in the power grid, which is why transmission system operators TSOs often utilize it. Additionally, the DC approximation can also be applied in other areas such as grid development planning.

Comparing the DC power flow approximation and the AC power flow model from~\ref{eq:ch2_ps_9}, there are three significant simplifications made in the former:
\begin{enumerate}
    \item $R \ll X$, resistance $R$ is much less than reactance $X$ in a line;
    \item $\theta_k - \theta_j \thickapprox 0$, the difference of angles between two connected busses is small;
    \item $v_k \thickapprox v_{nom}$, the voltage amplitude is close to the nominal voltage at each bus.
\end{enumerate}

The first simplification $R \ll X$ neglects the losses in the system solved by the DC approximation. This simplification can be justified by the fact that in high-voltage grids reactance $X$ is higher than resistance $R$ in line. It directly affects the admittance of the transmission line $y$. As previously mentioned, admittance $y$ is the reciprocal of impedance $z$ where the impedance is the complex number of resistance $R$ as the real part and reactance $X$ as the imaginary part. Accordingly, admittance $y$ is equal: 
\begin{equation}
 \label{eq:ch2_ps_11}
 y_{k,j} = \frac{1}{z_{k,j}} = \frac{1}{R_{k,j} + \mathbf{i}X_{k,j}}
= \frac{R_{k,j}}{R_{k,j}^2 + X_{k,j}^2} - \mathbf{i}\frac{X_{k,j}}{R_{k,j}^2 + X_{k,j}^2}
\end{equation}
where conductance $G_{k,j}$ and susceptance $B_{k,j}$ are defined as:
\begin{subequations}\label{eq:ch2_ps_12}
\begin{align}
G_{k,j} = \frac{R_{k,j}}{R_{k,j}^2 + X_{k,j}^2}
 \\
 B_{k,j} = \frac{- X_{k,j}}{R_{k,j}^2 + X_{k,j}^2} 
\end{align}
\end{subequations}
Since simplification implies that $R \ll X$ we get:
\begin{subequations}\label{eq:ch2_ps_13}
\begin{align}
G_{k,j} \to 0
 \\
 B_{k,j} \to \frac{-1}{X_{k,j}}
\end{align}
\end{subequations}
Accordingly, the AC power flow equations from~\ref{eq:ch2_ps_9} are simplified into:
\begin{subequations}\label{eq:ch2_ps_14}
\begin{align}
 p_{k,j} = v_k v_j \left( B_{k,j}\sin(\theta_{k} - \theta_{j})\right) \\
 q_{k,j} = v_k v_j \left( B_{k,j}\cos(\theta_{k} - \theta_{j})\right)
\end{align}
\end{subequations}

The second simplification $\theta_k - \theta_j \thickapprox 0$ linearized the trigonometric functions sin and cos, giving:  
\begin{subequations}\label{eq:ch2_ps_15}
\begin{align}
 p_{k,j} = v_k v_j \left( B_{k,j}(\theta_{k} - \theta_{j})\right) \\
 q_{k,j} = v_k v_j B_{k,j}
\end{align}
\end{subequations}

Finally, the last simplification $v_k \thickapprox v_{nom}$ made the voltage amplitude $v$ constant and therefore the power flow model completely linear resulting in:
\begin{subequations}\label{eq:ch2_ps_16}
\begin{align}
 p_{k,j} =  B_{k,j}(\theta_{k} - \theta_{j}) \label{eq:ch2_ps_16.1}\\
 q_{k,j} = B_{k,j} \label{eq:ch2_ps_16.2}
\end{align}
\end{subequations}

Eq.~\ref{eq:ch2_ps_16.2} can be ignored since the parameter $B$ is known for each line. Accordingly, the first equation (Eq.~\ref{eq:ch2_ps_16.1}) plays a key role in solving the DC power flow as a function of unknown $\theta$. Following the vectorized equation~\ref{eq:ch2_ps_4} for the AC power flow, in a similar way, we can represent it for the DC power flow as:
\begin{equation}
 \label{eq:ch2_ps_17}
 \begin{bmatrix} p_{1}  \\ \cdot \cdot \\ p_{j} \\ \cdot \cdot \\ p_{m}\end{bmatrix}
 =
 Y
\begin{bmatrix} \theta_1  \\ \cdot \cdot \\ \theta_j \\ \cdot  \cdot \\ \theta_m \end{bmatrix}
\end{equation}
This system of linear equations can be easily solved using any linear solver.

\section{Optimal Power Flow} \label{ch2_opf}

In this section, we will talk about optimal power flow (OPF) which consists of the power flow as the key component of the problem. OPF is an essential tool for power markets and power security, which involves solving optimization problems based on power flow equations. It is performed annually for system planning or daily, hourly, or even every five minutes for day-ahead planning. OPF was first introduced by Carpentier in 1962, where he dealt with the economic dispatch problem~\cite{carpentier1962contribution}. Since then, the formulation of OPF has evolved to solve various types of problems.

In general, OPF involves constrained optimization problems that consist of variables optimizing an objective function, equality constraints such as power balance and power flow equations, and inequality constraints including variable bounds. Depending on the type of OPF, the variables, constraints, and objective function may change.

The AC-OPF is the key method for solving Transmission System Operator (TSO) problems, but it remains challenging. It is complex economically, electrically, and computationally due to the need for multi-part nonlinear pricing in an efficient market, additional non-linearities caused by AC-PF, and its non-convexity, including binary and integer variables, that increases computational complexity in optimization. Despite several decades since OPF is formulated, fast and robust solution techniques for solving the full AC-OPF are still lacking.

OPF can solve various types of problems such as economic dispatch, unit commitment, optimal topology, and long-term planning. In this section, we focus on the economic dispatch problem with continuous variables since it is discussed further in the contribution part. We do not discuss other problems involving binary and integer variables. Moreover, in this section, we review deterministic and stochastic OPF and state-of-the-art solutions in this field.

\subsection{Deterministic OPF}

The deterministic or conventional OPF problem can be formulated in an abstract form as follows:
\begin{subequations}\label{eq:det_opf}
\begin{align}
 &\min_{p_g, q_g, v, \theta} 
 f(p_g) \label{eq:det_opf_a}
 \\
 &\text{s.t.~} F(p, q, v, \theta) = 0, \label{eq:det_opf_b}\\
 &~~~~~ p_{g,i}^{\min} \leq p_{g,i} \leq p_{g,i}^{\max}, \quad \forall g\in {\cal \mathcal{G}} \label{eq:det_opf_c}\\
 &~~~~~ q_{g, i}^{\min} \leq q_{g, i} \leq q_{g, i}^{\max}, \quad \forall i\in {\cal \mathcal{G}} \label{eq:det_opf_d}\\
 &~~~~~ v_{j}^{\min} \leq v_{j}\leq v_{j}^{\max}, \quad \forall j\in \mathcal{B} \label{eq:det_opf_e}\\
 &~~~~~ s_{k,j} \leq s_{k,j}^{\max}, ~~~~~~~~~~ \forall k \sim j  \label{eq:det_opf_f}\\
 &~~~~~ \theta_{slack} = 0  \label{eq:det_opf_g}
\end{align}
\end{subequations}
where $f$ is the objective (cost) function; $F$ is the AC power flow balance equations from~\ref{eq:ch2_ps_10}; $p_g$ and $q_g$ are active and reactive power injections of the controlable generation; $\theta_{slack}$ is voltage angel in Slack bus; $s_{k,j}$ is a value of apparent power flow along the line $e_{k,j}$ defined as $s_{k,j} = \sqrt{p^2_{k,j} + q^2_{k,j}}$.

The corresponding OPF formulation~\ref{eq:det_opf} has been adopted for the economic dispatch problem. The economic dispatch problem aims to determine the most cost-effective output levels for power generators while satisfying the physical and load-related limitations of the power system. To achieve this, an optimization problem is formulated with a cost function~\ref{eq:det_opf_a}, which is minimized to find the optimal solution for the economic dispatch OPF problem. Usually, the cost function of the economic dispatch OPF is the function of the total real power generation $p_g$. This cost function is typically represented by a quadratic or piecewise linear function. For the purpose of this discussion, we will focus on the quadratic cost function, which is expressed as follows.
\begin{gather}\label{eq:det_opf_obj}
f(p_g) = \sum_{i\in \mathcal{G}} \left\{ c_{2,i} p^2_{g,i} + c_{1,i} p_{g,i} + c_{0,i}\right\}
\end{gather}
where $\{c_{2,i}, c_{1,i}, c_{0,i}\}_{i \in \mathcal{G}} \geq 0 $ are scalar cost coefficients.

Equality constraints~\ref{eq:det_opf_b} in an optimization problem are mathematical expressions that represent the relationship between the optimization variables, usually in the form of equations. In the context of an AC-OPF, the power flow equations~\ref{eq:ch2_ps_9} and power balance equations~\ref{eq:ch2_ps_10} are fundamental physical laws that must be incorporated as equality constraints. The voltage angle differences are of primary importance than individual angles in the power flow equations. Accordingly, the voltage angle of the slack bus~\ref{eq:det_opf_g} can be selected as an additional equality constraint by setting its value to zero. This simplifies the calculation of voltage angle differences and facilitates the solution of the OPF problem.

Inequality constraints, which represent the physical limitations of the system elements, are also included in the OPF problem. These constraints also represent the function of the optimization variables by restricting the function value within a specified range. For example, in the context of the economic dispatch problem, the most common inequality constraints reflect the physical limitations of the system, such as minimum and maximum power output constraints~\ref{eq:det_opf_c} 
-~\ref{eq:det_opf_d}, voltage level constraints~\ref{eq:det_opf_e} and transmission line flow constraints~\ref{eq:det_opf_f}. These constraints ensure that the solution to the optimization problem is feasible and physically realizable.

During steady-state operations at PV buses, the AC-OPF in~\ref{eq:det_opf} returns the generation set points $p_g$ and $q_g$, while voltage magnitudes $v$ need to be fixed. On the other hand, active and reactive power loads are typically fixed for PQ buses.

\subsubsection{Interior-point method} Solving the economic dispatch optimization problem in~\ref{eq:det_opf} is challenging, as it involves non-linear and non-convex functions. However, one of the most effective techniques for solving large-scale optimization problems, and in this case, a full AC-OPF economic dispatch problem like this in~\ref{eq:det_opf} is the interior-point method~\cite{interior1, interior2}. In this section, we will give a brief explanation of the main aspects of this method.

Mathematical optimization involves finding the optimal values of decision variables that minimize the objective function, subject to a set of constraints, including equality and inequality constraints. Accordingly, to explain how the interior point method works, we introduce simplified expressions where $x$ is the decision variable, $c_{eq}$ represents all equality constraints, and $c_{in}$ symbolizes all inequality constraints. Therefore, the problem of mathematical optimization can be written in a simplified form as follows:
\begin{subequations}\label{eq:int-point}
\begin{align}
 &\min_{x} 
 f(x) \label{eq:int-point_a}
 \\
 &\text{s.t.~} c_{eq}(x) = 0 \label{eq:int-point_b}\\
 &~~~~~ c_{in}(x) \geq 0 \label{eq:int-point_c}
\end{align}
\end{subequations}

Barrier methods and interior-point methods are often used interchangeably because interior-point methods create a surrogate model of the optimization problem. This model is created by transforming inequality constraints into equality constraints and substituting the objective function with a logarithmic barrier function. Accordingly, formulation~\ref{eq:int-point} can be rewritten as follows:
\begin{subequations}\label{eq:barier}
\begin{align}
 &\min_{x, s_{sl}} 
 f(x) - \psi\sum_{i=1}^{|c_{in}|} log(s_{sl_i}) \label{eq:barier_a}
 \\
 &\text{s.t.~} c_{eq}(x) = 0 \label{eq:barier_b}\\
 &~~~~~ c_{in}(x) - s_{sl} = 0 \label{eq:barier_c}
\end{align}
\end{subequations}
where the objective function includes a barrier term with a barrier parameter $\psi$ ($\psi > 0$) and the slack variables $s_{sl}$ represented as a vector. From formulation~\ref{eq:barier}, it is obvious that the barrier term enforces an inequality constraint on the slack variables $s_{sl_i} \geq 0$ through the logarithmic function. While the barrier problem and the original non-linear program from~\ref{eq:int-point} are not equivalent, the solution of the barrier problem approaches the solution of the original optimization problem as the barrier parameter approaches zero. In other words, the barrier problem provides an approximate solution to the original problem by gradually relaxing the constraint until it is no longer a factor.

The Karush-Kuhn-Tucker (KKT) conditions~\cite{gordon2012karush} are fundamental to the basic interior-point algorithm, which is used to solve optimization problems. These conditions involve four sets of equations, where the first two originate from the system Lagrangian's first-order condition. Thus, the first two sets of equations require the derivatives with respect to the decision and slack variables to be equal to zero. The third set of equations relates to the equality constraints~\ref{eq:barier_b}, while the fourth set refers to the inequality constraintc~\ref{eq:barier_c}. KKT conditions are formulated as follows:
\begin{subequations}\label{eq:kkt}
\begin{align}
 & \bigtriangledown f(x) - J^T_{eq}(x)r - J^T_{in}(x)q \\
 & \bigtriangledown Sq - \psi e_{ind} = 0\\
 & ~ c_{eq}(x) = 0 \\
 & ~ c_{in}(x) - s_{sl} = 0 
\end{align}
\end{subequations}
In KKT formulation, $J^T_{eq}(x)$ and $J^T_{in}(x)$ are transposed Jacobian matrices of the equality and inequality functions, respectively; $r$ and $q$ are corresponding Lagrange multipliers; $S$ is a diagonal matrix of the slack variables ($S=diag(s_{sl})$); $e_{ind}$ is a vector of diagonal elements of identity matrix $I$ ($e_{ind} = [1,...,1]^T$).

The KKT conditions create a non-linear system by realizing a relationship between the primal ($x$, $s_{sl}$) and dual variables ($r$, $q$). Concisely, a non-linear system can be expressed by a vector-valued function $F_{KKT}=(x, s_{sl}, r, q) = 0$. This system needs to be solved to obtain the optimal solution. One common approach to solving the non-linear system is to use the Newton-Raphson method. This method can be applied by looking for a search direction based on the following linear problem:
\begin{equation}
 \label{eq:ntw_raps}
 J_{KKT}(x, s_{sl}, r, q)p_{dir} = -F_{KKT}(x, s_{sl}, r, q)
\end{equation}
where $J_{KKT}$ is the Jacobian matrix of the KKT system; $p_{dir}$ is the search direction vector. The full form of the $J_{KKT}$ is:
\begin{equation}
 \label{eq:J_kkt}
 J_{KKT}(x, s_{sl}, r, q) = \begin{bmatrix} \bigtriangledown_{xx}L & 0 & -J^T_{eq}(x) & -J^T_{eq}(x) \\  0 & Q & 0 & S \\ J^T_{eq}(x)  & 0 & 0 & 0 \\ J^T_{eq}(x) & -I & 0 & 0 \end{bmatrix}
\end{equation}
where $Q = diag(q)$; $\bigtriangledown_{xx}L$ is the Hessian of Lagrangian system $L$ defined as:
\begin{equation}
\label{eq:solve}
   L(x, s_{sl}, r, q) = f(x) - r^T c_{eq}(x) - q^T (c_{in}(x) - s_{sl})
\end{equation}
The search direction vector $p_{dir}$ is:
\begin{equation}
  p_{dir} = \begin{bmatrix} p_{dir, x} \\ p_{dir, s_{sl}} \\ p_{dir, r} \\ p_{dir, q} \\
  \end{bmatrix} 
\end{equation}

Iteratively applying the Newton-Raphson method to a formulation~\ref{eq:solve}, also called a primal-dual system, allows converging to the optimal solution of the optimization problem. It means that in iteration $n+1$ we obtain $p^{n+1}_{dir}$ and accordingly update variables ($x$, $s_{sl}$, $r$, $q$) and barrier parameter $\psi$ until convergence criteria is not satisfied, as follows:
\begin{subequations}
\begin{align}
 & x^{n+1} = x^{n} + \alpha^{n+1}_{dir, x} p^{n+1}_{dir, x}\\
 & s_{sl}^{n+1} = s_{sl}^{n} + \alpha^{n+1}_{dir, s_{sl}} p^{n+1}_{dir, s_{sl}}\\
 & r^{n+1} = r^{n} + \alpha^{n+1}_{dir, r} p^{n+1}_{dir, r}\\
 & q^{n+1} = q^{n} + \alpha^{n+1}_{dir, q} p^{n+1}_{dir, q}\\
 & \psi^{n+1} = f_{\psi}(\psi^{n}, x^{n+1}, s_{sl}^{n+1}, r^{n+1}, q^{n+1})
\end{align}
\end{subequations}
where $\alpha^{dir}$ is the line search parameter for different variables;  
$f_{\psi}$ is a function that computes the updated value of $\psi$ from the previous value and the updated variables. Various methods have been studied to enhance the convergence rates of optimization algorithms. These methods utilize different heuristic functions $f_{\psi}$ for calculating line search and barrier parameters.

The interior-point method is a common and reliable approach for solving AC-OPF problems. However, solving AC-OPF with the interior-point method can be costly due to the need for computing the Hessian of the Lagrangian $\bigtriangledown_{xx}L$ at each iteration step. Solving large-scale problems with the interior-point method can be particularly challenging, as the computational time scales superlinearly with system size. To overcome these challenges, there are several strategies available. One approach is to simplify the original OPF problem by using DC power flow equations, while another is to compute approximate solutions. These strategies can make the problem more manageable and reduce the computational burden.

\subsubsection{Convex relaxations}
Convex relaxations are a set of approximations used to tackle non-convex optimization problems. The underlying concept is to approximate the original problem with a convex optimization problem, which is easier to solve. In contrast to non-convex problems, where local minima may not be the global minimum, in convex problems, any local minimum is also the global minimum. Thus, there are several efficient algorithms available with excellent convergence rates to solve convex optimization problems \cite{convex1}.

Convex relaxations can be illustrated through the economic dispatch problem, which can be formulated using the basic BIM approach. With some slight adjustments, the economic dispatch problem can be reformulated into a quadratic programming (QP) problem using the complex voltages of the buses. In the QP optimization problem utilized for convex relaxations, the objective function and all inequality constraints have a quadratic form in the complex voltages~\cite{relaxation1, relaxation2}. Thus QP form of the convex relaxations is:
\begin{subequations}\label{eq:conv-rel}
\begin{align}
 &\min_{\bar{v} \in \mathbb{C}} 
 \bar{v}^{*}C\bar{v} 
 \\
 &\text{s.t.~} \bar{v}^{*}M_{herm} \bar{v} \leq m_{herm} 
\end{align}
\end{subequations}
where $M_{herm}$ is Hermitian matrix; $m_{herm}$ is real-valued vectors of the corresponding inequality constraints; sunbscript $*$ denotes conjugate transpose. However, QP formulation~\ref{eq:conv-rel} is still a non-convex. To make the QP problem convex, the expression $\bar{w} = \bar{v}\bar{v}^{*}$ is included in formulation~\ref{eq:conv-rel}. $\bar{w}$ is positive semidefinite matrix with rank 1 ($\bar{w} \succeq 0$, rank($\bar{w}$) = 1). By utilizing the identity of the trace function for any Hermitian matrix, formulation~\ref{eq:conv-rel} can be reformulated as:
\begin{subequations} \label{eq:sdp}
\begin{align}
 &\min_{\bar{w} \in \mathbb{S}} 
 tr(C\bar{w}) 
 \\
 &\text{s.t.~} tr(M_{herm} \bar{w}) \leq m_{herm} \\
 & \bar{w} \succeq 0 \\
 & rank(\bar{w}) = 1
\end{align}
\end{subequations}
where $\mathbb{S}$ is the space of $m$ x $m$ symmetric matrices. Finally, the formulation~\ref{eq:sdp} is convex in $\bar{w}$ and represents a semidefinite programming problem (SDP), known as the SDP relaxation of the OPF. Convex relaxation approaches possess an advantage over other approximation techniques in that the infeasibility of the relaxed problem directly implies the infeasibility of the original problem. Despite the recent success of regularized semidefinite programming~\cite{krechetov2018entropy}, which allows to produce tighter bounds, their scalability does not allow applications to large-scale power systems problems. 

\subsubsection{Linearized DC-OPF} 
The DC-OPF approach is a commonly used approximation in power systems. This approach implies replacing the AC power flow balance equation with the DC power flow balance equation described in~\ref{dc-approx}. The DC approximation linearized the original OPF problem by removing several variables and constraints (see Eq.~\ref{eq:ch2_ps_16}). Accordingly, DC-OPF creates a linear programming problem that can be solved very efficiently by interior-point methods or simplex methods. However, one major limitation is that the solution obtained from DC-OPF may not be feasible in AC-OPF~\cite{dc_opf1}. This leads to restarting the DC-OPF calculation and tightening some constraints.

\subsection{Stochastic OPF}

This section discusses stochastic OPF techniques that handle the uncertainty of power injections at the bus. This uncertainty primarily arises from the increasing use of RES like wind and solar, as well as the implementation of smarter grids that allow for the deferral of load and storage devices. Stochastic optimal power flow is a mathematical optimization technique used in power systems to determine the optimal operating conditions of the system under uncertainty. It considers the stochastic nature of the power grid variables, such as the fluctuating load and the uncertain renewable energy sources (RES). Accordingly, firstly we need to understand how these variables are modeled under uncertainty. 

The deterministic AC OPF assumes that the power injections are precisely known at the time of scheduling. In practice, this is not true, as both load and renewable generation might vary from their forecasted value. Deviations in forecasted values are usually caused by forecast errors, external fluctuations, or intra-day electricity trading. For this reason, it is important to account for the impact of injection uncertainties on system operation to ensure secure operation. Accordingly, we assume that active power uncertainties are modeled as the sum of the deterministic forecasted value $p$ and a real-time fluctuation $\omega$ as:
\begin{gather} \label{eq:uncert_1}
 {p}_{l}(\omega) = p_{l} + \omega\\
 {p}_{r}(\omega) = p_{r} + \omega
\end{gather}
where $\omega$ is an independent random variable with zero mean and known standard deviation $\sigma_{\omega}$; $p_l$ and $p_r$ are load and RES active power injections. In this thesis, the reactive power injections of load and RES correspond to the same uncertainty realization while maintaining a given constant power factor $cos(\phi)$ as: 
\begin{gather}\label{eq:uncert_2}
 {q}_{l}(\omega) = \gamma{p}_{l}(\omega) = \gamma p_{l} + \gamma \omega\\
 {q}_{r}(\omega) = \gamma{p}_{r}(\omega) = \gamma p_{r} + \gamma \omega
\end{gather}
where $\gamma = \sqrt{\frac{1-cos^2(\phi)}{cos^2(\phi)}}$ is the power ratio. Thus, the ratio of active and reactive power injections remains unchanged during fluctuations.

Following uncertainty realizations $\omega$, the controllable generators adapt their generation to maintain the total power balance and feasibility. We use an affine
policy representing the automatic generation control (AGC) \cite{roald2017chance} to balance active power generation. Therefore, the total power mismatch $\Omega = \sum_{j\in \mathcal{G}} \omega_j$ is divided among the generators according to the participation factors $\alpha$ based on the following generation control policy:
\begin{subequations}\label{eq:uncert_3}
\begin{align}
    p_{g,j}(\omega) = p_{g,j} + \alpha_j \Omega, ~~~~~~ \forall j \in \mathcal{G}  \\
    \sum_{j\in \mathcal{G}} \alpha_j = 1   ~~~~~~~~~~~~~~~~~~~~~~~~~~~~
\end{align}
\end{subequations}

The reactive power is controlled locally at buses $\mathcal{G}$, keeping their outputs constant at buses $\mathcal{L}$. Generators at buses $\mathcal{G}$ change reactive power outputs by $\delta q(\omega)$ to keep voltage magnitudes constant on these buses,
\begin{subequations}\label{eq:uncert_4}
\begin{align}
    q_{j}(\omega) = q_{j} + \delta q_j(\omega) , ~~~~~~~~~~ \forall j \in \mathcal{G}  \\
    q_{j}(\omega) = q_{j}, ~~~~~~~~~~~~~~~~~ \forall j \in \mathcal{L}
\end{align}
\end{subequations}

In contrast, the voltage magnitudes is fixed at buses $\mathcal{G}$ and fluctuates at buses $\mathcal{L}$,
\begin{subequations}\label{eq:uncert_5}
\begin{align}
    v_{j}(\omega) = v_{j} + \delta v_j(\omega) , ~~~~~~~~~~ \forall j \in \mathcal{L}  \\
    v_{j}(\omega) = v_{j}, ~~~~~~~~~~~~~~~~~ \forall j \in \mathcal{G}
\end{align}
\end{subequations}

Let $p_{k,j}$ and $q_{k,j}$ denote the active and reactive power flows from bus $k$ to bus $j$ along line $(k, j) \in \mathcal{E}$. In the AC power flow formulation, the active and reactive power flows on each transmission line depend non-linearly on the voltage magnitudes $v$ and voltage angles $\theta$. Consider all possible fluctuations $\omega$ within the uncertainty set $W$, the power flow equations from~\ref{eq:ch2_ps_9} are given by:  
\begin{subequations}\label{eq:uncert_6}
\begin{align}
    p_{k,j}(\omega) = v_k(\omega) v_j(\omega)\left(G_{k,j}cos\left(\theta_{k,j}(\omega)\right) + B_{k,j}sin\left(\theta_{k,j}(\omega)\right)\right) \\
    q_{k,j}(\omega) = v_k(\omega) v_j(\omega)\left(G_{k,j}sin\left(\theta_{k,j}(\omega)\right) - B_{k,j}cos\left(\theta_{k,j}(\omega)\right)\right)
\end{align}
\end{subequations}
where $\theta_{k,j}(\omega) = \theta_k(\omega) - \theta_j(\omega)$. According to~\ref{eq:ch2_ps_10}, power balanced equation under uncertainty is given as:
\begin{subequations}\label{eq:uncert_7}
\begin{align}
    p_k(\omega) = v^2_k(\omega)G_{k,k} + \sum_{j}^{m} p_{k,j}(\omega) \\
    q_k(\omega) = -v^2_k(\omega) B_{k,k} +  \sum_{j}^{m} q_{k,j}(\omega)
\end{align}
\end{subequations}

Deviations in power injections cause changes in power flows throughout the system. Accordingly, the apparent power flow fluctuates in the lines as:

\begin{equation}\label{eq:uncert_8}
    s_{k,j} (\omega) = s_{k,j} + \delta s_{k,j}(\omega), ~~~~~~~~ \forall k\sim j \in \mathcal{E}
\end{equation}

Finally, similar to Eg.~\ref{eq:det_opf} and taking into account Eg.~\ref{eq:uncert_1} -~\ref{eq:uncert_8}, the stochastic OPF can be define as: 
\begin{subequations}\label{eq:stoch_OPF}
\begin{align}
 & \min_{p_g, q_g, v, \theta, \alpha} 
 f(p_g(\omega))
 \\
 &\text{s.t.~} F(p(\omega), q(\omega), v(\omega), \theta(\omega)) = 0, ~~\forall \omega \in W \\
 &~~~~~ p_{g,i}^{\min} \leq p_{g,i}(\omega) \leq p_{g,i}^{\max}, \quad ~~~~~~~~~ \forall g\in {\cal \mathcal{G}} \\
 &~~~~~ q_{g, i}^{\min} \leq q_{g, i}(\omega) \leq q_{g, i}^{\max}, \quad ~~~~~~~~~ \forall i\in {\cal \mathcal{G}} \\
 &~~~~~ v_{j}^{\min} \leq v_{j}(\omega) \leq v_{j}^{\max}, \quad ~~~~~~~~~ \forall j\in \mathcal{B} \\
 &~~~~~ s_{k,j}(\omega) \leq s_{k,j}^{\max}, ~~~~~~~~~~~~~~~~~~~ \forall k \sim j \\
 &~~~~~ \theta_{slack} = 0  
\end{align}
\end{subequations}

This section will focus on the chance-constrained (CC) approach for modeling and solving Stochastic OPF while acknowledging that there are other approaches such as robust optimization and probabilistic optimal power flow (P-OPF).

\subsubsection{Chance-constrained OPF}
Chance constrained approach uses probability-based constraints that rely on a specific probability distribution of uncertainty to limit the chances of constraint violations below a desired threshold. Essentially, these constraints ensure that the probability of violating any constraints remains under a set limit. The full AC chance-constrained OPF is formulated as follows:

\vspace{-.8em}
\begin{subequations}\label{cc-opf}
\begin{align}
 &\min_{p_g, q_q, v, \theta} \sum_{i\in \mathcal{G}} \mathbb{E}[c_i(p_{g,i}(\omega))] \\
 &\text{s.t.~} F(\theta(\omega), v(\omega), p(\omega), q(\omega)) = 0, ~~~~~~~ \forall \omega\in W  \label{cc-opf_b}\\
 &~~~~~ \mathbb{P}(p_{g_i}(\omega)  \leq p^{max}_{g_i}) \geq 1 - \epsilon_p, ~~~~~~~~~~ \forall i \in \mathcal{G} \label{cc-opf_c}\\
 &~~~~~ \mathbb{P}(p_{g_i}(\omega)  \geq p^{min}_{g_i}) \geq 1 - \epsilon_p, ~~~~~~~~~~ \forall i \in \mathcal{G}\\
 &~~~~~ \mathbb{P}(q_{g_i}(\omega)  \leq q^{max}_{g_i}) \geq 1 - \epsilon_q, ~~~~~~~~~~ \forall i \in \mathcal{G}\\
 &~~~~~ \mathbb{P}(q_{g_i}(\omega)  \geq q^{min}_{g_i}) \geq 1 - \epsilon_q, ~~~~~~~~~~ \forall i \in \mathcal{G}\\
 &~~~~~ \mathbb{P}(v_{j}(\omega)  \leq v^{max}_{j}) \geq 1 - \epsilon_v, ~~~~~~~~~~~ \forall i \in \mathcal{N}\\
 &~~~~~ \mathbb{P}(v_{j}(\omega)  \geq v^{min}_{j}) \geq 1 - \epsilon_v, ~~~~~~~~~~~ \forall i \in \mathcal{N}\\
 &~~~~~ \mathbb{P}(s_{kj}(\omega)  \leq s^{max}_{kj}) \geq 1 - \epsilon_s, ~~~~~~~~~~ \forall k\sim j \in \mathcal{E} \label{cc-opf_i} 
\end{align}
\end{subequations}
where expectations $\mathbb{E}$ and probabilities $\mathbb{P}$ are defined over the distribution of $\omega$. 

The power outputs of conventional generators, voltage magnitudes at buses and apparent power flow in the lines are constrained using the separate chance constraints~\ref{cc-opf_c}–\ref{cc-opf_i}. Moreover, chance constraints can be modeled as joint chance constraints, which ensure that all
constraints hold jointly with a pre-described probability~\cite{cc1, cc2}. The inequality constraints are formulated probabilistically, i.e., must be satisfied with the prescribed probability. The prescribed probability is controlled by a choice of acceptable violation probabilities $\epsilon_p$, $\epsilon_q$, $\epsilon_v$ and $\epsilon_s$. Accordingly, the concept of chance constraint is used to define an $\epsilon$-reliability set, which allows TSOs to safely execute additional control actions within that set, with a low probability of violating a constraint. A lower acceptable violation probability $\epsilon$ ensures more reliable system operation, but at a higher cost. In contrast, a higher acceptable violation probability $\epsilon$ is riskier and provides no guarantee that such controls will be available.

CC-OPF formulation~\ref{cc-opf} is intractable and causes challenging computation as it involves computing multi-dimensional probability integrals. However, recent advancements in chance-constrained optimization have inspired the creation of state-of-the-art CC-OPF applications~\cite{cc_state_of_art_1, cc_state_of_art_2}.

The scenario-based approach is a commonly used method for solving the CC-OPF~\cite{cc_state_of_art_1}. This approach involves replacing the chance constraint with a finite set of constraints that correspond to different potential outcomes of the uncertain parameters. The key advantage of this approach is that it assumes that all relevant functions are convex with respect to the decision variables, making the problem easier to solve. The scenario-based approach provides probabilistic guarantees by using a set of scenarios that increases linearly with the number of decision variables. In simpler terms, this method uses a finite set of constraints to represent different potential outcomes of uncertain parameters, and it assumes that the problem is solvable through convex functions. There are also some alternative approaches to the scenario-based approach. One of them is given in~\cite{cc_state_of_art_2}. This approach solves a robust problem with bounded uncertainty, where bounded uncertainties are computed using a scenario approach. In addition, convexity is not required in this approach, and the number of scenarios does not depend on the number of decision variables. Accordingly, this alternative approach reduces computational costs or provides less conservative guarantees. In \cite{lukashevich2021power}, the authors used internal approximations to the feasibility set later on providing more efficient algorithms to generate samples (scenarios) for reducing the complexity of the optimization methods \cite{owen2019importance,lukashevich2021importance,lukashevich2021power}. Furthermore, the resulting deterministic problem may have a special structure allowing for more efficient numerical methods~\cite{anikin2022efficient,krechetov2018entropy}

Since CC-OPF with full AC power flow equation is very difficult to solve, relaxation and linear approximation are often used to make problems easier. Full AC power flow equation in CC-OPF with a polynomial chaos expansion is proposed in~\cite{muhlpfordt2019chance}. Analytical reformulation of the CC-ACOPF based on linear sensitivity factors is proposed in~\cite{schmidli2016stochastic}. In~\cite{roald2017chance} and~\cite{roald2017power}, a deterministic AC-OPF problem was proposed that is solved iteratively and estimates the uncertainty limits by a linear approximation at the operating point. However, this method cannot guarantee the convergence of the algorithm and finding the global optimum. 

In recent years, convex relaxation has been used to solve the AC CC-OPF problem~\cite{halilbavsic2018convex, venzke2017convex, baker2017efficient, lubin2019chance, xie2017distributionally}. Semi-definite programming (SDP) relaxation and sample-based methods are used in~\cite{venzke2017convex}. However, it is still difficult to solve the problem reformulated in this way, since all the relaxed solutions in the set of uncertainties have a physical meaning. Since SDP relaxation is computationally demanding, the second-order cone programming (SOCP) relaxation is applied in~\cite{halilbavsic2018convex} to achieve faster computations than SDP in~\cite{venzke2017convex}. In addition, in~\cite{baker2017efficient} the convex optimization problem is reformulated from a non-convex joint CC-OPF using the improved Boole’s inequality. The SOCP problem is also addressed in~\cite{lubin2019chance} by linearizing the power balance equation around the forecasted operation point. Similarly, an exact SOCP is proposed in~\cite{xie2017distributionally} to reformulate a distributionally robust CC-OPF with known first and second moments. However, convex relaxation methods have not yet been developed for practical application in industry. Furthermore, CC-OPF problem can be combined with the contingency analysis and detection~\cite{anikin2022efficient,burashnikova2022ranking,mikhalev2020bayesian,stulov2020learning}.

The literature review indicates that DC CC-OPF based on a linearized DC power flow equation is widely used due to its impressive efficiency and scalability~\cite{bienstock2014chance, roald2016corrective, lubin2015robust, hou2020chance, pena2020dc}. However, these methods oversimplify the power system by neglecting voltage and reactive power, which can lead to solutions that deviate significantly and compromise the secure operation of the system. This is especially depicted in systems with a strong relationship between active and reactive power~\cite{castillo2015successive}. To address this issue, a more accurate linearization method for power flow equations is needed, particularly in stochastic scenarios. This requires retaining the essential characteristics of the AC power flow model as much as possible while making the model deterministic and linear. Such an approach would provide a more reliable and accurate solution, especially in complex power systems with significant uncertainties. We also should mention a series of papers that potentially allow us to reduce the computational time in the above-mentioned problems~\cite{grinis2022differentiable, kadilenko2023adjoint,grinis2016soliton} based on efficient approximations of complicated numerical problems.

In summary, the CC-OPF approach offers potential cost savings by allowing violations of operational constraints with low probability, while still meeting forecasted operating conditions. This approach is particularly useful for intraday operations where forecast uncertainty is low, but may not be as suitable for day-ahead operations where uncertainty is higher~\cite{capitanescu2016critical}. However, the disadvantage of CC-OPF is that still solving the full AC CC-OPF is challenging and computationally intensive, while approximation methods can compromise the safe operation of the system and are still not available in practice. To overcome these issues, the contribution of this thesis is a developing novel data-driven based approach (part II) to address these drawbacks.

\section{Synthetic dataset generation} \label{ch2_dataset}

In this section, we describe how we created a synthetic dataset by simulating a power system. This generated dataset is essential for the contribution part of our research. One significant advantage of using a synthetic dataset is that it allows us to conduct a controlled experiment where we have complete knowledge and control over all the variables involved.

Motivated by~\cite{donnot2019deep}, we generated the synthetic dataset using three different parts of injection sampling:
\begin{enumerate}
    \item active power loads and RES;
    \item reactive power loads and RES
    \item active power (controllable) generation
\end{enumerate}

\subsection{Sampling active power loads and RES}
To simulate real active powers for $i$-th load and $j$-th renewable source, we define stochasticity as:
\begin{subequations}
\begin{align}
 p_l^i = \eta^i p_{l_{ref}}^i, ~ i \in \mathcal{L} \\
 p_{r}^j = \nu^j p_{r_{ref}}^j, ~ j \in \mathcal{L} 
\end{align}
\end{subequations}
where $p_{l_{ref}}^i$ and $p_{r_{ref}}^j$ are fixed reference values of loads and RES; $\eta^{i}$ and $\nu^{j}$ are a random variables that follow a required distribution. Further, we will consider log-normal distribution. As in~\cite{donnot2019deep}, we separate random variables $\eta^{i}$ and $\nu^{j}$ into two components:
\begin{subequations}
\begin{align}
 \eta^i = \eta_{corr} \eta^i_{uncorr}, ~ i \in \mathcal{L} \\
 \nu^j = \nu_{corr} \nu^j_{uncorr}, ~ j \in \mathcal{L}
\end{align}
\end{subequations}
where $\eta_{corr}$ and $\nu_{corr}$ denotes spatio-temporal correlations, while $\eta^j_{uncorr}$ and $\nu^j_{uncorr}$ are local variations. We propose that both variables have log-normal distribution. We consider correlated variables $\eta_{corr}$ at all loads and $\nu_{corr}$ at all RES to take the same values with mean and standard deviation, ($\mu(\eta_{corr}) = -1$, $\mu(\nu_{corr}) = 0.2$) and ($\sigma(\eta_{corr}) =0.1$, $\sigma(\nu_{corr}) =0.4$). In contrast, uncorrelated variables are different for each load and renewable source, distributed with means ($\mu(\eta_{corr}) = 1$, $\mu(\nu_{corr}) = 1$) and standard deviations ($\sigma(\eta_{corr}) =0.05$, $\sigma(\nu_{corr}) =0.3$).

\subsection{Sampling reactive power loads and renewable sources}
Reactive powers are sampled as a fixed power ratio from the active injections as follows:
\begin{subequations}
\begin{align}
 q_l^i = \gamma^i p_{l}^i, ~i \in \mathcal{L} \\ 
 q_{rs}^j = \gamma^j p_{rs}^j, ~j \in \mathcal{L}
\end{align}
\end{subequations}
The power ratio is derived form the reference values $\gamma^i = \frac{q_{l_{ref}}^i}{p_{l_{ref}}^i}$,~ $\gamma^j = \frac{q_{rs_{ref}}^j}{q_{rs_{ref}}^j}$.

\subsection{Sampling active power generation}
Since the voltages of the generation at PV buses are fixed to their nominal values, we only considered the sampling of the active power generation $p_g$. To sample generation correctly, we have to ensure that the power balance condition is satisfied: 
\begin{equation}\label{eq:los}
 \sum_{k=1}^{n_u} p^{k}_g = \sum_{i=1}^{n_L} p^{i}_l - \sum_{j=1}^{n_R} p^{j}_{rs} + losses\\ 
\end{equation}
Consequently, the total power generation is approximately some percentage $\varrho$ higher than the total demand, covering losses in the power system. In our case, this percentage $\varrho$ is calculated as the ratio of the referent values between the total generation and the total load for our test systems\footnote{For tested IEEE 9 bus system $\varrho$=1.0139 that means the total losses are 1.39\% of the total output. Similarly, for IEEE 39 bus system $\varrho$=1.0086 and IEEE 118 $\varrho$=1.0322}. Following~\cite{donnot2019deep}, we first formulate the part in which each production sample has to target the total demand increased by losses percentage as:
\begin{gather*}
 \Breve{p}_g^{k} = \psi^k \cdot \varrho \cdot \frac{\sum_{i=1}^{n_L} p^{i}_l}{\sum_{k=1}^{n_u} p^{k}_{g_{ref}}} \cdot p^{k}_{g_{ref}} 
\end{gather*}
where $\psi^k \sim \mathcal{U}[0.8,1.2]$ is the uniform distribution to ensure no deterministic behavior for the productions.

We applied the second step to ensure that the relation $\sum_{k=1}^{n_u} p^{k}_g + \sum_{j=1}^{n_R} p^{j}_{rs} = \varrho \cdot \sum_{i=1}^{n_L} p^{i}_l$ holds perfectly, as:
\begin{gather*}
 p_g^{k} = \Breve{p}_g^{k} \cdot \varrho \cdot \frac{\sum_{i=1}^{n_L} p^{i}_l}{\sum_{k=1}^{n_u} \Breve{p}^{k}_g} 
\end{gather*}
\section{Conclusion} \label{ch2_conclusion}
This chapter is dedicated to introducing the fundamentals of the power grid, with a specific focus on the transmission power system. The state of the power system is described by power system variables that can be calculated using the power flow equations. The AC power flow equation is explained in detail, and its simplification and linearization using DC power flow equations are demonstrated.

The chapter also explores the purpose of OPF and its solutions, which are limited to economic dispatch problems since other tasks are beyond the scope of this thesis. The discussion includes both deterministic and stochastic OPF, with an emphasis on the chance-constrained approach for solving stochastic OPF. This approach can save costs by allowing violations of operational constraints with low probability, but it remains challenging to solve.

Additionally, the chapter outlines a process for generating the synthetic dataset required for the contribution part of this thesis. This dataset will aid in the conducting of the proposed contribution, and the chapter provides a brief explanation of its generation.

Overall, this chapter provides a thorough introduction to the fundamental concepts of power systems, with a particular emphasis on the transmission power system. It is a valuable resource for readers to develop a comprehensive understanding of this complex and vital field.

 \chapter{Gaussian Process}
\label{ch:3}
\section{Introduction} \label{ch3_intro}

Chapter~\ref{ch:3} serves to provide the reader with the fundamentals of machine learning, with a specific emphasis on supervised learning. The concept of supervised learning will be clearly explained, along with a thorough explanation of key terms within the field.

The second part of the chapter focuses on the Gaussian Process, as a model for solving regression problems within the context of supervised learning. The Gaussian Process model will be explored in detail, specifically within the domain of regression.

Overall, the chapter is divided into four sections, with section~\ref{ch3_ml} serving as an introduction to the main concepts of machine learning, emphasizing supervised learning. Section~\ref{ch3_gp} dives deeper into the specifics of the Gaussian Process model, while section~\ref{ch3_conclusion} provides a concise summary of the chapter's key points.

\section{Supervised Learning} \label{ch3_ml}

In this section, we will provide a brief overview of the fundamental concepts of machine learning (ML). Subsequently, we will delve into the task of supervised learning in machine learning and explore the associated terminology in greater depth. By providing a detailed explanation of these key concepts, we aim to facilitate a better understanding of supervised learning tasks.

\subsection{Fundamentals of machine learning}
Machine learning is a rapidly evolving field that has seen significant advancements in recent years. It is a field of artificial intelligence that focuses on developing algorithms and models that can learn from data and make predictions or decisions based on that learning. It has become increasingly important in recent years due to its ability to work with large amounts of data, so-called big data, and has the potential to revolutionize the way of solving problems and automating tasks. Therefore, one of the key advantages of machine learning is that it can handle large amounts of data much more efficiently than traditional methods. This is because machine learning algorithms can extract patterns and structure from the data, allowing them to make more accurate predictions and decisions. As a result, machine learning is increasingly being used in applications such as image and speech recognition, natural language processing, and fraud detection~\cite{vinoth2022fundamentals, zhou2021machine}.

There are three primary types of machine learning tasks:
\begin{enumerate}
    \item supervised learning;
    \item unsupervised learning;
    \item reinforcement learning.
\end{enumerate}
Each type of task is suited to different types of problems and data sets. The types of machine learning are visually illustrated in Fig.~\ref{fig:ml_fundamentals}.
\begin{figure}[H]
\centering
\vspace*{-10pt}
\includegraphics[width=1.\textwidth]{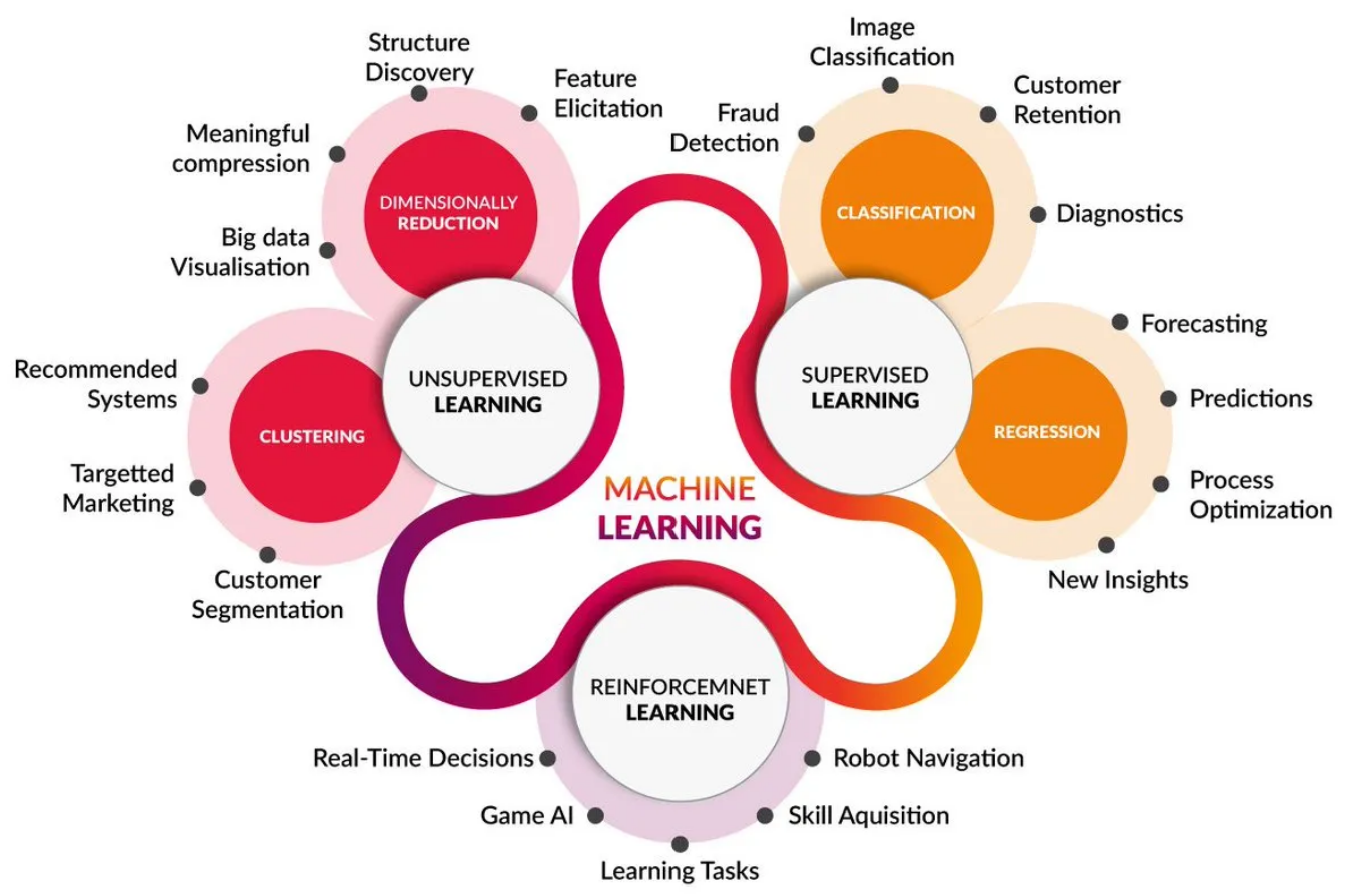}
\caption[Types of machine learning tasks; \newline the figure is taken from \href{https://medium.com/almabetter/machine-learning-fundamentals-for-beginners-70f409b1197e}{https://medium.com}]{Types of machine learning tasks; the figure is taken from  \href{https://medium.com/almabetter/machine-learning-fundamentals-for-beginners-70f409b1197e}{https://medium.com}.}
\label{fig:ml_fundamentals}
\end{figure}

Supervised learning is one of the most widely used types of machine learning tasks. It involves training a model on labeled data, where the desired output is already known. This allows the model to learn to predict the output for new input data. In classification tasks, the output is a label or category, while in regression tasks, the output is a continuous value. Supervised learning has many applications, including spam detection, medical diagnosis, weather forecasting, etc~\cite{cunningham2008supervised, hastie2009overview}.

Unsupervised learning is another important type of machine learning task. It involves training a model on unlabeled data, where the desired output is not known. The objective of unsupervised learning is to discover patterns or structures in the data. This can be useful for tasks such as clustering, where the goal is to group similar data points together, or for dimensionality reduction, where the goal is to reduce the number of features in the data~\cite{ghahramani2004unsupervised, usama2019unsupervised}.

Reinforcement learning is a type of machine learning that involves training a model to make decisions in an environment to maximize a reward signal. This type of learning is commonly used in robotics and game playing, where the model learns to take actions that will lead to a desired outcome based on feedback from the environment. Reinforcement learning is a powerful tool for creating intelligent agents that can interact with the world in a sophisticated way~\cite{kaelbling1996reinforcement, li2017deep}.

In addition to these primary machine learning tasks, there are also related concepts such as deep learning, semi-supervised learning, transfer learning, and adversarial learning. Deep learning is a subfield of machine learning that involves training models with multiple layers of neural networks~\cite{lecun2015deep}. This allows the models to learn complex representations of the input data, making them highly effective for tasks such as image and speech recognition. Deep learning has seen significant advancements in recent years, leading to breakthroughs in areas such as computer vision and natural language processing. Semi-supervised learning involves training a model on a combination of labeled and unlabeled data~\cite{van2020survey}, while transfer learning involves using a pre-trained model as a starting point for a new task~\cite{weiss2016survey}. Adversarial learning involves training models to be robust against attacks or adversarial inputs~\cite{lowd2005adversarial}.

Overall, machine learning is a powerful tool that has the potential to transform the way we solve problems and automate tasks. It requires expertise in mathematics, statistics, and computer science, but can lead to significant advancements in a wide range of fields. As data continues to become more abundant, the importance of machine learning is only expected to grow.

\subsection{Supervised learning}
Supervised learning is a widely used machine learning technique that involves learning the probabilistic relationship between input data $x^{in} \in \mathcal{X}$ and corresponding output data $y^{out} \in \mathcal{Y}$. This is accomplished through the use of a labeled training set $\{(x^{in}_{i}, y^{out}_{i}) | i = 1, ..., m_{s}\}$, where each sample is a pair consisting of an input observation and its associated output. The ultimate objective is to develop a prediction function $f$ that can accurately predict the output for new, previously unseen samples.

In this section, we will focus specifically on supervised learning for the regression task, which involves predicting a continuous output variable. It is important to note that supervised learning also includes the classification task, which involves predicting a discrete output variable. However, this topic will not be covered here. The main difference between these two frameworks lies in the nature of the output set $\mathcal{Y}$.

To train a supervised learning algorithm, a loss function $\ell_c$ is utilized to measure the degree of difference between the predicted output and the true output, also known as the label. The goal of the algorithm is to minimize the average error on the training set, which is referred to as the empirical error. This process of minimizing error is achieved through the principle of Empirical Risk Minimization (ERM)~\cite{burashnikova2022large}. The aim is to minimize empirical error, and thus reduce the generalization error. Generalization error implies an error of the prediction function made on new unseen samples. The relationship between empirical error and generalization error, aiming of finding a balance between a low empirical error and a high complexity of the prediction function by choosing the appropriate learning function, is a key area of study in statistical learning theory~\cite{vapnik1999nature}. This process of choosing the right learning function is known as the Structural Risk Minimization (SRM) principle~\cite{burashnikova2022large}.

\subsubsection{Fundamental assumptions}
In statistics, a crucial assumption is that all samples are independently and identically distributed (i.i.d.) from an unknown population distribution $\mathcal{D}$. This means that a sample set denoted as $\mathcal{s}$ and consisting of pairs of input samples $x^{in}$ and their corresponding outputs $y^{out}$, is generated randomly and without any dependence between samples. This assumption is essential for determining the representativeness of a training or test dataset, where the input samples and their corresponding outputs are generated from the same source.

Another essential notation in statistic learning is an error (risk or loss). In the machine learning community, this error is measured through the loss function. The loss function evaluates the error between prediction f($x^{in}$) and actual (desired) output $y^{out}$. Therefore, the loss function implies a distance over predicted and desired output and can be expressed in general form as:
\begin{equation*}
    \ell_c(f(x^{in}), y^{out})
\end{equation*}

\subsubsection{Generalization error and empirical error}

According to $\ell_c$, the error on all examples ($f(x^{in}), y^{out}$) can be defined. This error is called a generalization error and is given as:
\begin{equation}
\mathcal{L}(f) = \mathbb E_{ \mathcal D } \left[ \ell_c(x^{in}, y^{out}) \right]
\end{equation}

The prediction function $f$ is responsible for minimizing the error on new, unseen data, which is crucial for achieving good performance. However, it is impossible to directly estimate this error because the distribution of the population is unknown. To address this, we can optimize the average error on the training set by searching for the most suitable function $f$ in set of functions $\mathcal{F}$. By doing so, we can estimate the generalization error on $\mathcal{S}$, also called the empirical error, as:
\begin{equation}
 \label{eq:EmpError}
\hat{\mathcal{L}}(f,S) = \frac{1}{m_s} \sum_{i=1}^{m_s} \ell_c(f(x_i^{in}), y_i^{out}) 
\end{equation}

In regression problems, the choice of an appropriate loss function is crucial in accurately measuring the difference between predicted and actual values. The most commonly used loss functions for error measuring in regression problems are Mean Absolute Error (MAE), Mean Squared Error (MSE), Root Mean Squared Error (RMSE), and Mean Squared Logarithmic Error (MSLE).

\textbf{MAE} is a simple and robust loss function that measures the average absolute difference between the predicted and actual values. It is particularly useful in cases where outliers are present and variables do not have a strict Gaussian distribution. The MAE is defined as:
\begin{equation}
    \mathcal{L} = \frac{1}{m_s} \sum_{i=1}^{m_s} \|f(x_i^{in}) - y_i^{out} \|
\end{equation}

\textbf{MSE} measures the average squared difference between the predicted and actual values. It is sensitive to outliers and larger errors are penalized more heavily. Accordingly, MSE is suitable for problems where minimizing large errors is important. MSE is defined as:
\begin{equation}
    \mathcal{L} = \frac{1}{m_s} \sum_{i=1}^{m_s} (f(x_i^{in}) - y_i^{out})^{2}
\end{equation}

\textbf{RMSE} is the square root of MSE. It is more useful in practice since it is measured in the same units as the actual values. Therefore, the results are easier to interpret. RMSE is defined as:
\begin{equation}
    \mathcal{L} = \sqrt{\frac{1}{m_s} \sum_{i=1}^{m_s} (f(x_i^{in}) - y_i^{out})^{2}}
\end{equation}

\textbf{MSLE} measures the average squared logarithmic difference between the predicted and actual values. It is particularly beneficial in scenarios where the target variable has a wide range of values, and the objective is to minimize the relative error instead of the absolute error. MSLE is defined as:
\begin{equation}
    \mathcal{L} = \frac{1}{m_s} \sum_{i=1}^{m_s} (log(f(x_i^{in})) - log(y_i^{out}))^{2}
\end{equation}

\subsubsection{ERM principle}
To learn the prediction function $f$, we use the input and output data from the training dataset to find the best fit between them. The ERM principle suggests that if the training samples from set $S$ are representative of the distribution $D$, then the empirical error is a reliable estimate of the generalization error. By minimizing the empirical error on a given training set, we can also minimize the generalization error. In summary, the ERM principle returns as a way to optimize the function $f$ by minimizing the empirical error on the training set as follows:
\begin{equation}
    f_S = arg~\min_{f \in \mathcal{F}}~ \frac{1}{m_s} \sum_{i=1}^{m_s} \ell_c(f(x_i^{in}), y_i^{out}) 
\end{equation}

\subsubsection{Learning algorithm complexity}
In supervised learning, the ability of a learning algorithm to generalize to new, unseen data is reflected in the complexity of the learned function. To evaluate function complexity, we compare the empirical error (the error on the training data) and the generalization error (the error on the unseen data). If the empirical error is small, but the generalization error is high, this indicates \textit{overfitting}, meaning that the learned function is too complex for the training data. To avoid \textit{overfitting}, simpler functions should be learned. Apart from identifying an appropriate learning function $f$, overfitting can be prevented by implementing various techniques such as early stopping, pruning, regularization, etc.

However, if the learning function is too simple, it may not have sufficient expressiveness to capture the underlying patterns in the data, leading to \textit{underfitting}. In this case, both the empirical and generalization errors will be high. Therefore, it is important to strike a balance between function complexity and performance. This trade-off between low empirical error and complex function is known as the \textit{bias-variance} trade-off. Achieving a balance between function complexity and generalization is crucial to prevent \textit{overfitting} or \textit{underfitting} in machine learning. This involves finding the right level of learning function complexity that can effectively capture the underlying patterns in the data, while also being simple enough to avoid \textit{overfitting}. In Fig.~\ref{fig:ERM_property}, all properties of ERM (\textit{overfitting}, \textit{underfitting}, and \textit{bias-variance} trade-off) are illustrated with a toy example for a regression problem.
\begin{figure}[H]
\centering
\vspace*{-10pt}
\includegraphics[width=1.\textwidth]{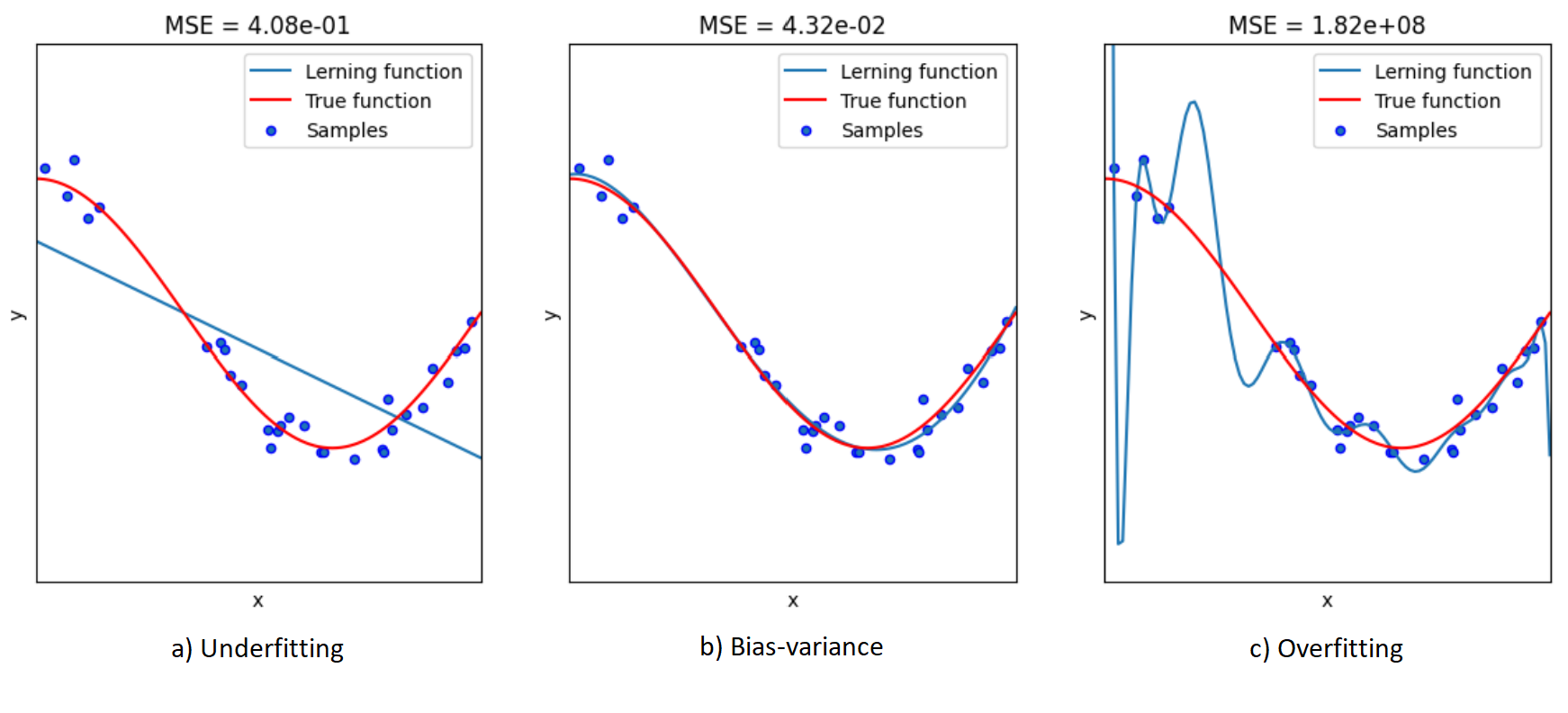}
\caption[ERM properties]{Illustration of the ERM properties on a regression problem using a toy function.}
\label{fig:ERM_property}
\end{figure}

\subsubsection{SRM principle}
The SRM principle aims to provide a bias-variance trade-off in supervised learning by finding the most suitable learning function $f$ from set $\mathcal{F}$. However, selecting an optimal learning function can be a challenging task. To simplify this process, one can use a simple strategy that involves applying the ERM algorithm to find a function and then calculating the bound on the generalization error. The learning function that minimizes this bound provides the best theoretical guarantee on the generalization error compared to other function candidates. Therefore, this function can be selected as the appropriate learning function $f$. By following this approach, one can effectively minimize the risk of overfitting and improve the generalization performance of a supervised learning model.
\section{Gaussian Process}\label{ch3_gp}

In recent years, deep neural networks have gained popularity in the supervised machine learning field, while the Gaussian Process remains a widely used framework in probabilistic machine learning. The key difference between these models lies in how they make predictions. Neural networks optimize weights to fit functions and return deterministic outputs, whereas the Gaussian process uses kernel functions to interpolate training data and returns probabilistic outputs distributed as a Gaussian. As a result, the Gaussian Process is a nonparametric model that can provide measures of uncertainty for each point prediction.

A comprehensive introduction to the Gaussian Process is given in~\cite{rasmussen2006gaussian}. The research has shown that GP can be successfully established to propagate uncertainty in multi-step predictions, as demonstrated in~\cite{girard2002gaussian}. Consequently, the Gaussian Process has been effectively used as a predictive model in model predictive control~\cite{murray2003adaptive, kocijan2004gaussian}. However, this thesis focuses on one-step prediction by propagating input uncertainties to the predicted output. The idea of using the Gaussian Process in the stochastic OPF problem was inspired by~\cite{girard2002gaussian} and~\cite{hewing2019cautious}, who found that cautious control and uncertainty propagation in model predictive control produced effective results.

\subsection{Gaussian Process Framework}

A Gaussian Process is a statistical model based on Bayesian inference that defines a distribution over functions. In a Gaussian process, every point in a function is modeled as a random variable, and the joint distribution of all these variables is a multivariate Gaussian distribution.
The GP function $f(x^{in})$ can be represented as an infinite vector, where each input $x^{in}$ in the vector corresponds to the function value at a specific input. However, working with an infinite object is not practical. The advantage of a Gaussian process is that it allows us to perform inference using a finite subset of points and still obtain the same results as we would with the entire infinite vector. According to~\cite{rasmussen2006gaussian}, the Gaussian Process is a collection of random variables, any finite number of which have a joint Gaussian distribution.

For the given input $x^{in}$, the output $y^{out}$ is given by:
\begin{equation}\label{eq:output}
    y^{out} = f(x^{in}) + \zeta
\end{equation}
where $\zeta \sim \mathcal{N}(0, \sigma^{2}_{\zeta})$ is Gaussian measurement noise and $f: \mathcal{R}^{\mathbb{D}} \to \mathcal{R}$. The GP function $f(x^{in})$ is defined by a mean function $\mu(x_{p}^{in})$ and a covariance function $k(x_{p}^{in}, x_{q}^{in})$:
\begin{subequations}\label{eq:gp_mean_cov}
\begin{align}
 \mu(x_{p}^{in}) = \mathbb{E}[f(x_{p}^{in})] ~~~~~~~~~~~~~~~~~~~~~~~~~\\
 k(x_{p}^{in}, x_{q}^{in}) = \mathbb{E}[(f(x_{p}^{in}) - \mu(x_{p}^{in}))(f(x_{q}^{in}) - \mu(x_{q}^{in}))]
\end{align}
\end{subequations}
where the mean function represents the expected value of the function at any point, while the covariance function measures the correlation between the values of the function at different points. Accordingly, the GP function can be defined as:
\begin{equation}
    f(x^{in}) = GP(\mu(x_{p}^{in}), k(x_{p}^{in}, x_{q}^{in}))
\end{equation}

\subsection{Bayesian inference}
The Gaussian Process is a statistical model that relies on Bayesian inference. According to~\cite{gelman1995bayesian}, Bayesian inference involves building a probability model based on a given set of training data, where a set of training data is defined as:
\begin{equation*}
    D = \{(x^{in}_i, y^{out}_i),~~i=1,...,m_s\}
\end{equation*}
In Bayesian inference, two terms are used: prior inference and posterior inference. Prior inference implies inference without data, while posterior inference is inference with data.

\subsubsection{Prior inference}
The Gaussian process uses training data to interpolate between them via kernel function. However, the quality of the interpolation depends on good prior inference, which involves guessing the function without relying solely on the available training data. This is particularly important for regions of the function where no training data are available.

To make an accurate prior inference, it is important to define the mean and covariance functions from~\ref{eq:gp_mean_cov}. The most commonly used prior mean function is zero, but if the behavior of the physical system is known, choosing a different type of prior mean function may be more appropriate (such as a linear function). However, the choice of covariance function, which takes into account the features of the data, is the most critical aspect of prior inference. There are many different types of covariance functions~\cite{rasmussen2006gaussian}. Moreover, combining these functions can often result in a better fit for the model. The most popular and used covariance functions, depending on the type of problem, are:  
\begin{enumerate}
    \item Squared exponential (SE): $k(x_{p}^{in}, x_{q}^{in}) = exp(-\frac{1}{2}\frac{|x_{p}^{in} - x_{q}^{in}|^2}{l^2})$
    \item Matern: $k(x_{p}^{in}, x_{q}^{in}) = \frac{2^{1-\nu}}{\Gamma(\nu)} (\frac{\sqrt{2\nu}|x_{p}^{in} - x_{q}^{in}|}{l})^{\nu} K_{\nu} (\frac{\sqrt{2\nu}|x_{p}^{in} - x_{q}^{in}|}{l}) $
    \item Rational quadratic (RQ): $(1+|x_{p}^{in} - x_{q}^{in}|)^{-\iota}$
    \item Periodic: $exp(-\frac{2sin^2(\frac{(x_{p}^{in} - x_{q}^{in})}{2})}{l^2})$
\end{enumerate}
where $\nu$, $l$, $\iota$ are hyperparameters; $K_{\nu}(\cdot)$ is a modified Bessel function; $\Gamma(\cdot)$ is the gamma function. Enumerated covariance functions are illustrated in Fig.~\ref{fig:kernel_func}. 

\begin{figure}[H]
\centering
\vspace*{-10pt}
\includegraphics[width=0.78\textwidth]{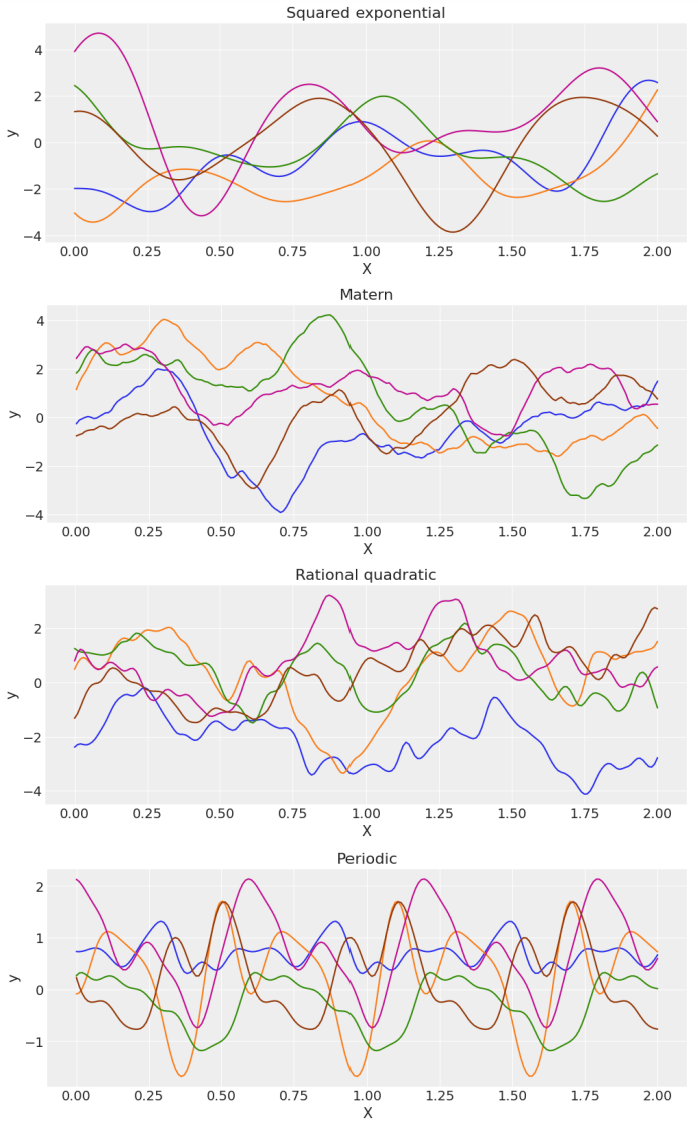}
\caption[Covariance functions]{Illustration of the most popular covariance functions.}
\label{fig:kernel_func}
\end{figure}

In this thesis, we will utilize a squared exponential covariance function. This function is well-suited for our problem due to its inherent smoothness, which makes it capable of effectively fitting non-linear functions while avoiding superfluous local minima. Specifically, we will be employing a variant of the squared exponential function known as Squared Exponential Automatic Relevance Determination (SEard). In SEard, a unique length scale $l$ hyperparameter is assigned for each input, resulting in a more specialized and effective model. SEard covariance function is given as:
\begin{equation}
    k(x_{p}^{in}, x_{q}^{in}) = \sigma^2_f exp(-\frac{1}{2}(x_{p}^{in} - x_{q}^{in})\Lambda^{-1} (x_{p}^{in} - x_{q}^{in}))
\end{equation}
where $\sigma^2_f$ is a signal variance and $\Lambda = diag([l^2_1,l^2_2,...,l^2_{\mathbb{D}}])$ is diagonal matrix of length-scales.

According to the SE function, covariance $k$ is close to one when input points are close indicating that the points are similar. On the other hand, as the distance between the points increases, the covariance decreases exponentially, indicating that the points are not similar.

\subsubsection{Posterior inference}

As previously mentioned, prior inference evaluates the model without data, based on hyperparameters. While this is an important part of GP, it has limitations compared to the posterior part, which takes data into account. To address the differences between the prior and posterior, as well as to insight into how the Gaussian Process Regression (GPR) model works, we will consider a toy sine function. Assuming a noise signal $y^{out}$, then the SE function can be expressed as:
\begin{equation}
    k(x_{p}^{in}, x_{q}^{in}) = \sigma^2_f exp(-\frac{1}{2}(x_{p}^{in} - x_{q}^{in})\Lambda^{-1} (x_{p}^{in} - x_{q}^{in})) + \sigma^2_{\zeta}\delta_{pq}
\end{equation}
where $\sigma^2_{\zeta}$ is noise variance that provides avoiding over-fitting the model on the noise; $\delta_{pq}$ is an element of the identity matrix $I$. Accordingly, the vector $\Theta=[\Lambda, \sigma^{2}_f, \sigma^{2}_n]$ consists of hyperparameters that need to be optimized during GPR training. Here we will illustrate with a toy sine function how each of these hyperparameters affects model fit.

In general, for the given input data matrix $X^{in} = [x^{in}_i]^T_{1\leq i\leq m_s} \in \mathbb{R}^{m_s \times \mathcal{D}}$, each output dimension $y^{out} \in \mathbb{R}^{m_s}$ is learnt independently. A GP with prior mean function $\mu(\cdot)$ and chosen kernel $k(\cdot,\cdot)$ in each output dimension, models the output $y^{out}$ as normally distributed data:
\begin{equation}\label{eq:output_dist_1} 
 y^{out} \sim \mathcal{N}(\mu(X^{in}), K + \sigma^2_{\zeta}I )
\end{equation}
where $\sigma^2_{\zeta}$ is the variance of measurement noise $\zeta$ in \eqref{eq:output}; identity matrix $I$ is $I = diag(\delta_{pq})$; $\mu(X^{in}) = [\mu(x^{in}_1),...,\mu(x^{in}_{m_s})]^T$, and $K_{pq} = k(x^{in}_{pg},x^{in}_{pg})$ is the Gram matrix of the sampled data points. This work is based on the common zero prior mean $\mu(x^{in})=0$. 

In this comparison, we will analyze the prior, with zero mean and variance $\sigma^{2}_{f} = 0.5$, and the posterior distribution of the sine curve. The prior distribution indicates that the sine curve will usually be within 95\% (percentile) of the zero mean, as illustrated by the 10 random functions generated from the prior distribution in Fig~\ref{fig:prior}. Fig~\ref{fig:prior} depicts a zero mean function as a red dashed line, while 95\% of the signal variance is represented by the red area. 
\begin{figure}[H]
\centering
\includegraphics[width=0.78\textwidth]{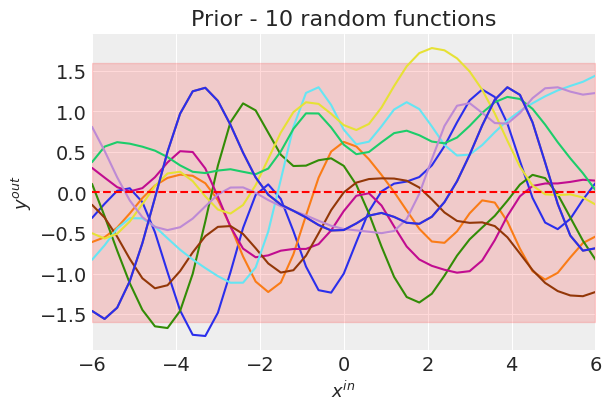}
\caption[Prior distribution]{Illustration of 10 random functions with the prior distribution.}
\label{fig:prior}
\end{figure}
However, when training points are available, the posterior distribution changes. Fig.~\ref{fig:poster} shows the same 10 random functions, but with the posterior distribution after 8 training points have been added. We can see that the functions are closer to the training points, indicating that if a new unseen point is close to the training points, it is highly probable that it will be close to the spline between the known points. Nevertheless, if the unseen point is too far from the training points, the posterior distribution will revert to the prior distribution. This implies that the prior distribution still holds for points that are far from the training data. 

To verify the aforementioned claim, Fig.~\ref{fig:posteriror_test} illustrates examples of the sine curve with and without some training points. It is evident that uncertainty increases when the training data points are widely spaced, and the predicted distribution approaches the prior distribution. In other words, the prior distribution becomes more significant when we have fewer data or data that is not well distributed in the space, and the predicted distribution becomes less confident.
\begin{figure}[H]
\centering
\vspace*{-10pt}
\includegraphics[width=0.78\textwidth]{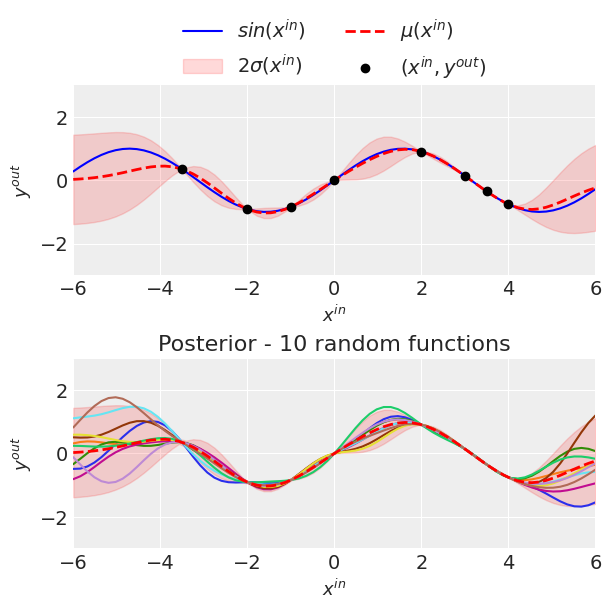}
\caption[Posterior distribution]{Illustration of 10 random functions with the posterior distribution distributions based on 8 training data.}
\label{fig:poster}
\end{figure}

In this context, we can examine how different hyperparameters impact the GP function. One hyperparameter of interest is the noise variance $\sigma^{2}_{\zeta}$, which can disrupt the signal. To better understand how noise variance affects GP prediction, in Fig.~\ref{fig:posteriror_test_2} we visualize two scenarios with noise variance $\sigma^{2}_{\zeta} = 0.2$ and zero. In the absence of noise, the GP function can exactly fit all the training data points, and the posterior variance at these points is zero. However, when we consider the possibility of noise, we become uncertain about the true location of the signal, and this uncertainty leads to higher posterior variance at the training points. The GP mean, which represents the best estimate of the underlying function, tries to fit the data points while accounting for this uncertainty. In the presence of noise, the GP mean may not pass through all the training points, trying to find the simplest path that fits the data points while accounting for the noise.

Unlike noise variance, signal variance $\sigma^{2}_{f}$ is a key factor in predicting the certainty of a model's output, as it is independent of the training data and reflects the underlying variability of the signal being modeled. While it does not affect the mean prediction, it does have a significant impact on the variance of the predictions. When the signal variance is low, the model's uncertainty is naive when predicting data far from training data, which can result in overly optimistic predictions. Conversely, when the signal variance is high, the model's uncertainty is more conservative and it becomes more difficult to determine the underlying signal from the noise. This is illustrated in Fig.~\ref{fig:posteriror_test_3} where we compare signal variances of 0.5 and 3.  The appropriate signal variance will depend on the specific application and the level of uncertainty precision required in the model's predictions.

\begin{figure}[H]
\centering
\includegraphics[width=0.78\textwidth]{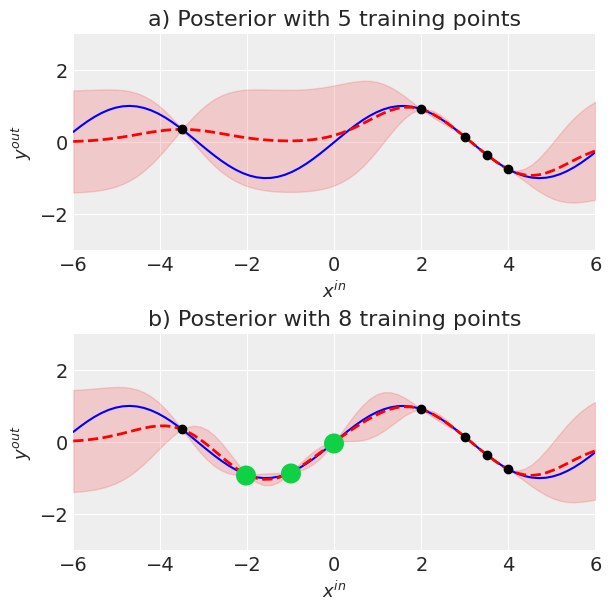}
\caption[Comperation of posterior distributions]{Comparing the posterior distribution with different numbers of training points. a) 5 training points are used to fit GP; b) 3 new training points are added (green dots).}
\label{fig:posteriror_test}
\end{figure}

The length-scale $l^{2}$ is a crucial hyperparameter in GP modeling. It determines the shape of the spline that links the training points in the GP function. A low length-scale will result in a steep and aggressive function, while a higher length-scale will produce a smoother one. Therefore, the length scale affects the flexibility of the GP function, and choosing the appropriate value is crucial for achieving good performance in finding new data. The impact of the length-scale hyperparameter is demonstrated in Fig.~\ref{fig:posteriror_test_4} for the $l^{2} = 1$ and $l^{2} = 0.2$.

\begin{figure}[H]
\centering
\vspace*{-10pt}
\includegraphics[width=1.\textwidth]{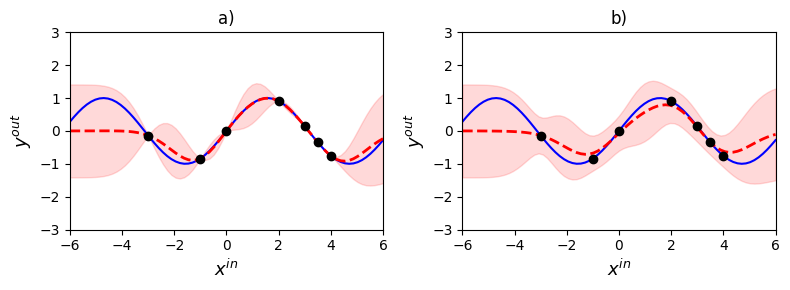}
\caption[Posterior distribution - adjusting noise variance ]{Posterior distribution obtained by adjusting noise variance $\sigma^2_{\zeta}$. a) signal with $\sigma^2_{\zeta} = 0$; b) signal with $\sigma^2_{\zeta} = 0.2$.}
\label{fig:posteriror_test_2}
\end{figure}

\begin{figure}[H]
\centering
\vspace*{-10pt}
\includegraphics[width=1.\textwidth]{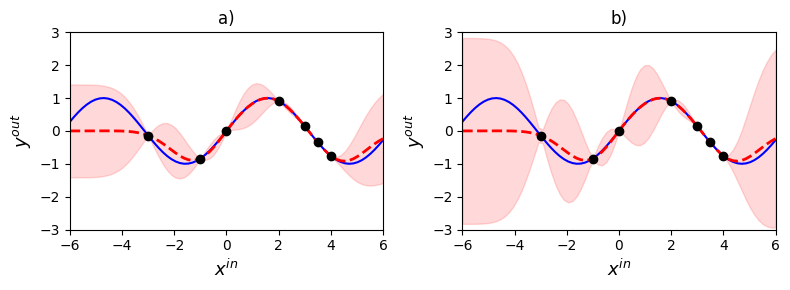}
\caption[Posterior distribution - adjusting signal variance]{Posterior distribution obtained by adjusting signal variance $\sigma^2_{f}$. a) signal with $\sigma^2_{f} = 0.5$; b) signal with $\sigma^2_{f} = 3$.}
\label{fig:posteriror_test_3}
\end{figure}

In this exploration, we investigate how the distribution and spread of training data in space can impact the fitting of the GP function. The ideal distribution of training data points, where the sin function is perfectly fitted by the GP function, is demonstrated in Fig.~\ref{fig:posteriror_test_4} a). However, this ideal distribution is rare or even impossible in practice. To ensure a good spread of data, a Latin hypercube design can be used to distribute the training data, as shown in Fig.~\ref{fig:posteriror_test_4} b). Fig.~\ref{fig:posteriror_test_4} c) displays the worst-case distribution where the GP function fails to fit the sin function at all. In practice, it is common to encounter examples where the distribution fits some parts of the function well while leaving other parts equal to the prior, as illustrated in Fig.~\ref{fig:posteriror_test_4} d). Therefore, achieving a high-quality distribution of training data points is one of the crucial factors in obtaining a good GP model.

\begin{figure}[H]
\centering
\vspace*{-10pt}
\includegraphics[width=1.\textwidth]{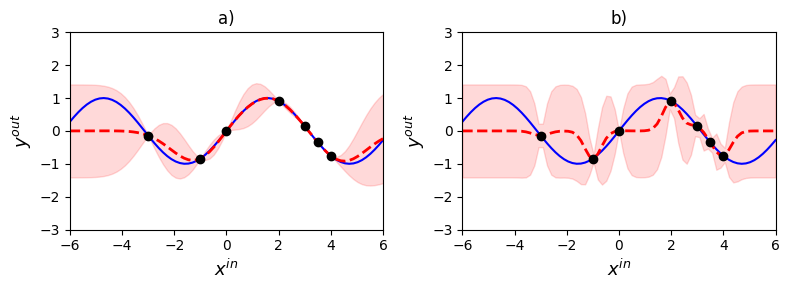}
\caption[Posterior distribution - adjusting length-scales]{Posterior distribution obtained by adjusting length-scales $l^2$. a) signal with $l^2 = 1$; b) signal with $l^2 = 0.2$.}
\label{fig:posteriror_test_4}
\end{figure}

\begin{figure}[H]
\centering
\vspace*{-10pt}
\includegraphics[width=1.\textwidth]{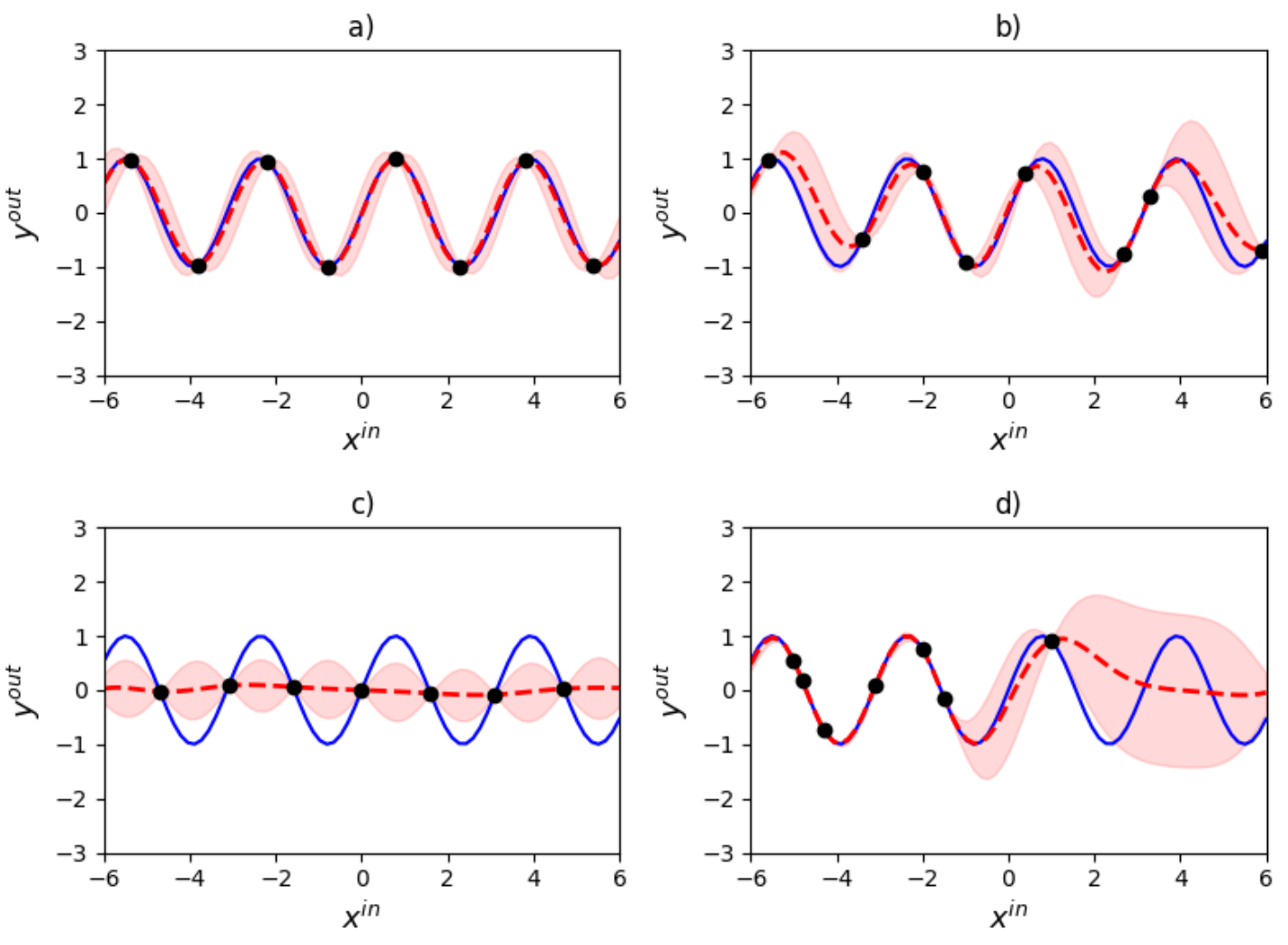}
\caption[Illustrations of the GP results using different sampling distributions]{Illustrations of the GP results using different sampling distributions with the same hyperparameters. All cases have 10 points in the distribution. a) data points are distributed manually to cover the minimum/maximum of input and output data; b) data points are distributed using the Latin hypercube design; c) data points are uniformly distributed with the worst case distribution; d) data points are uniformly distributed with a random distribution.}
\label{fig:dif_distributions}
\end{figure}

\subsection{Optimization of hyperparameters}
In the previous section, it was demonstrated that hyperparameters $\Theta = [\Lambda, \sigma^2_{f}, \sigma^2_{\zeta}]$ play a crucial role in determining the behavior of a GP function. However, it is not feasible to manually set hyperparameter values during the training process. Therefore, it is essential to optimize hyperparameters automatically. To achieve this, we use the Maximum Likelihood Estimate (MLE) method with the log marginal likelihood. In this method, we treat the hyperparameters as unknown parameters and try to find their optimal values by maximizing the likelihood function. To simplify the optimization process, we assume that the length-scales of the GP function are independent of each other. This means that the hyperparameter matrix, denoted by $\Lambda$, is a diagonal matrix. 

The likelihood function is the probability of the observed data given the hyperparameters. Following~\cite{rasmussen2006gaussian}, the MLE with log marginal likelihood is given as:
\begin{equation}\label{eq:mle}
 \begin{split}
 \log~p(y^{out}|X^{in}, \Theta) = \int p(y^{out}|\mathrm{w}, X^{in}, \Theta)\mathrm{d}\mathrm{w} ~~~~~~~~~~~~~~~~~~~~~~~ \\ = \frac{1}{2}(y^{out} - \mu(X^{in}))^{T}(K + \sigma^2_{\zeta}I)(y^{out} - \mu(X^{in})) \\ - \frac{1}{2}\log|K + \sigma^2_{\zeta}I| - \frac{N}{2}\log 2 \pi
 \end{split}
\end{equation}
where $\mathrm{w}$ are the weights of all parameters, and $\mu(X^{in})$ represents a prior mean value. Since we are considering a GP with zero prior mean, Eq.~\ref{eq:mle} can be simplified as:
\begin{equation}\label{eq:mle_2}
 \log~p(y^{out}|X^{in}, \Theta) = \frac{1}{2}(y^{out})^{T}(K + \sigma^2_{\zeta}I)y^{out} - \frac{1}{2}\log|K + \sigma^2_{\zeta}I| - \frac{m_s}{2}\log 2 \pi
\end{equation}
where each term of the Eq.~\ref{eq:mle_2} has interpretative roles. For example, the first term $\frac{1}{2}(y^{out})^{T}(K + \sigma^2_{\zeta}I) y^{out}$ represents the data-fit term, which measures how well the GP function fits the observed data using only the available samples. This term decreases with the length-scales of the covariance function, which means that the GP function becomes less flexible. The second term $\frac{1}{2}\log|K + \sigma^2_{\zeta}I|$ represents the complexity term, which penalizes overly complex models and encourages simplicity. This term increases as the length-scales of the covariance function increase, which means that the GP function becomes less complex. Finally, the last term $\frac{m_s}{2}\log 2 \pi$ defines a normalization constant.  

By maximizing the log marginal likelihood, we can find the optimal hyperparameters for the GP function. However, by using the negative log marginal likelihood as the objective function, any non-linear solver can be utilized. Therefore, the objective function will be minimized according to hyperparameters. The negative log marginal likelihood is a non-convex function, meaning that it may have multiple local minima. To overcome this, it is common practice to use a multi-start process with different initial guesses for the hyperparameters. This involves running the optimization algorithm multiple times with different initial values for the hyperparameters and selecting the solution with the lowest negative log marginal likelihood as the optimal solution. Accordingly, the optimal hyperparameters are found as:
\begin{equation}\label{eq:mle_3}
\Theta^{*} = argmin(-\log~p(y^{out}|X^{in}, \Theta))
\end{equation}
By optimizing the hyperparameters using the negative log marginal likelihood, we can ensure that the GP function fits the observed data well and achieves good generalization performance on new data. However, the optimization process can be computationally expensive, especially for large datasets. Therefore, it is important to use efficient optimization algorithms. In this thesis, we utilize the Sequential Least Squares Programming (SLSQP)~\cite{kraft1988software} solver to determine the optimal solution.

\subsection{Prediction with deterministic inputs}
Predictive distribution of noise-free model $y^{out}_{*}$ for a given unseen sample data $x^{in}_*$ can be represent as marginal distribution with zero mean:
\begin{equation}
 y^{out}_{*} \sim \mathcal{N}(0, K(x^{in}_{*}, x^{in}_{*}))
\end{equation}

By combining the marginal distribution with the distribution of the GP using the training samples, we obtain a joint distribution:
\begin{equation}\label{eq:joint_dist}
 p(y^{out}_{*}, y^{out}) \sim \mathcal{N}\left(\begin{bmatrix}
 0\\
 0
 \end{bmatrix}, 
 \begin{bmatrix}
 K(X^{in}, X^{in}) & k(X^{in}, x^{in}_*)\\
 k(x^{in}_*, X^{in}) & k(x^{in}_*, x^{in}_*)
 \end{bmatrix}\right)
\end{equation}
In the case of the nosy signal, the joint distribution can be generalized:
\begin{equation}\label{eq:joint_dist_1}
 p(y^{out}_{*} | y^{out}) \sim \mathcal{N}\left(\begin{bmatrix}
 0\\
 0
 \end{bmatrix}, 
 \begin{bmatrix}
 K(X^{in}, X^{in}) + \sigma_{\zeta}^{2}I & k(X^{in}, x^{in}_*)\\
 k(x^{in}_*, X^{in}) & k(x^{in}_*, x^{in}_*)
 \end{bmatrix}\right)
\end{equation}

Since the joint distribution of the variables is Gaussian, the resulting distribution is conditioned on the training
data $p(y^{out}_{*}|y^{out}) = \mathcal{N}(\mu(x^{in}_*), \sigma^2(x^{in}_*))$ is also Gaussian \cite{rasmussen2003gaussian} with:
\begin{subequations}\label{eq:gp_output_1}
\begin{align}
 \mu(x^{in}_*) = k^T_*(K + \sigma^2_{\zeta}I)^{-1}y^{out} \label{eq:gp_00} \\
 \sigma^2(x^{in}_*) = k(x^{in}_*, x^{in}_*) - k^T_*(K + \sigma^2_{\zeta}I)^{-1}k_* \label{eq:gp_01}
\end{align}
\end{subequations}
where $k_* = k(X^{in}, x^{in}_*) $ is the vector of covariance functions and $K(X^{in}, X^{in})$ is a Gram matrix. 

In general, the predictive distribution of multivariate output dimension $y^{out}_{*}$ can be written as follows:
\begin{subequations}\label{eq:gp_output_2}
\begin{align}
 \mu(x^{in}_*) = [\mu_1(x^{in}_*), \mu_2(x^{in}_*), ..., \mu_{n_y}(x^{in}_*)]^T \\
 \Sigma(x^{in}_*) = diag([\sigma^2_1(x^{in}_*), \sigma^2_2(x^{in}_*), ..., \sigma^2_{n_y}(x^{in}_*)])
\end{align}
\end{subequations}

\subsection{Prediction with uncertain inputs}\label{sec:uncer_pred}
Up until now, we have assumed that the input to the GP is deterministic, while the output is Gaussian with errors caused by measurement or modeling inaccuracies. However, if we are dealing with random input, where the uncertainties in the input need to be taken into account when propagating through the GP model, we can assume that the input follows a Gaussian distribution, denoted as $x^{in}_* = \mathcal{N}(\mu_{x^{in}_*}, \Sigma_{x^{in}_*})$. To obtain the distribution of the predicted output, which is also uncertain, we need to integrate over the input distribution using the following approach:
\begin{equation}\label{eq:input_dist}
 p(y^{out}_* | \mu_{x^{in}_*}, \Sigma_{x^{in}_*}) = \int p(y^{out}_*|x^{in}_*)\, p(x^{in}_*| \mu_{x^{in}_*}, \Sigma_{x^{in}_*}) \,dx^{in}_*
\end{equation}
In the case of the general kernel functions, it is not possible to compute the posterior distribution analytically because the Gaussian input is mapped through a non-linear function. To address this issue, the predictive distribution can be approximated using a Gaussian distribution and utilize various methods to calculate its statistics, allowing propagation of the uncertainty through the model. In our context, we consider two techniques to propagate input uncertainties: Taylor Approximation (TA) and Exact Moment Matching (EM). These techniques are used to approximate complex functions and probability distributions, respectively. For a more comprehensive understanding of these techniques, we suggest referring to the works of Girard and Deisenroth \cite{girard2003gaussian, deisenroth2010efficient}.

\subsubsection{Taylor Approximation}
By using TA approximation, the mean output $\mu(x^{in}_{*})$ is approximated by its first-order Taylor expansion. For the output variance $\sigma^2(x^{in}_{*})$, we use the first and second-order Taylor expansions around the mean output $\mu^{in}_{x_*}$ to approximate it.

When using a first-order Taylor expansion, the mean prediction at a random input $x^{in}_{*}$ does not offer any correction beyond what is already provided by the zero-order, as in \eqref{eq:gp_00}:
\begin{equation}\label{eq:mean_TA1}
 \mu_{TA}(x^{in}_*) = k^T_*(\mu_{x^{in}_*})(K + \sigma^2_{\zeta} I)^{-1}y^{out}
\end{equation}

The first-order Taylor Approximation (TA1) of the output variance is given as follows:
\begin{equation}\label{sigma_TA1}
 \sigma^2_{TA1}(x^{in}_*) = \sigma^2(\mu_{x^{in}_*}) + \frac{\partial \mu_{TA}(\mu_{x^{in}_*})}{\partial \mu_{x^{in}_*}}^T \Sigma_{x^{in}_*} \frac{\partial \mu_{TA}(\mu_{x^{in}_*})}{\partial \mu_{x^{in}_*}} 
\end{equation}
as well as the second-order Taylor Approximation (TA2):
\begin{equation}\label{sigma_TA2}
 \sigma^2_{TA2}(x^{in}_*) = \sigma^2_{TA1}(x^{in}_*) + \frac{1}{2}tr\left(\frac{\partial^2 \sigma^2(\mu_{x^{in}_*})}{\partial \mu_{x^{in}_*} \partial\mu_{x^{in}_*}^T}\Sigma_{x^{in}_*}\right)
\end{equation}
where $\sigma^2(\mu_{x^{in}_*})$ corresponds to~\ref{eq:gp_01}. In contrast to the deterministic input case in~\ref{eq:gp_output_1}, using the first and second-order Taylor expansion leads to correction terms for the posterior variance and covariance. These correction terms consider both the gradient of the posterior mean and the variance of the input of the GP. As a result, according to~\cite{hewing2019cautious}, the complexity of prediction with TA1 is $O(n_y n^2_x m_s^2)$. However, TA2 approximation requires additional computational effort as it involves solving for the second derivative.

\subsubsection{Exact Moment Matching}
For a zero prior mean function and SEard kernel, the predicted mean and variance can be computed analytically using the law of iterated expectations (Fubini's theorem)~\cite{deisenroth2010efficient}. Therefore, the predicted mean can be expressed as:
\begin{equation}\label{eq:mean_EM}
 \mu_{EM}(x^{in}_*) = k^T_*(\mu_{x^{in}_*}, \Sigma_{x^{in}_*})(K + \sigma^2_n I)^{-1}y
\end{equation}
where the covariance function $k_*$ is simultaneously a function of input mean and covariance:
\begin{equation}\label{eq:se_kernel_EM}
 \begin{split}
 k^T_*(\mu_{x^{in}_*}, \Sigma_{x^{in}_*}) = \sigma_f^{2}|\Sigma_{x^{in}_*}\Lambda^{-1} + I|^{-\frac{1}{2}} \cdot ~~~~~~~~~~~~~~~~~~\\
 ~~~~~~~~exp(-\frac{1}{2}(x^{in}-\mu_{x^{in}_*})^T(\Sigma_{x^{in}_*} + \Lambda)^{-1}(x^{in}-\mu_{x^{in}_*})) 
 \end{split}
\end{equation}

The variance of the predictive distribution is defined as follows:
\begin{equation}\label{eq:sigma_EM}
 \sigma^2_{EM}(x^{in}_*) = \sigma_f^2 - tr((K + \sigma^2_{\zeta} I)^{-1}Q) + \beta^TQ\beta - \mu_{x^{in}_*}^2
\end{equation}
in which $\beta = (K + \sigma^2_{\zeta} I)^{-1}y^{out}$ and elements of $Q \in \mathbb{R}^{m_s\mathsf{x}m_s}$ are computed as:
\begin{equation}\label{eq:q_EM}
 \begin{split}
 [Q]_{ij} = \frac{k_*(x^{in}_i,\mu_{x^{in}_*}) k_*(x^{in}_j,\mu_{x^{in}_*})}{|2\Sigma_{x^{in}_*}\Lambda^{-1} + I|^{\frac{1}{2}}} \cdot ~~~~~~~~~~~~~\\exp((z_{ij}-\mu_{x^{in}_*})^T(\Sigma_{x^{in}_*}+\frac{1}{2}\Lambda)^{-1}\Sigma_{x^{in}_*}\Lambda^{-1}(z_{ij} - \mu_{x^{in}_*}))
 \end{split}
\end{equation}
where $z_{ij} = \frac{1}{2}(x^{in}_i + x^{in}_j)$. According to to~\cite{hewing2019cautious}, the computational complexity in prediction with EM approximation is given as $O(n_y^2 n_x m_s^2)$

GP is limited in its applicability for relatively small systems or slow systems in dynamic contexts due to the fact that their computational complexity is directly influenced by the input and output dimensions, as well as the number of training points. This means that as these dimensions and the number of training points increase, the computational complexity also increases, which can make GP computationally difficult for larger systems. Therefore, the use of GP may be limited in such scenarios.

\subsection{Computational complexity}

As emphasized in the previous subsection, GP is a limited model in the context of scalability. Specifically, as the size of the system and the number of samples increases, the computational complexity of the GP model also increases. One of the primary reasons for this computational complexity is the computation of the inverse covariance matrix $(K + \sigma^2_{\zeta}I)^{-1}$. This matrix is an essential component of GP and its calculation becomes increasingly difficult as the size of the data set grows. This can make the use of GP challenging and computationally expensive in large-scale systems.

To minimize the computational effort in inverting $(K + \sigma^2_{\zeta}I)^{-1}$ during prediction, it is possible to apply the Cholesky decomposition. Cholesky decomposition decompose matrix into $(K + \sigma^2_{\zeta}I)^{-1} = L_{ch}L_{ch}^T$, where $L_{ch}$ is the lower triangular matrix with all positive elements. Reducing the computational complexity is possible by pre-computing both the Cholesky $L_{ch}$ and the linear $\beta_{ch}$ terms as:
\begin{subequations}\label{eq:chol_1}
\begin{align}
 L_{ch} = chol(K + \sigma^2_{\zeta}I) \\
 \beta_{ch} = L_{ch}^T \setminus (L_{ch} \setminus y^{out})
\end{align}
\end{subequations}
where $\setminus$ is the left matrix division.

Hence the mean and variance can be calculated as:
\begin{subequations}\label{eq:chol_2}
\begin{align}
 \mu(x^{in}_*) = k^T_* \beta_{ch}\\
 \sigma^2(x^{in}_*) = k(x^{in}_*,x^{in}_*) - \tau_{ch}^T\tau_{ch}
\end{align}
\end{subequations}
where $\tau_{ch} = L_{ch}\setminus k_*$.

\subsection{Sparse approximation}
To handle the increasing computational complexity that can arise from large training data, one approach is to use inducing points to create a sparse approximation of the GP. This approach can significantly reduce the computational cost while still maintaining a high level of accuracy. However, selecting the appropriate set of inducing points is critical to obtaining a good sparse approximation. The inducing points should be representative of the entire input space to ensure that the resulting approximation captures the full complexity of the original GP. If the selection of inducing points is suboptimal, the accuracy of the approximation may suffer. Therefore, while sparse approximation can be a useful tool, the process of selecting inducing points requires careful consideration to achieve a successful approximation.

The literature has proposed several sparse methods that use inducing points, such as those presented by~\cite{seeger2003fast, smola2001sparse, quinonero2005unifying}. In this thesis, we adopt the variational learning approach, which jointly infers the inducing inputs and the kernel hyperparameters by maximizing a lower bound of the actual log marginal likelihood~\cite{titsias2009variational}. The idea behind sparse Gaussian Processes is selecting a small set of points, known as inducing points, denoted by, $X^{in}_m = [x^{in}_i]^T_{1\leq i\leq m_m} \in \mathbb{R}^{m_m \times n_x}, ~ m_m \leq m_s$, to improve the approximation of the marginal likelihood. By using this subsample, the approximate posterior GP mean and covariance function depends on a mean vector $\mu_m$ and a covariance matrix $A_m$ as:
\begin{subequations}
    \begin{align}
    \widehat{\mu}(x^{in}_*) = k^T_{m*}(K_{mm} + \sigma^2_{m}I)^{-1}\mu_{m} ~~~~~~~~~~~~~~~~~~~~~~ \\
    \widehat{\sigma}^2(x^{in}_*)  = k(x^{in}_*, x^{in}_*) - k^T_{m*}(K_{mm} + \sigma^2_{m}I)^{-1}k_{m*} + k^T_{m*}K_{mm}A_mK_{mm}k_{m*}
    \end{align}
\end{subequations}
where 
\begin{gather*}
 \mu_{m} = \sigma^{-2}_{m} K_{mm} \Gamma K_{mn}y^{out}, \quad A_m = K_{mm} \Gamma K_{mm}\\
 \Gamma = (K_{mm} + \sigma^{-2}_{m} K_{mn} K_{nm})^{-1}
\end{gather*}
where $K_{mm}$ is $m_m\times m_m$ covariance matrix of $\{X^{in}_i\}_{i=1}^{m_m}$; $K_{mn}$ is $m_m\times m_s$ covariance matrix between $\{X^{in}_i\}_{i=1}^{m_m}$ and the training set $X^{in}$; $k_{m*}$ is the cross-covariance vector. 

In general, sparse approximation is a technique that reduces the time complexity of GP prediction from $O(m_s^3)$ to $O(m_s m_m^2)$, while still preserving important information from the original data. This means that instead of having to compute all pairwise distances between $m_s$ data points, we only need to compute $m_s$ distances using a subset of informative points $m_m \leq m_s$. The resulting prediction accuracy is nearly as good as the full GP but with significantly less computational cost.

\section{Conclusion} \label{ch3_conclusion}
This chapter has presented a concise introduction to the fundamental concepts of machine learning, with a focus on supervised learning and regression tasks. The Gaussian process framework has been explained in detail as a powerful tool for supervised regression modeling.

While unsupervised and semi-supervised learning are two additional important frameworks in machine learning, they fall outside the scope of this thesis and have not been covered.

The purpose of this chapter is to provide a foundation for the subsequent introduction and application of the Gaussian process framework, which is the primary focus of this thesis.
 ~\\~\\~\\~\\~\\~\\~\\~\\~\\~\\~\\~\\~\\
\addcontentsline{toc}{chapter}{Part II: Contribution} \label{part_2}
\begin{center}{\Huge \textsc{\underline{Part II}}\\~\\ \textsc{Contribution}}\end{center}\normalsize
 \chapter{Data Driven CC-OPF using Gaussian Process}
\label{ch:4}
\section{Introduction} \label{ch4_intro}

This chapter presents our first contribution in the form of a full GP CC-OPF approach that replaces the traditional AC-PF balance equations with GPR. The GPR is learned from the input and output data of the full AC-PF and is embedded in the CC-OPF problem. To propagate input uncertainties, various approximation techniques such as Taylor Approximation (TA) and Exact Moment Matching (EM) are used. This allows an analytical reformulation of the CC-OPF to determine the optimal solution by linearizing around the expected operating point and assuming normally distributed deviations. The proposed approach has been evaluated on various system buses such as IEEE9, 39 and 118. The approach and results of this chapter have been published in the \textit{International Journal of Electrical Power \& Energy Systems (IJEPES)}~\cite{mile1}.

The rest of this chapter is organized as follows: Section~\ref{ch4_problem} introduces the problem statement, research question, and hypothesis, while Section~\ref{ch4_approach} presents the proposed approach. Section~\ref{ch4_results} discusses the experimental results that support the approach, and finally, Section~\ref{ch4_conclusion} provides a summary of the outcomes of the study and suggests areas for further research.
\section{Problem Statement} \label{ch4_problem}
The AC-OPF problem is crucial for efficient and reliable power systems and markets. However, it is complex in many ways: economically, electrically and computationally. In this research, the computational problem of ACOPF will be considered which is a consequence of the lack of a fast and robust solution technique for the full AC-OPF. Usually, approximation techniques are employed to obtain reasonably acceptable solutions for the AC-OPF. While these techniques may reduce computational costs, they can also compromise the reliability of solutions, resulting in billions of dollars in unnecessary costs and environmental damage from emissions and energy waste~\cite{cain2012history}. 

Improving the solving of AC-OPF is crucial, as seen from Energy Information Administration (EIA) data on wholesale electricity prices and US and World energy production~\cite{eia}. Improvements can increase market efficiency by reducing costs by 5\%, resulting in billions of dollars in savings per year. \cite{oneill2011recent} states that installing new software with an improved solution can cost less than $10$ million dollars, leading to potential benefit/cost ratios in the range of 10 to 1000. Therefore, small increases in dispatch efficiency can have significant economic benefits. Thus, efficient and reliable AC-OPF solutions are crucial for the optimal operation of power systems and markets. Investing in improving the computational methods for solving the AC-OPF problem can offer significant economic and environmental benefits.

The integration of renewable energy sources (RES) into modern power grids provides significant advantages in terms of reducing energy production costs and carbon emissions, thereby promoting environmental protection. However, the rise in RES generation, such as wind and solar, poses significant challenges for Transmission System Operators (TSOs) in solving AC-OPF. The intermittent nature of RES generation is the most significant challenge as it poses a significant threat to grid security and can result in power outages.

The traditional approach to power grid operation and control, which relies on a fixed generation profile, is not feasible for grids with a substantial share of RES. To address this challenge, there is a need for a new class of optimization and control algorithms that account for generation uncertainty. These algorithms should be capable of handling the challenges that come with integrating intermittent RES generation into the grid.

This thesis addresses the stochastic AC-OPF problem, which is made more challenging due to the inclusion of uncertain generation. Various approaches have been proposed to tackle stochastic OPF problems, including robust optimization (RO), probabilistic OPF (P-OPF), and chance-constrained (CC) optimization. In this work, we focus on the Chance-Constrained Optimal Power Flow (CC-OPF) approach~\cite{morillo2022distribution, wu2019chance}, which aims to minimize power generation costs while providing optimal power dispatch in the face of changing demand and operational conditions of a power grid. The CC-OPF problem ensures that the operational constraints, such as nodal voltage limits and transmission capacity limits, are met with a high probability~\cite{lubin2019chance, roald2017chance}. This provides a guarantee of grid resilience against sudden changes in RES generation. However, the chance-constrained (CC) OPF remains a challenging optimization problem both in terms of solution accuracy and computational complexity~\cite{venzke2020chance}.

Researchers have been actively exploring ways to solve the CC-OPF problem, which remains a challenging optimization problem. Previous studies, such as \cite{calafiore2006scenario} and \cite{vrakopoulou2013probabilistic}, have employed Monte Carlo (MC) scenario-based CC-OPF with nonlinear AC-PF equations. However, this approach is not scalable when the system size or the number of scenarios increases \cite{mezghani2020stochastic}. To address this issue, advanced sampling policies have been proposed, such as importance sampling \cite{mezghani2020stochastic, lukashevich2021importance, lukashevich2021power}, and active sampling \cite{capitanescu2012cautious, owen2019importance}. These methods aim to reduce the number of required samples while maintaining solution accuracy, making it possible to tackle larger and more complex CC-OPF problems. Unlike the scenario-based approach for solving CC-OPF, this study focuses on the analytical approach, which offers better computational performance for large-scale power systems by using certain distribution functions to model uncertainty. Although many papers have proposed convex/linear approximations to the chance-constrained problem~\cite{lubin2015robust, roald2016corrective, du2021chance}, most of them are conservative and cannot handle large variations in power generation and/or demand. Another line of research employs the nonlinear Polynomial chaos expansion to model uncertainties in power generation and demand, but this approach often lacks good computational performance~\cite{muhlpfordt2019chance, muhlpfordt2016solving}. Additionally, most of the prior work relies on accurate knowledge of power system parameters, which may not be available due to insufficient and irregular equipment calibration. This challenge becomes even more critical in micro-grids and low-voltage grids where real-time metering capabilities are reduced, and parametric uncertainties are prevalent.

According to the aforementioned problems, this thesis is driven by the need to address the challenges posed by the computational complexity of solving stochastic CC-OPF problems without compromising the quality of optimal solutions. To this end, the research question is focused on how to strike a trade-off between the computational complexity and optimal solutions of the problem. The hypothesis put forward is that a data-driven approach, leveraging machine learning techniques, can enhance the solution of CC-OPF, leading to a trade-off between computational complexity and optimal solutions.

This chapter introduces a novel approach to solving the CC-OPF problem that balances computational complexity and solution accuracy. Instead of traditional approaches that rely on power balance equations, we propose using Gaussian process regression (GPR) to learn an approximation to power flow equations based on empirical samples of RES and load fluctuations. GPR is a model-free machine learning algorithm that can fit any smooth nonlinear function and offers a reasonable complexity-accuracy tradeoff~\cite{dudley2010sample, schulz2018tutorial, liu2020gaussian}. Our approach differs from prior work, which has focused on deterministic OPF formulations using GPR \cite{pareek2020gaussian, xu2020probabilistic} or neural network-based power-flow models \cite{pan2020deepopf, kekatos}. Such models do not consider CC-OPF with input uncertainty. Neural network-based stochastic AC-OPF has been proposed in \cite{gupta2021dnn}, but considers scenario-based CC-OPF. Instead, our GP-based formulation accounts for both parametric and input uncertainties, including analytic CC-OPF constraints. This approach exploits uncertainty propagation in Gaussian processes, resulting in a more accurate and efficient solution to the CC-OPF problem.

\section{Proposed Approach} \label{ch4_approach}

The problem given in~\ref{cc-opf} is not tractable to solve. There are two main reasons for this. Firstly,~\ref{cc-opf_b} is a semi-infinite problem, meaning that the set of fluctuations $W$ is uncountable. Secondly, the chance constraints~\ref{cc-opf_c}~-~\ref{cc-opf_i} are not tractable for solving by optimization solver. Accordingly, we reformulated~\ref{cc-opf} as a nonlinear programming (NLP) problem with learned GPR in the constraints, such that an optimization solver can
handle it.

Following~\cite{bienstock2014chance}, the objective function is reformulated as:
\begin{gather}\label{eq:obj} 
 \mathbb{E}[c_i(p_{g,i}(\omega))] = \sum_{i\in \mathcal{G}} \left\{ c_{2,i}(p_{g,i}^{2} + tr(\Sigma_{\omega})\alpha_{i}^{2}) + c_{1,i} p_{g,i} + c_{0,i}\right\}
\end{gather}
where $\{c_{2,i}, c_{1,i}, c_{0,i}\}_{i=1}^{\mathcal{G}} \geq 0 $ are scalar cost coefficients and $\alpha$ is the participation factor. Note that $\alpha$ (participation factor) is not a standard (fixed) value, but variables to be optimized with constraints $\sum_{i=1}^{\mathcal{G}}\alpha_i=1$ and $\alpha_k \geq 0$. Accordingly, the objective function is a convex quadratic function of $p_{g}$ and $\alpha$.

Reformulating chance constraints to the tractable form implies reformulating both output and decision input constraints. According to \cite{hewing2019cautious}, a GPR individual output probabilities could be expressed in terms of the mean $\mu$
and variance $\sigma^2$:
\begin{gather}\label{eq:pr_y}
 \begin{cases}
 \texttt{Prob}(\mu_{i} + \sqrt{\sigma^2_{i}} \leq y_i^{out_{\max}}) \leq 1-\epsilon_{i}\\
 \texttt{Prob}(\mu_{i} - \sqrt{\sigma^2_{i}} \geq y_i^{out_{\min}}) \leq 1-\epsilon_{i}\\
 \end{cases}
\end{gather}
that is equivalent to:
\begin{gather}\label{eq:pr_y_1}
 y_i^{out_{\min}} + r_{i} \sqrt{\sigma^2_{i}} \leq \mu_{i} \leq y_i^{out_{\max}} - r_{i} \sqrt{\sigma^2_{i}}
\end{gather}
where $r_{i}$ represents the quantile function $\Phi^{-1}(1-\epsilon_{i})$ of the standard normal distribution and $i=1,...,n_y$; $n_y$ is number of outputs.

Similarly, the decision input constraints are~\cite{bienstock2014chance}:
\begin{gather}
 \begin{cases}
 \texttt{Prob}(p_{g,i} + \alpha_i\, tr(\Sigma_{\omega})\leq p_{g,i}^{max}) \leq 1-\epsilon_{p_{g,i}}\\
 \texttt{Prob}(p_{g,i} - \alpha_i\, tr(\Sigma_{\omega}) \geq p_{g,i}^{min}) \leq 1-\epsilon_{p_{g,i}}
 \end{cases}
\end{gather}
As in \eqref{eq:pr_y}, this formulation is equivalent to:
\begin{gather}\label{eq:pr_u_1}
 p_{g,i}^{min} + r_{p_{g,i}} \alpha_i\, tr(\Sigma_{\omega}) \leq p_{g,i} \leq p_{g,i}^{max} - r_{p_{g,i}} \alpha_i\, tr(\Sigma_{\omega})
\end{gather}
where $r_{p_{g,i}} = \Phi^{-1}(1-\epsilon_{p_{g,i}})$ and $i\in \mathcal{G}$.

By taking into account of the formulations in \eqref{eq:obj}, \eqref{eq:pr_y_1}, and \eqref{eq:pr_u_1}, the following \textbf{full} GP CC-OPF is derived:
\begin{subequations}\label{eq:opf4}
\begin{align}
 &\min_{p_g, \alpha, \mu, \sigma} \sum_{i\in \mathcal{G}} \{ c_{2,i}(p_{g,i}^{2} + tr(\Sigma_{\omega})\alpha_{i}^{2}) + c_{1,i}p_{g,i} + c_{0,i}\} \label{eq:Gp-cost}\\
 &\text{s.t.~} \sum_{i\in \mathcal{G}} \alpha_i = 1, ~ \alpha_i \geq 0 \\
 &~~~~~ \sum_{i\in \mathcal{G}} p_{g,i} = \sum_{i\in \mathcal{L}} p_{l_i} - \sum_{j\in \mathcal{L}} p_{rs_j} \label{eq:balance} \\
 &~~~~~~ \mu_i = \mu_i(x^{in}_*) \label{eq:balance_mean}\\ 
 &~~~~~~ \sigma_i^{2} = \sigma_i^{2}(x^{in}_*) \label{eq:balance_var}\\
 &~~~~~~ y_i^{out_{min}} + \lambda_{i} \leq \mu_{i} \leq y_i^{out_{max}} - \lambda_{i}\label{eq:const_var1}\\
 &~~~~~~ p_{g,i}^{min} + \lambda_{p_{g,i}} \leq p_{g,i} \leq p_{g,i}^{max} - \lambda_{p_{g,i}} \label{eq:const_var2}
\end{align}
\end{subequations}
where $\lambda_{i} = r_{i} \sqrt{\sigma^2_{i}}$ for $i\in n_y$ and $\lambda_{p_{g,i}} = r_{p_{g,i}} \alpha_i\, \sqrt{tr(\Sigma_{\omega})}$ are uncertainty margins. By $\sqrt{tr(\Sigma_{\omega})}$ the standard deviation of the total active power imbalance $\Omega$ is represented. Equation \eqref{eq:balance} describes the balance equation in which decision variables of controllable generators have to be satisfied in the power system keeping the balance between generation and consumption, while \eqref{eq:balance_mean} and \eqref{eq:balance_var} represent GPR output mean and variance formulation. Thus, we replace the standard AC power flow equation with a data-driven method using these equations.

The proposed OPF optimization method using the GP-based model results in a non-convex optimization problem that can be challenging to solve. However, for the chosen twice differentiable SE kernel and zero prior means, the second-order derivative information of all quantities is available. Thus, the problem can be solved using sequential quadratic programming or the nonlinear interior-point method. We used a primal-dual interior-point algorithm with a filter line-search approach, and we have used the IPOPT~\cite{wachter2006implementation} solver in the non-linear optimization framework CasADi~\cite{andersson2019casadi}. To solve this optimization problem, we assume simultaneous approaches that simultaneously treat $p_g$, $\alpha$, $\mu$, and $\sigma^2$ as optimization variables.

\subsection{Full GP CC-OPF Pipeline}
The proposed full GP CC-OPF problem is solved using a three-stage procedure. In the first stage, historical data or a simulator is used to collect fluctuations of power demand and RES power generation, along with dependent variables. In the second stage, Gaussian process regression is employed to provide a data-driven probabilistic approximation for the dependent variables in the power balance equations, which are much more tractable than the PF equations. This approximation simplifies the optimization problem significantly. A synthetic example illustrating this procedure is shown in Figure~\ref{fig:gp_1}. In the third stage, the probability for constraint violations, referred to as uncertainty margins, is computed using Eq.~\ref{eq:pr_y_1} and~\ref{eq:pr_u_1}. This probability has a straightforward expression for Gaussian process-based modeling with input uncertainty and enables the CC-OPF problem to be transformed into a deterministic optimization problem. Solving this problem gives us a minimal cost of power dispatch that meets the probabilistic security constraints.

\begin{figure}[H]
\centering
\includegraphics[width=0.78\textwidth]{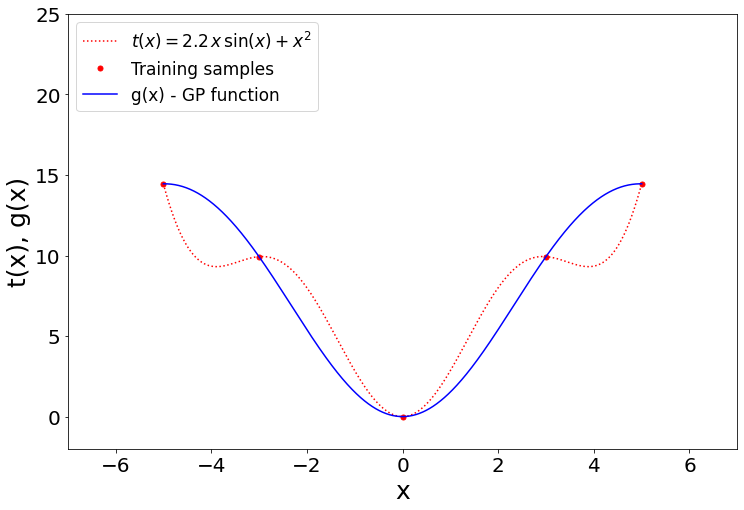}
\caption[Synthetic example of the power of the Gaussian process regression]{A synthetic example illustrates the power of the Gaussian process regression. Function $t(x) = 2.2 x\sin x + x^2$ stands for the initial equation and $g(x)$ is its data-driven approximation via the Gaussian process with the squared exponential kernel. We refer to the approximation as a GP-function.}
\label{fig:gp_1}
\end{figure}

\subsection{Input Covariance}
As discussed in Section~\ref{sec:uncer_pred}, we make an assumption that the unseen input points $x^{in}_{*}$ for the GPR that follow a normal distribution with a vector mean denoted by $\mu_{x_*} \in \mathbb{R}^{n_x}$ and a covariance matrix denoted by $\Sigma_{x_*} \in \mathbb{R}^{n_x\mathsf{x} n_x}$. The mean values correspond to the optimized control variables, while the uncertain parameters are forecasted as:
\begin{gather*} 
 \mu_{x^{in}_*} = [p_g^\top, p_{l}^\top, p_{rs}^\top]^\top
\end{gather*}
where $p_g=[p^1_g, p^2_g, ..., p^{n_{\mathcal{G}}}_g]^\top$, $p_l=[p^1_l, p^2_l, ..., p^{n_{\mathcal{L}}}_l]^\top$ and $p_{rs}=[p^1_{rs}, p^2_{rs}, ..., p^{n_{\mathcal{L}}}_{rs}]^\top$.
Due to the linear feedback, the input covariance matrix $\Sigma_{x_*}$ needs to be correctly defined as $p_g$ is correlated with uncertainty $\omega$ in $p_l$ and $p_{rs}$. This matrix is:
\begin{equation} 
 \Sigma_{x^{in}_*} =
 \begin{bmatrix}
 \Sigma_g & \Sigma_{g\omega}\\\
 \Sigma_{g\omega}^\top & \Sigma_{\omega}
 \end{bmatrix}
\end{equation}
so that $\Sigma_g \in \mathbb{R}^{n_{\mathcal{G}} \mathsf{x} n_{\mathcal{G}}}$, $\Sigma_{\omega} \in \mathbb{R}^{n_{\mathcal{L}} \mathsf{x} n_\mathcal{L}}$ and $\Sigma_{g\omega} \in \mathbb{R}^{n_{\mathcal{G}} \mathsf{x} n_\mathcal{L}}$ are covariance sub-matrices.

The covariance sub-matrix $\Sigma_{\omega}$ represents uncertain variations in loads and renewable sources. We consider this matrix as diagonal:
\begin{gather}\label{cov_1}
 \Sigma_{\omega} =
 \begin{bmatrix}
 \sigma^2_{l_1} & ... & 0 & 0 & ... & 0 \\
 ... & ... & ... & ... & ... & ...\\
 0 & ... & \sigma^2_{l_{\mathcal{L}}} & 0 & ... & 0\\
 0 & ... & 0 & \sigma^2_{rs_1} & ... & 0 \\
 ... & ... & ... & ... & ... & ...\\
 0 & ... & 0 & 0 & ... & \sigma^2_{rs_{\mathcal{L}}} \\
 \end{bmatrix}
\end{gather}
and assume that fluctuations between stochastic power injections are independent, consistent with prior work \cite{bienstock2014chance}. 

Sub-matrix $\Sigma_g$ refers to controllable generator deviations caused by loads and renewable sources fluctuations keeping the balance of the system. Since controllable variables $p_g$ are dependent on forecasted uncertain injections $p_l$ and $p_{rs}$, the deviations of $p_g$ are also dependent on fluctuations of $\omega$ as follows:

\begin{gather}\label{cov_2}
\Sigma_{g} =
 \begin{bmatrix}
 \alpha^2_1 tr(\Sigma_w) & ... & 
 \alpha_1 tr(\Sigma_w) \alpha_{n_{\mathcal{G}}}\\
 ... & ... & ... \\
 \alpha_{n_\mathcal{G}} tr(\Sigma_{\omega}) \alpha_1 & ... & \alpha^2_{n_\mathcal{G}} tr(\Sigma_{\omega})\\
 \end{bmatrix}
\end{gather}

The covariance elements between dependent decision variables and stochastic parameters are expressed in sub-matrix
$\Sigma_{g\omega}$ as:
\begin{gather}\label{cov_3}
 \Sigma_{g\omega} =
 \begin{bmatrix}
 \alpha_1\sigma^2_{l_1} & ... & \alpha_1\sigma^2_{l_{n_{\mathcal{L}}}} & \alpha_1\sigma^2_{rs_1} & ... & \alpha_1\sigma^2_{rs_{n_{\mathcal{L}}}} \\
 ... & ... & ... & ... & ... & ... \\
 \alpha_{n_{\mathcal{G}}}\sigma^2_{l_1} & ... & \alpha_{n_{\mathcal{G}}}\sigma^2_{l_{n_{\mathcal{L}}}} & \alpha_{n_{\mathcal{G}}}\sigma^2_{rs_1} & ... & \alpha_{n_{\mathcal{G}}}\sigma^2_{rs_{n_{\mathcal{L}}}} \\
 \end{bmatrix}
\end{gather}
\section{Experimental Setup and Results} \label{ch4_results}

The results presented in this chapter are obtained on an Intel Core i7-5500U CPU @ 2.40GHz and 8GB of RAM. To simulate the experiment, we use our python gp-ccopf\footnote{\url{https://github.com/mile888/gp_cc-opf}} framework developed for this article and pandapower~\cite{thurner2018pandapower} package to validate results.

\subsection{Case Study}
We use the $9$, $39$, and $118$-bus IEEE Test Systems to evaluate the performance and scalability of the proposed GP CC-OPF method. Simulations are conducted using pandapower~\cite{thurner2018pandapower} package for Python.

\textit{IEEE 9-bus Test Systems} consists of 3 generators and 3 loads on the high-voltage level of $V_{n} = 345kV$. We introduce two renewable source generators at buses 3 and 5 with a total forecast power output of 80 MW, which is approximately 25.4\% of the total active power demand. We assume that the reactive to real power ratio is 0.3. All loads and renewable sources are uncertain.

\textit{IEEE 39-bus Test Systems} consists of 10 generators and 21 loads with $V_{n} = 345kV$. Renewable source generators are placed at buses 1, 11, 14, 21, 23, and 28 with a total forecast power output of 1260 MW, which is approximately 20.2\% of the total active power demand. The assumed power ratio is 0.3. Also, we consider that all loads and renewable sources are assumed to be uncertain.

\textit{IEEE 118-bus Test Systems} consists of 54 generators and 99 loads with $V_{n} = 345kV$. The six renewable source generators are placed at buses 10, 27, 47, 51, 78, and 92 with a total forecast power output of 1260 MW, which is approximately 29.7\% of the total active power demand. All loads and RES are considered as uncertain elements with an assumed reactive to real power ratio of $0.3$.

The forecast errors are modeled as zero mean, multivariate Gaussian random variables with a standard deviation corresponding to 15\% of the forecasted loads ($\sigma_l = 0.15\, p_l$) and 30\% of the forecasted renewable sources ($\sigma_{rs} = 0.3\, p_{rs}$). The acceptable violation probabilities $\epsilon$ are set to $\epsilon_{p_g}=0.1\%$ and $\epsilon = 2.5\%$. Since generation constraints are physically impossible to violate, we assume a very small percentage of violations ($\epsilon_{p_g}$), as proposed in Swissgrid (the swiss system operator)~\cite{abbaspourtorbati2015swiss} for their reserve procurement process. In contrast, output constraints are soft constraints where some violations can be tolerated ($\epsilon$) if the magnitude and duration are not too large or removed through additional control actions such as generation re-dispatch~\cite{roald2017chance}.

\subsection{Evaluation Procedure and Metrics}
To evaluate the prediction quality of the GPR optimized using training data, we use the root mean squared error (RMSE) metric for each output $k$ as:
\begin{gather*}
 RMSE_k = \sqrt{\frac{1}{m_*}\sum_{i=1}^{m_*} (y^{out}_{i,k} - \hat{y}^{out}_{i,k})^2}
\end{gather*} 
where $y^{out}$ and $\hat{y}^{out}$ are the actual (AC-PF) and predicted (GPR) mean values for testing data, and $m_*$ is the number of test samples. The average evaluation score of all output variables is calculated as follows:
\begin{equation}\label{eq:rmse_1}
 RMSE = \frac{\sum_{k=1}^{n_y} RMSE_k}{n_y}
\end{equation}

Solving the data-driven CC-OPF with GP-based constraints for different test data, we compute the mean of output variables $\hat{y}^{out}$ (Eq.~\eqref{eq:balance_mean}) and report the RMSE against the variables computed by AC-PF. To analyze chance constraints feasibility, we compare the distribution of output variables, computed analytically in GP CC-OPF, with the true empirical spread computed via AC-PF with Monte-Carlo samples of uncertainty $\omega$ around the mean inputs (loads, RES, and optimized generator set-points). 

To measure the performance of the solution, we also compute the computation time, generation cost (Eq.~\eqref{eq:Gp-cost}), and the empirical spread of output variables for (a) model-based scenario CC-OPF \cite{vrakopoulou2013probabilistic}, \cite{mezghani2020stochastic}, (b) AC-OPF on individual test samples (i.e., full recourse), and (c) AC-OPF for base case (mean loads and RES). Note that AC-OPF on the base case does not optimize for input uncertainty and presents the worst-case violations. On the other hand, the full-recourse solves AC-OPF for each sample separately and is the benchmark for feasibility as it is not restricted to linear generator feedback.

\subsection{Models Performance}
Proposed GPR models for IEEE 9, 39, and 118-bus systems are trained on randomly sampled data, generated as discussed in Section \ref{ch2_dataset}. The GPR model of the IEEE 9-bus system consists of 8 inputs features and 15 outputs, while the IEEE 39 bus has 37 inputs and 74 outputs. Likewise, the GPR model of the IEEE 118-bus system has 159 inputs and 291 outputs. In the case of the IEEE 9 bus case, we applied 75 random samples for training and 25 for validation. For the IEEE 39 bus system, training is conducted on 200 samples, and 65 samples are used for validation. Similarly, for the IEEE 118 system, 400 training samples and 100 validation samples are used. The average RMSE metric in Eq. \eqref{eq:rmse_1} for IEEE 9, IEEE 39, and IEEE 118 bus are RMSE = $7.72\cdot 10^{-5}~p.u.$, RMSE = $8.91\cdot 10^{-4}~p.u.$, and RMSE = $2.83\cdot 10^{-2}~p.u.$, respectively. The small errors suggest that the models for both systems have good generalization for deterministic inputs and that the hyperparameters are well optimized.

Next, we use the GP CC-OPF method with three approximation methods: first-order Taylor approximation (TA1), second-order Taylor approximation (TA2), and Exact moment matching (EM). The iterative solution interior-point algorithm is applied to find a feasible solution in $k$ iterations. In our simulations, the algorithm converged within a relatively small number of iterations for the convergence tolerance of $\epsilon_{tol} = 10^{-5}$, as shown in Table \ref{table:table2}. Table \ref{table:table2} also presents each approximation method's RMSE for GP-output against AC-PF based outputs, for the same inputs. The RMSE results are the same for TA1 and TA2 algorithms since both algorithms have the same mean function \eqref{eq:mean_TA1}. In contrast, the EM mean function depends on input covariance; in our case, the RMSE is better. However, the CPU time of optimization is greater for EM, with a sharper increase for the larger system.

\begin{table}[h!]
\centering 
\begin{tabular}{ c | c | c | c | c  } 
\hline 
System & Parameters & TA1 & TA2 & EM\\ [0.5ex] 
\hline 
\multirow{2}{*}{\vspace{-0.6 cm} IEEE 9} & RMSE [p.u.] & $6.35 \cdot 10^{-3}$ & $6.35 \cdot 10^{-3}$ & $5.95\cdot 10^{-3}$\\\cline{2-5}
 & No. iteration & 16 & 14 & 16\\\cline{2-5}
 & CPU time [s] & 0.35 & 4.97 & 6.60\\\hline 
\multirow{3}{*}{IEEE 39} & RMSE [p.u.] & $1.37 \cdot 10^{-2}$ & $1.37 \cdot 10^{-2}$ & $1.35 \cdot 10^{-2}$\\\cline{2-5}
 & No. iteration & 18 & 18 & 20 \\\cline{2-5}
 & CPU time [s]& 19.38 & 717.6 & 1221.2 \\\hline 
 \multirow{4}{*}{\vspace{0.52 cm} IEEE 118} & RMSE [p.u.] & $8.37 \cdot 10^{-1}$ & $8.37 \cdot 10^{-1}$ & $8.21 \cdot 10^{-1}$\\\cline{2-5}
 & No. iteration & 20 & 21 & 21 \\\cline{2-5}
 & CPU time [s]& 212.1 & 2987.1 & 4102.3  
 \\\cline{2-5}
\hline 
\end{tabular}
\caption[Results of the full GP CC-OPF approach]{GP CC-OPF method results for IEEE 9, IEEE 39 and IEEE 118 bus systems.}
\label{table:table2}
\end{table}

\subsection{Chance-constraint feasibility and uncertainty margins}
We plot the spread of centered CC-OPF output variables (voltages at PQ buses, reactive powers at controllable generators, and apparent power flows on the lines) for the IEEE 9-bus system in Fig.~\ref{fig:waveforms}. The three standard deviations (3 STD) intervals for uncertainty margins are analytically derived using the estimated GP-variance,  see Eq.~\eqref{eq:balance_var}. The different approximations considered for uncertainty propagation are marked as  blue (TA1), red (TA2), and black (EM). The empirical spread for the same output variables is derived using both GP-function and AC-PF equation with 5000 Monte Carlo (MC) samples of the input uncertainty $\omega$. Brown and yellow bars represent the upper and lower margins of the MC-based GPR model, while dark and light green bars are for the upper and lower margins of the MC-based AC-PF equation. Due to the asymmetrical MC-based distributions owing to nonlinear AC-PF, the upper and lower uncertainty margins computed as 99.73\% and 0.27\% probability, respectively (3 STD), are depicted separately. 

\begin{figure}[t]
 \begin{center}
\includegraphics[width=0.84\textwidth,height=6.5cm]{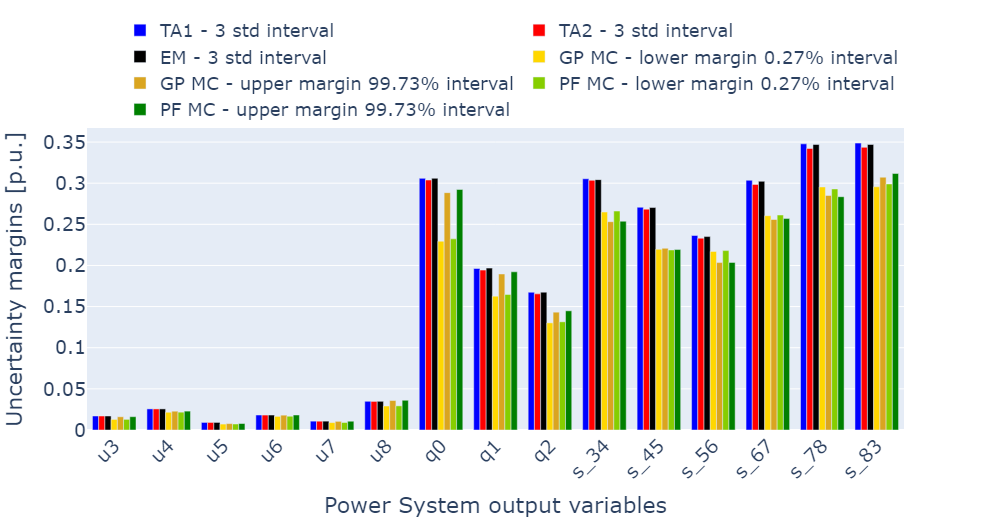}
 \caption[IEEE9 uncertainty margins]{Output uncertainty spread of IEEE 9.}\label{fig:waveforms}
 \end{center}
\end{figure}

Similar plots for nodal voltage, reactive power injections, and apparent power flows on lines of the $39$ and $118$ bus systems are presented in Fig.~\ref{fig:example_39} and Fig.~\ref{fig:example_118}. Note that part of the variables is shown in Fig.~\ref{fig:example_118} due to the impossibility of presenting all of them in the figure. Comparing the analytic results with the Monte Carlo based uncertainty margins, it is clear that they are close for most outputs, with a few exceptions where the analytic spread \emph{overestimate} the MC ones. 

\begin{figure}[!t]
 \centering
 \subfloat[Voltage output distributions]{\includegraphics[width=0.84\textwidth,height=6.1cm]{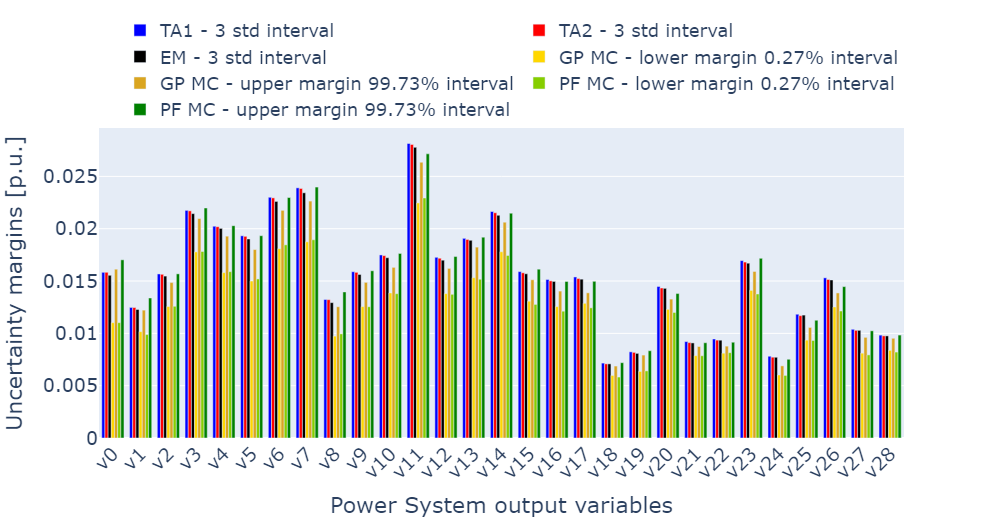}}%
 \qquad
 \subfloat[Reactive power output distributions]{\includegraphics[width=0.84\textwidth,height=6.1cm]{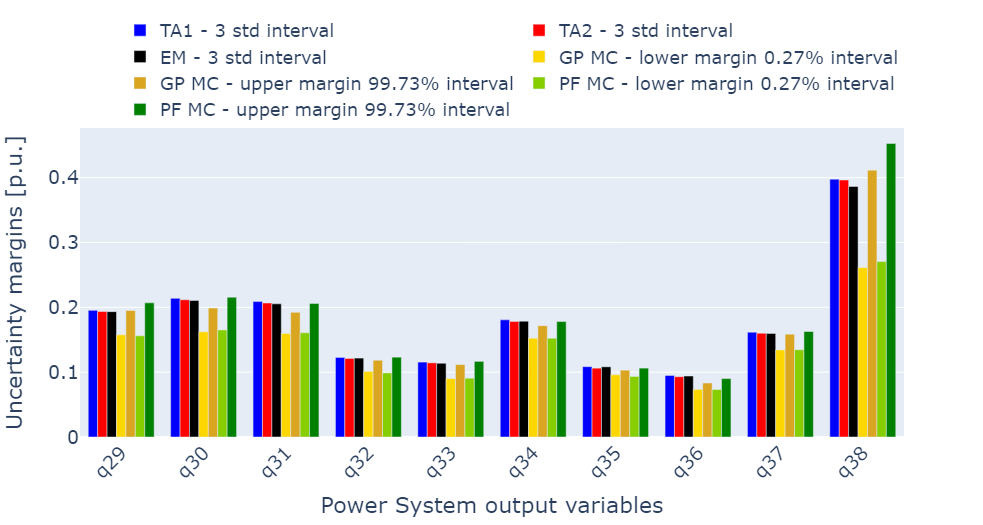} }%
 \qquad
 \subfloat[ Apparent power output distributions]{\includegraphics[width=0.84\textwidth,height=6.1cm]{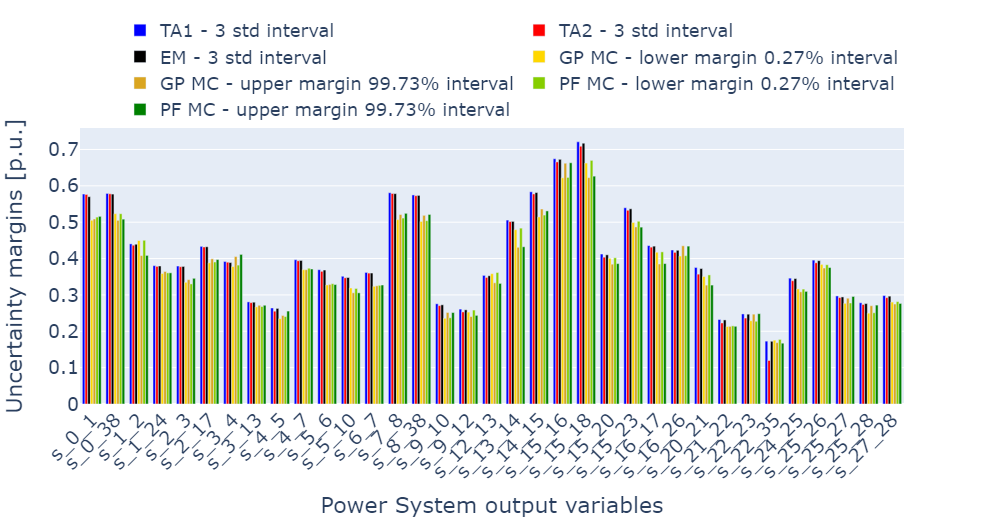}}%
 \caption[IEEE39 uncertainty margins]{Output uncertainty spread of IEEE 39.}%
 \label{fig:example_39}%
\end{figure}

\begin{figure}[!t]
 \centering
 \subfloat[\centering Voltage output distributions]{{\includegraphics[width=0.84\textwidth,height=6.1cm]{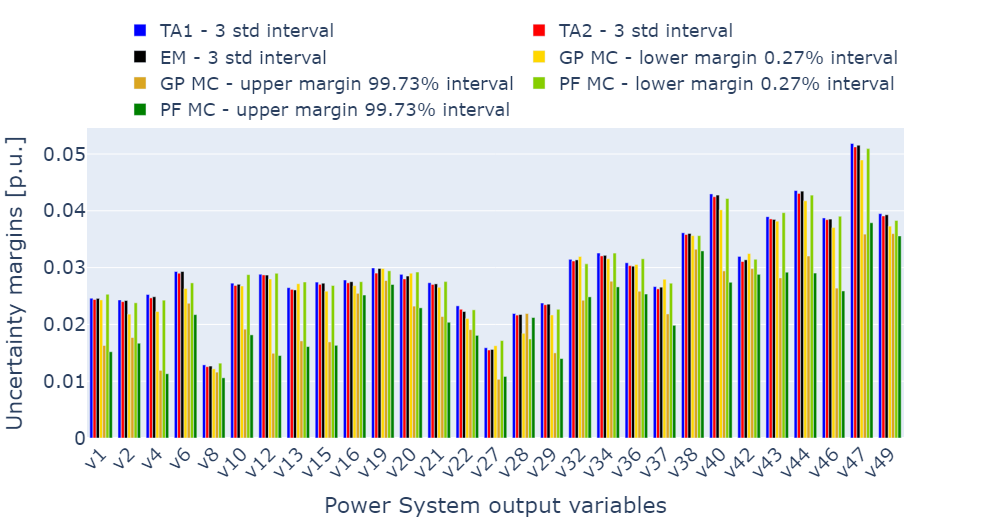} }}%
 \qquad
 \subfloat[\centering Reactive power output distributions]{{\includegraphics[width=0.84\textwidth,height=6.1cm]{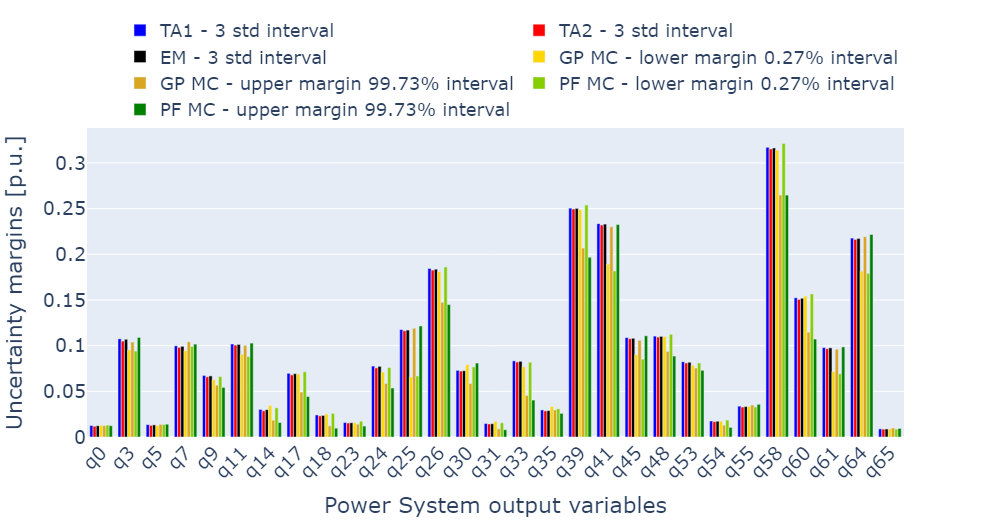} }}%
 \qquad
 \subfloat[\centering Apparent power output distributions]{{\includegraphics[width=0.84\textwidth,height=6.1cm]{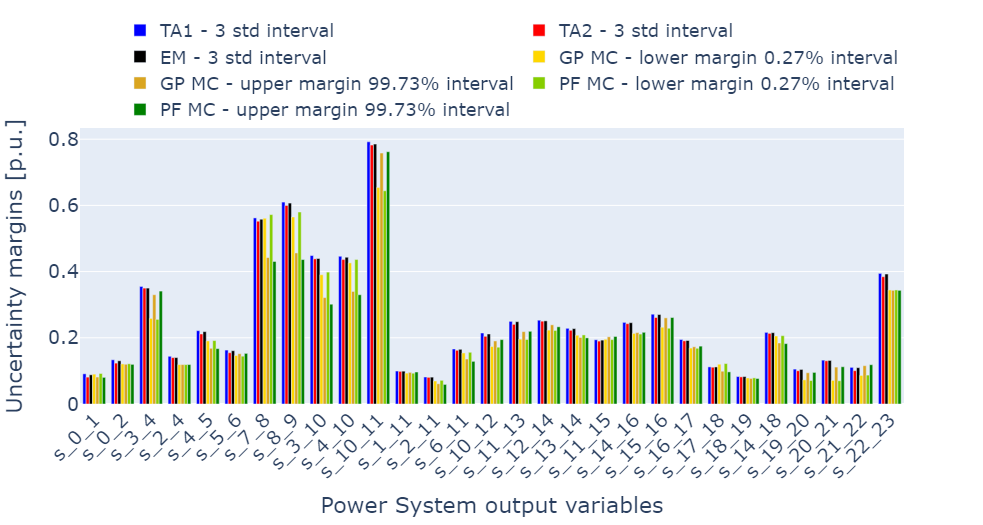} }}%
 \caption[IEEE118 uncertainty margins]{Output uncertainty spread of IEEE 118.}%
 \label{fig:example_118}%
\end{figure}

Note that as analytic margins satisfy the chance constraints, see Eq.~\eqref{eq:const_var1} and Eq.~\eqref{eq:const_var2}, the smaller MC empirical violations will also be within the chance-constrained limits, and hence feasible. Moreover, from Fig.~\ref{fig:waveforms}, Fig.~\ref{fig:example_39} and Fig.~\ref{fig:example_118}, MC margins for the GPR model and AC-PF equation are comparable, which implies that the GPR model can learn the nonlinear AC-PF accurately. For the IEEE 39 and 118 bus systems, we do observe small deviations between upper and/or lower margins for some variables, but this can be overcome by increasing the number of training data samples.

\subsection{Comparison with Model-based Scenario CC-OPF}
Here, we compare the cost, computational time, and CC-feasibility of the GP CC-OPF method (with TA1 approximation), with model-based scenario CC-OPF \cite{vrakopoulou2013probabilistic, mezghani2020stochastic}, with a varying number of scenarios. The empirical constraint violation for determined set-points is computed using $1000$ test MC samples. We also compare our method with two model-based setups:
\begin{itemize}
    \item[$(A)$] AC-OPF will full recourse, i.e., separate AC-OPF for each MC sample, 
    \item[$(B)$] and AC-OPF for the base case, i.e., AC-OPF for the mean input.
\end{itemize}

Note that though unrealistic, (A) considers generalized nonlinear feedback, while (B) doesn't optimize the participation factor $\alpha$ and uses equal generator participation for test MC samples. Hence they represent, respectively, model-based solutions with the best and worst feasible solutions. 

In Table~\ref{table:cost_fun}, we can observe that GP CC-OPF solution produces the desired reliability level of $\epsilon_y = 2.5\%$, with a cost between that of (A) and (B) for both test cases. Note that (A) solves separate AC-OPF for each MC sample and hence has no constraint violation but a higher cost. On the other hand, (B) 's set point is optimized by OPF for the mean loads and RES. Thus, it has a lower cost but large violations for test uncertainty. Their computational time is that of a single AC-OPF. For scenario CC-OPF, increasing the number of scenarios increases cost and reduces constraint violations for either test case. Although the number of scenarios necessary to get desired CC-feasibility is greater than $100$ for IEEE 39 and greater than what leads to a significantly higher computational time for scenario CC-OPF over GP CC-OPF where CC is implemented analytically. 

To study the empirical spread of output variables, we illustrate the voltage magnitude under uncertainty at bus 8 of IEEE9 in Fig.~\ref{fig:histcase9}, the voltage magnitude at bus 25 of IEEE39 in Fig.~\ref{fig:histcase39}, and the same variable at bus 47 of IEEE118 in Fig.~\ref{fig:histcase118}. As one can observe, voltage values for GP CC-OPF for the IEEE9 (Fig.~\ref{fig:histcase9a}) are centered further from the minimal voltage than the corresponding voltage values for 100 scenarios CC-OPF (Fig.~\ref{fig:histcase9b}). The latter leads to reduced violation of GP CC-OPF compared to scenario CC-OPF for the important lower voltage limit. Moreover, the latter's spread is more skewed to the constraint border. Note that (A) with full recourse has no violation while (B) for base-case has a higher violation. Similarly, for voltage at bus 25 of IEEE39 and bus 47 of IEEE118, the GP CC-OPF voltage spread (Fig.~\ref{fig:histcase39a} and~\ref{fig:histcase118a}) is further from the important upper limit compared to the solution of scenario CC-OPF (Fig.~\ref{fig:histcase39b} and~\ref{fig:histcase118b}).
\begin{table}[!t]
\centering
\begin{tabular}{l|l|l|l}
 \hline\hline
 System &
 \multicolumn{3}{c}{IEEE 9} \\\cline{2-4}
 \hline
 Parameters & Cost [\$] & Infeas. Prob. [\%] & Time (s)\\
 \hline
 A (full recourse)& $4.056\cdot 10^3$ & - & 0.86\\
 \hline
 B (base-case)& $3.467 \cdot 10^3$ & 9.76 & 0.86 \\
 \hline
 \textbf{GP CC-OPF} & $4.039 \cdot 10^3$ & 0.44 & 0.35\\
 \hline
 20 CC-OPF & $3.481 \cdot 10^3$ & 7.52 & 4.4\\
 \hline
 50 CC-OPF & $3.836 \cdot 10^3$ & 4.92 & 12.1\\
 \hline
 100 CC-OPF & $3.985 \cdot 10^3$ & 1.24 & 22.8\\
 \hline\hline
 \end{tabular}
 \begin{tabular}{l|l|l|l}
 \hline\hline
 System &
 \multicolumn{3}{c}{IEEE 39}\\\cline{2-4}
 \hline
 Parameters & Cost [\$] & Infeas. Prob. [\%] & Time (s)\\
 \hline
 A (full recourse) & $7.869\cdot 10^6$ & - & 0.93\\
 \hline
 B (base case)& $7.533 \cdot 10^6$ & 35.36 & 0.93 \\
 \hline
 \textbf{GP CC-OPF} & $7.752\cdot 10^6$ & 2.36 & 19.38 \\
 \hline
 50 CC-OPF & $7.642\cdot 10^6$ & 14.20 & 96.4 \\
 \hline
 100 CC-OPF & $7.696\cdot 10^6$ & 8.24 & 184.7 \\
 \hline
 200 CC-OPF& $7.813\cdot 10^6$ & 0.16 & 505.0 \\
 \hline\hline
 \end{tabular}
 \begin{tabular}{l|l|l|l}
 \hline\hline
 System &
 \multicolumn{3}{c}{IEEE 118}\\\cline{2-4}
 \hline
 Parameters & Cost [\$] & Infeas. Prob. [\%] & Time (s)\\
 \hline
 A (full recourse) & $26.91\cdot 10^6$ & - & 1.59\\
 \hline
 B (base case)& $26.39 \cdot 10^6$ & 49.97 & 1.59 \\
 \hline
 \textbf{GP CC-OPF} & $26.91\cdot 10^6$ & 2.47 & 212.1 \\
 \hline
 100 CC-OPF & $26.52\cdot 10^6$ & 28.12 & 683.3 \\
 \hline
 200 CC-OPF & $26.71\cdot 10^6$ & 19.62 & 1856.5 \\
 \hline
 500 CC-OPF& $26.90\cdot 10^6$ & 4.78 & 39537.1 \\
 \hline\hline
 \end{tabular}
 \caption[Cost functions and probability of violations]{Cost function values and probability of violation of a constraint at a solution}
 \label{table:cost_fun}
\end{table}

\begin{figure}[!t] 
 \centering
 \subfloat[Spread of voltages for GP CC-OPF and (B) AC-OPF for base case.]{
\includegraphics[width=0.48\textwidth,height=4.1cm]{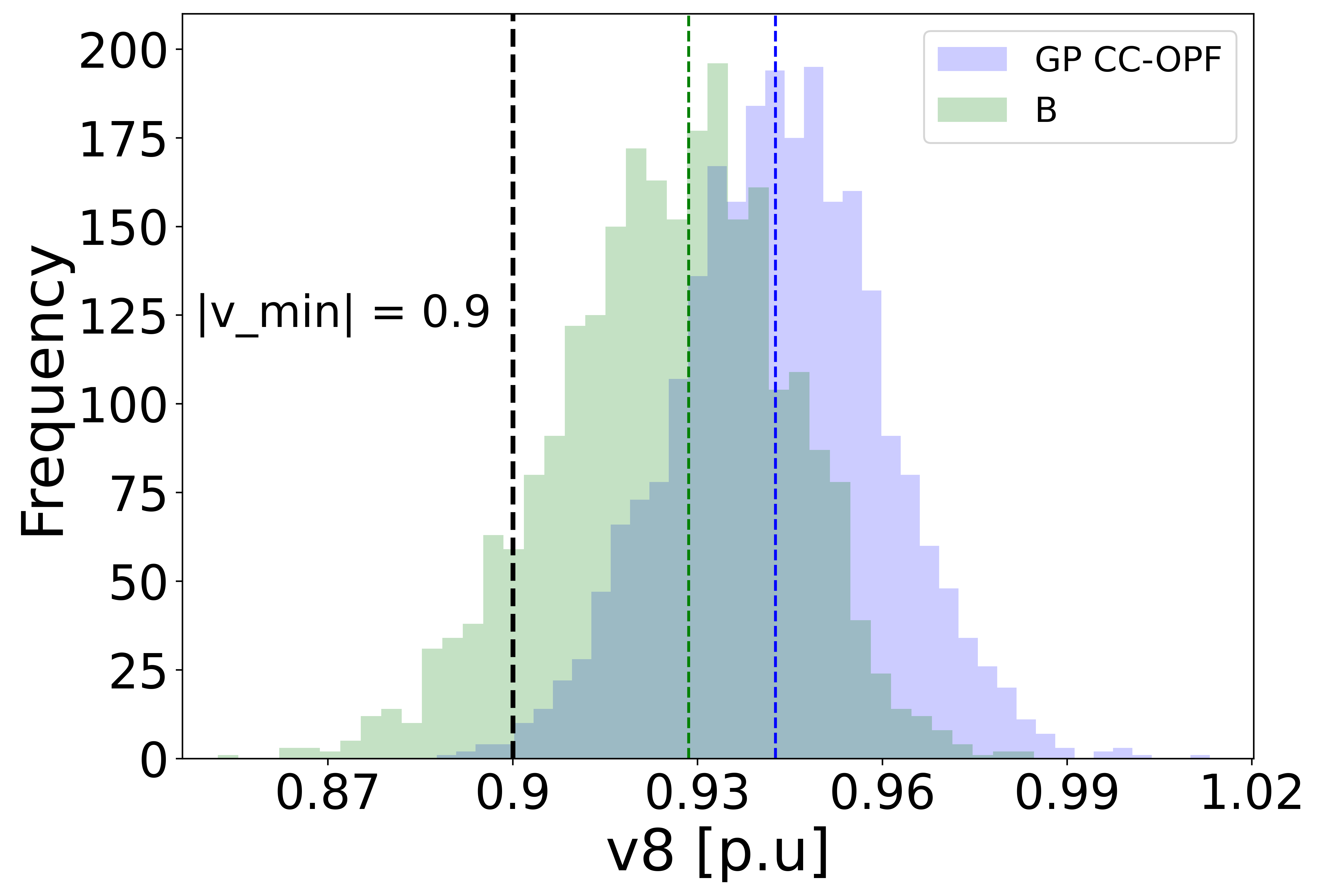}%
  \label{fig:histcase9a}%
  }
 \hfill%
 \subfloat[Spread of voltages for scenario CC-OPF and (A) AC-OPF with full recourse.]{ \includegraphics[width=0.48\textwidth,height=4.1cm]{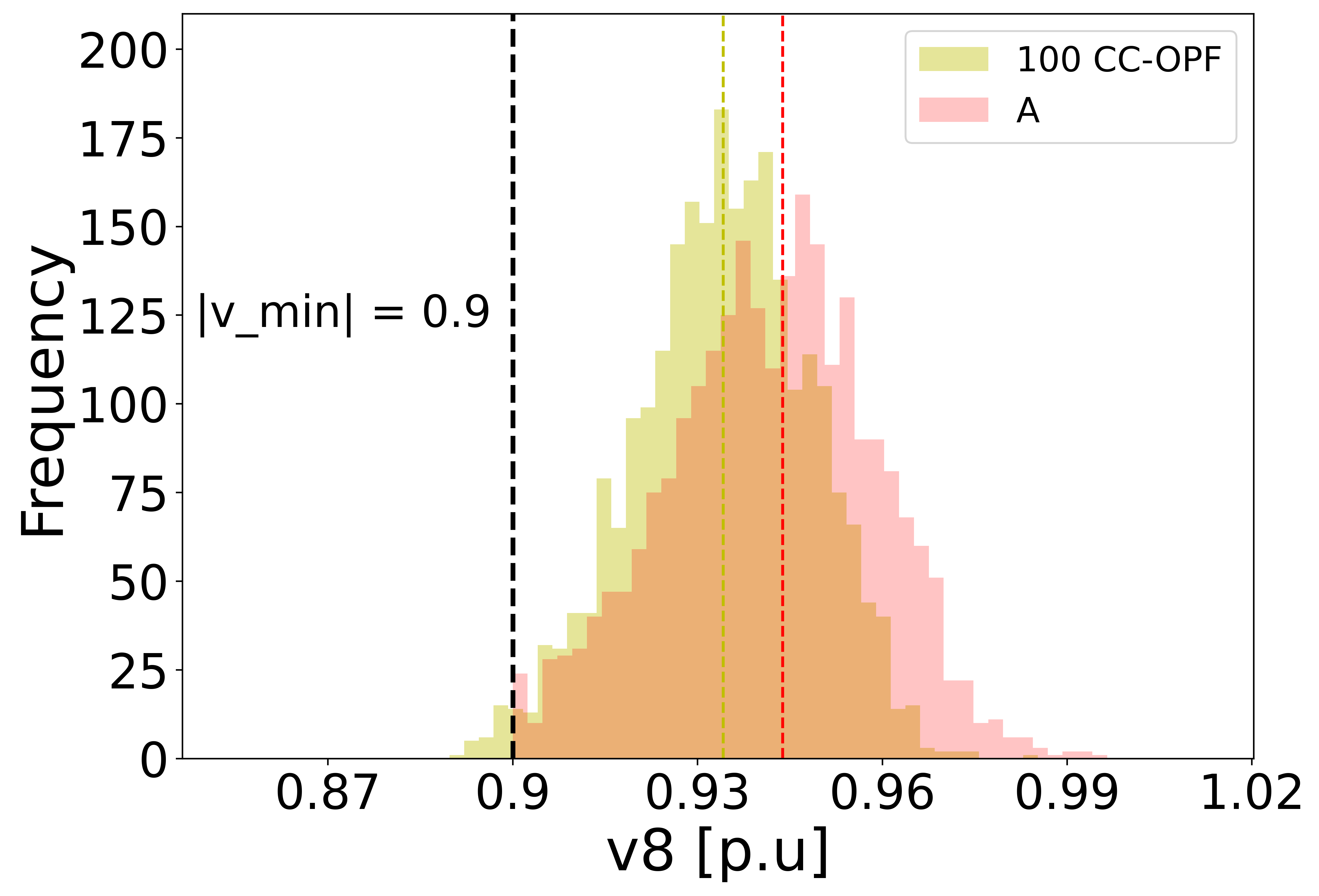}
  \label{fig:histcase9b}%
  }%
 \caption[IEEE9 empirical spread of voltages]{Case IEEE-9. Empirical Spread of voltages at Bus 8 under uncertainty.}
 \label{fig:histcase9}%
\end{figure}

\begin{figure}[!t] 
 \centering
 \subfloat[Spread of voltages for GP CC-OPF and (B) AC-OPF for the base case.]{%
\includegraphics[width=0.48\textwidth,height=4.1cm]{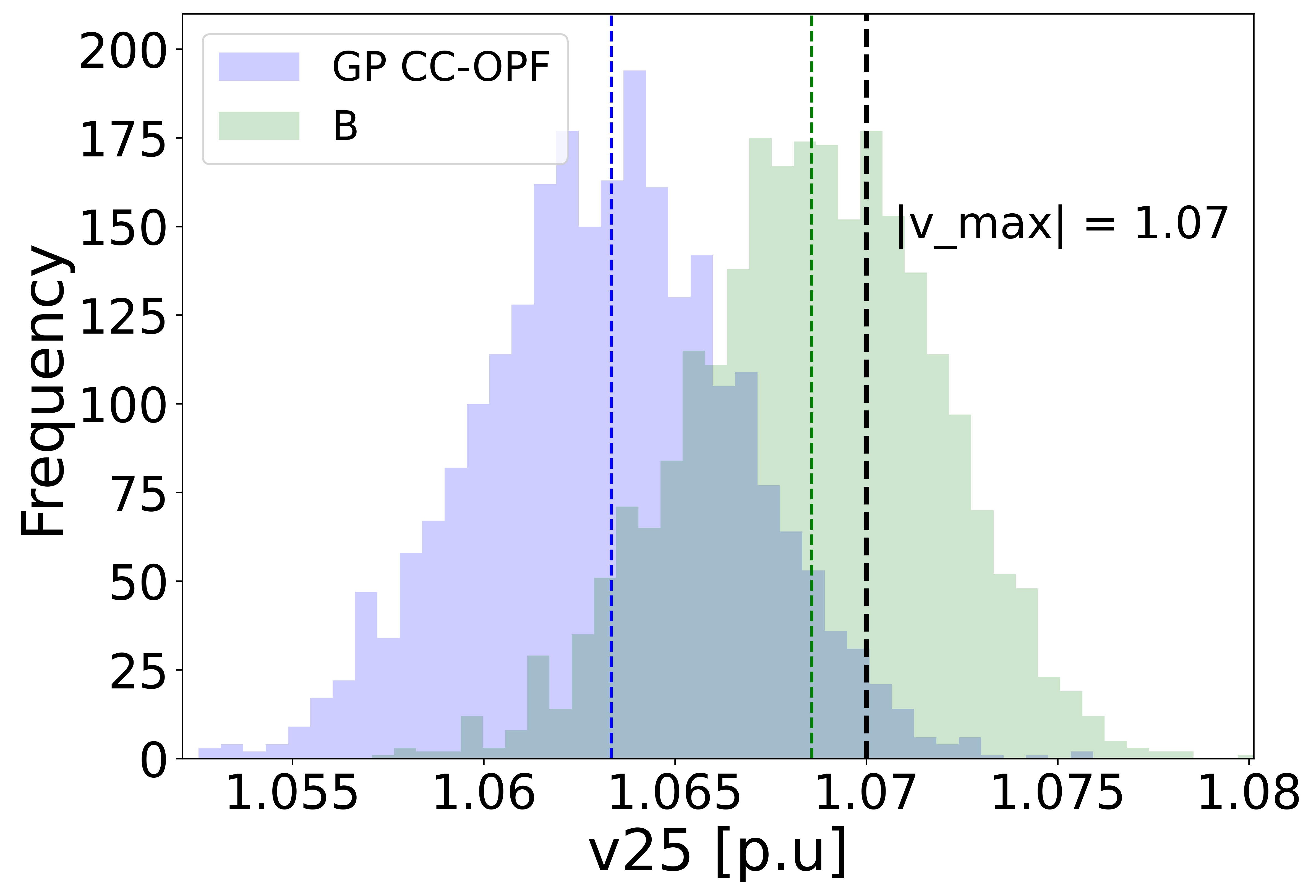}%
  \label{fig:histcase39a}%
  }%
 \hfill%
 \subfloat[The spread of voltages for scenario CC-OPF and (A) AC-OPF with full recourse.]{%
\includegraphics[width=0.48\textwidth,height=4.1cm]{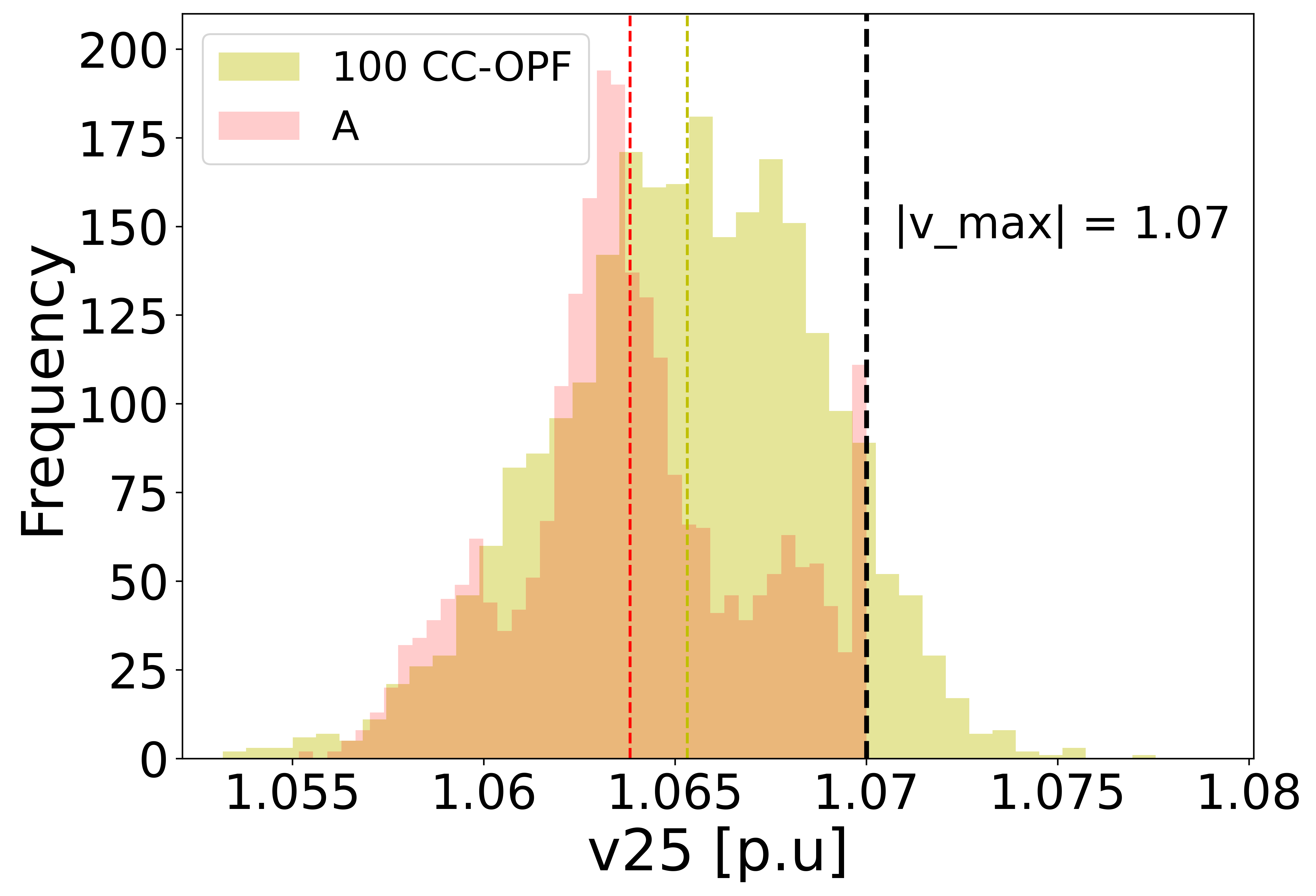}
  \label{fig:histcase39b}%
  }%
 \caption[IEEE39 empirical spread of voltages]{Case IEEE-39. Empirical Spread of voltages at Bus 25 under uncertainty.}
 \label{fig:histcase39}%
\end{figure}

\begin{figure}[ht] 
 \centering
 \subfloat[Spread of voltages for GP CC-OPF and (B) AC-OPF for the base case.]{%
\includegraphics[width=0.48\textwidth,height=4.1cm]{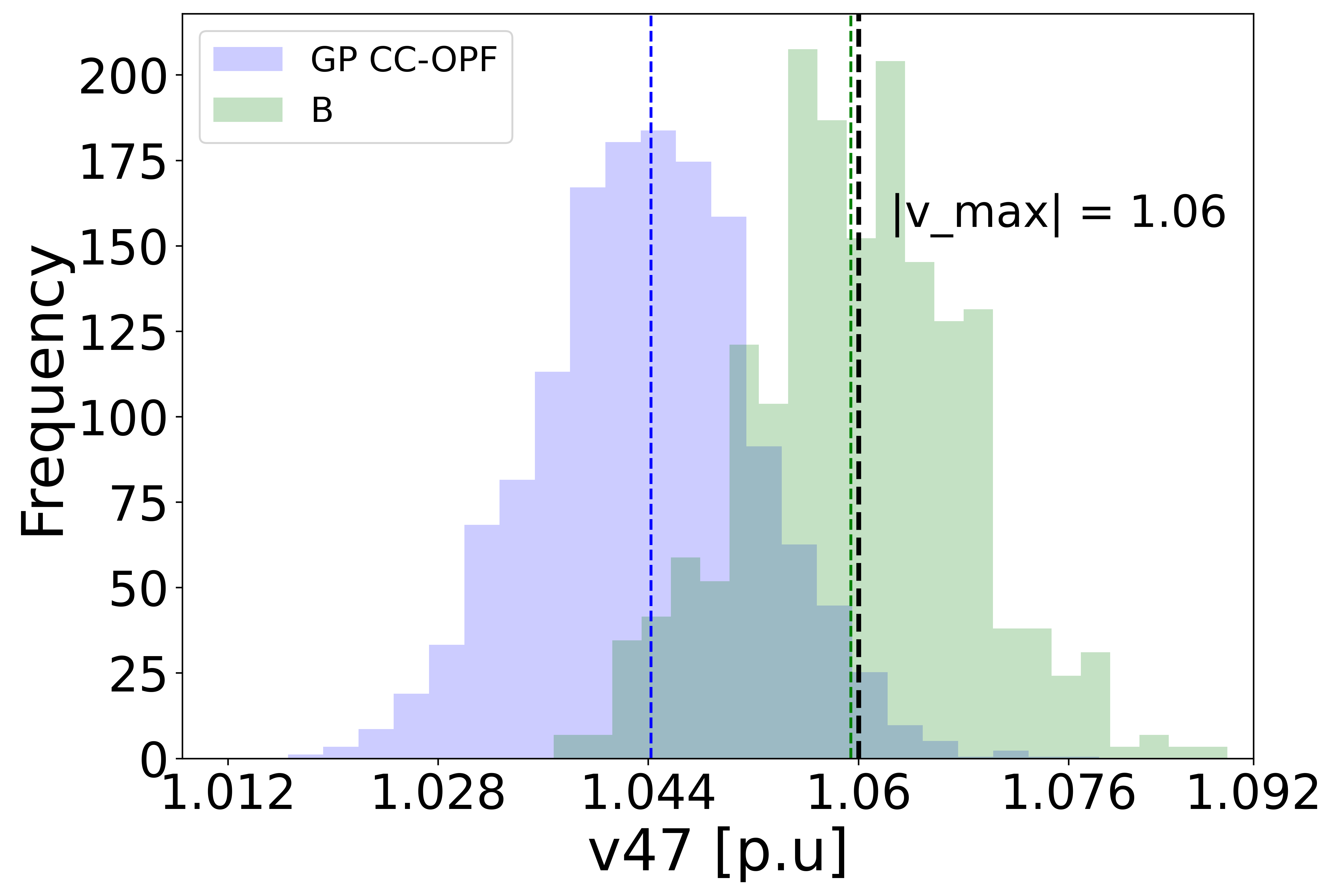}%
  \label{fig:histcase118a}%
  }%
 \hfill%
 \subfloat[The spread of voltages for scenario CC-OPF and (A) AC-OPF with full recourse.]{%
\includegraphics[width=0.48\textwidth,height=4.1cm]{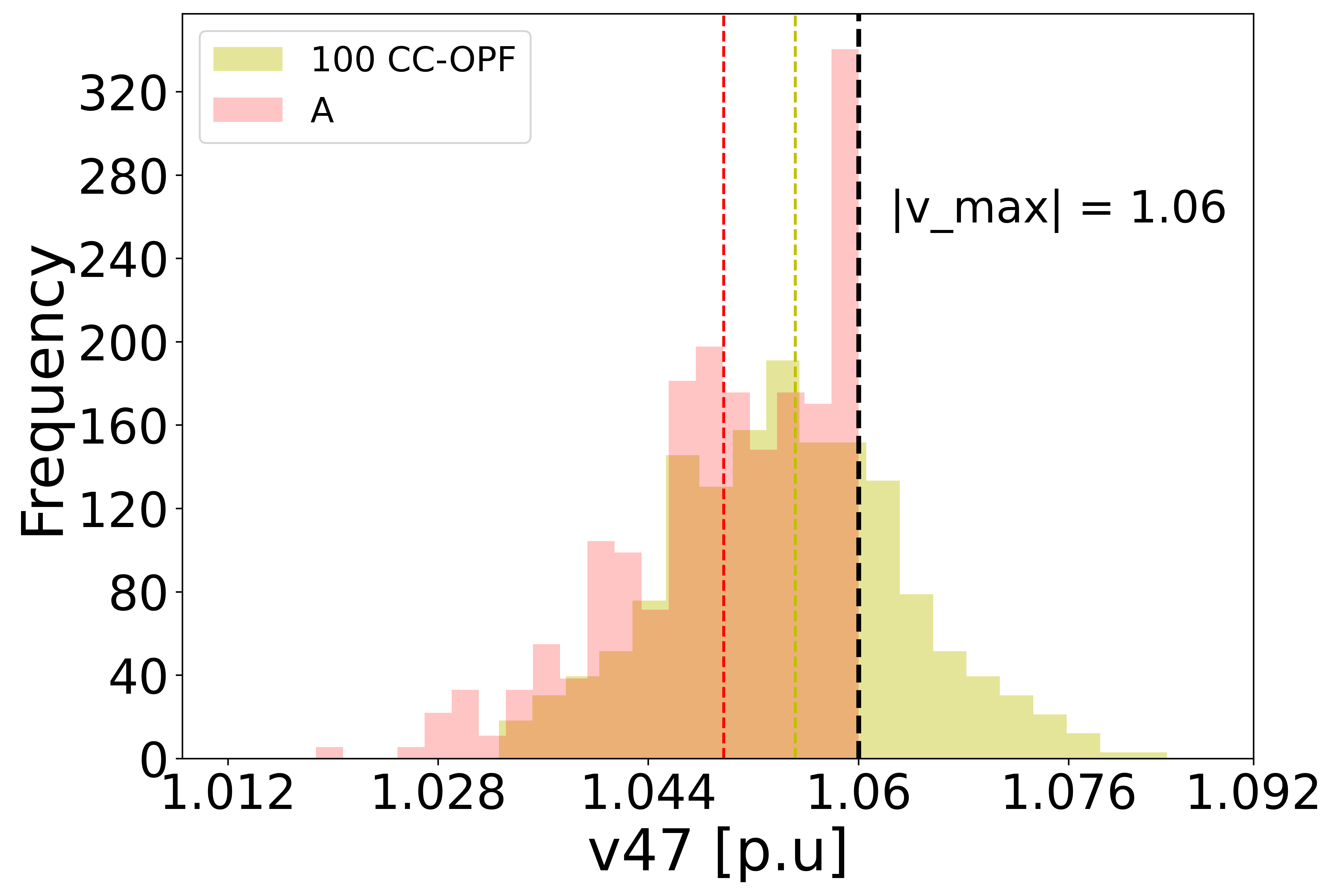}
  \label{fig:histcase118b}%
  }%
 \caption[IEEE118 empirical spread of voltages]{Case IEEE-118. Empirical Spread of voltages at Bus 47 under uncertainty.}
 \label{fig:histcase118}%
\end{figure}

\subsection{Discussion}\label{sec:40-Discussion}
Comparing the results of different approximation methods within GP CC-OPF, we can conclude that all three methods (TA1, TA2, and EM) approximate uncertainty margins very well with small probability violations for MC test samples (Figs.~\ref{fig:waveforms},~\ref{fig:example_39}). Table \ref{table:table2} shows that with the system scaling, the accuracy (RMSE) of the mean performance decreases. It can be explained by the fact that for the IEEE 39 and IEEE 118 bus systems, we considered 200 and 400 training data samples randomly sampled in space. More samples and active learning techniques can help improving the accuracy of large systems. The presented new approach for solving the CC-OPF problem could be equally efficiently applicable for power networks of essentially larger size than the test networks used for the algorithm testing. In general, the problem of GPR is scalability, where the computational complexity of training and prediction increases with the number of samples and input/output dimensions. Due to the inclusion of the second derivative term, TA2 and EM are more complicated than TA1 as confirmed by the CPU time results in Table \ref{table:table2}. Computational complexity problems can be overcome by using sparse GPR and dimensionality reduction techniques \cite{hewing2019cautious}, which we plan to explore in the future. Generally, from the obtained results that the TA1 method gives similar performance accuracy as TA2 and EM, but with less complexity.

\section{Conclusion} \label{ch4_conclusion}
In this chapter, we have introduced a novel stochastic OPF method that is model-free and data-driven. The key idea is to replace the standard AC power flow equation and security constraints with a Gaussian process regression model. We have shown that even with a relatively small number of samples, the GP function can accurately model the AC-PF equations and efficiently propagate uncertainty from input to output.

In our experiments, we have demonstrated that applying the first Taylor Approximation method in GP CC-OPF outperforms other proposed approximation methods and two widely used sample-based approaches. Despite the notable accuracy of the data-driven CC-OPF, it is limited by the propagation space of the training data. To address this issue, we propose a new hybrid GP CC-OPF method in the next \textbf{Chapter \ref{ch:5}} that combines linear and non-linear approximation methods. Additionally, we propose improving the scalability of the proposed approach using sparse GPR formulations.

 \chapter{Data Driven CC-OPF using Hybrid Sparse Gaussian Process}
\label{ch:5}
\section{Introduction} \label{ch5_intro}
In this chapter, we introduce an improved version of the data-driven stochastic OPF method proposed in \textbf{Chapter~\ref{ch:4}}. This improved approach is a hybrid method that combines the linear DC-PF balance equation with the GPR-based estimation of the residuals between DC and AC power flow equations. By incorporating the DC power flow equation, our method is more robust and less sensitive to the propagation space of the training data. To further improve the scalability of the method, we use sparse techniques for the GPR kernel matrix. For the input uncertainty propagation, the first-order Taylor Approximation (TA1) is applied. As in \textbf{Chapter~\ref{ch:4}}, the proposed approach has been evaluated on  IEEE9, 39 and 118 bus systems. The approach and results of this chapter have been presented and published at the \textit{IEEE Belgrade PowerTech 2023} conference~\cite{mile2}.

The remaining parts of this chapter are organized as follows: In Section~\ref{ch5_motivation}, we explain the reasons that motivated us to develop a hybrid approach for GP CC-OPF. Section~\ref{ch5_hyb_approach} presents the mathematical formulation of the proposed hybrid approach. The experimental results that demonstrate the effectiveness of the proposed method are presented in Section~\ref{ch5_results}. Finally, in Section~\ref{ch5_conclusion}, we summarize the key findings of this study.
\section{Motivation} \label{ch5_motivation}

In \textbf{Chapter~\ref{ch:4}}, we introduce a novel data-driven GP CC-OPF approach that is based on the full AC-PF equations. However, this approach may not always be robust and its effectiveness heavily depends on the distribution and spread of training data in the space. If the training data do not cover the space properly, the GPR model may tend towards the prior mean, which is zero, resulting in an inadequate approximation of the AC-PF function. To illustrate this drawback, we provide a toy example in Fig.~\ref{fig:hybr_image}
\begin{figure}[!t] 
 \centering
 \subfloat[Full approach: GP learned on historical data;]{
\includegraphics[width=0.48\textwidth,height=5.5cm]{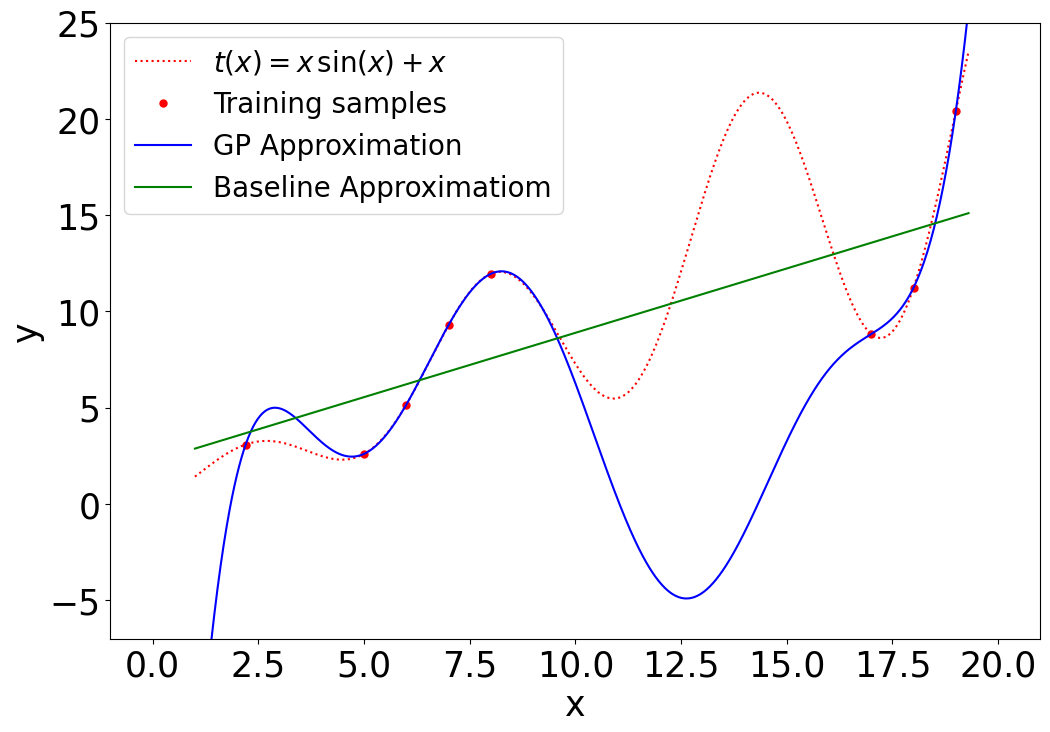}%
  \label{fig:hybr_image_a}%
  }
 \hfill%
 \subfloat[Hybrid approach: GP learned on  residuals between the linear model and data.]{ \includegraphics[width=0.48\textwidth,height=5.5cm]{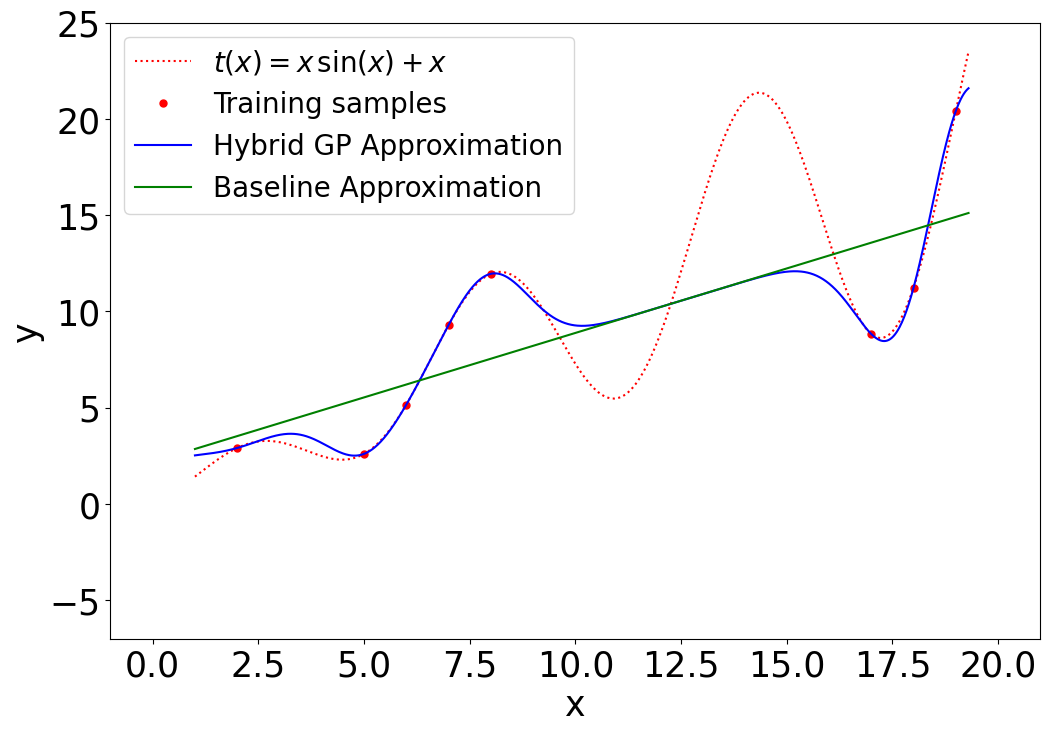}
 \label{fig:hybr_image_b}} 
 \caption[An example of illustrating the advantages of a hybrid approach]{Non-linear function $x \sin(x) + x$ (in red) is approximated with GP (blue; left) based on several observations (red dots). The hybrid GP approximation (blue; right) uses a linear proxy (green) to improve the approximation.}
 \label{fig:hybr_image}%
\end{figure}

The disadvantage of the full approach is that GP is learned from historical data on a full non-linear AC model using the entire synthetic data space. This means that this model will make a significant prediction error if the historical data is poorly distributed in space due to
the technical nature of the system. A simple one-dimensional example can be observed in Fig.~\ref{fig:hybr_image_a}. One can see the difference in regression results between $x=10$ and $x=17.5$ for this method. A simple GPR will tend to predict zero prior in this region since the absence of training data. To overcome this problem and build a more robust approach, we propose a hybrid chance constraints GP-based structure. Whereas hybrid GPR has non-zero prior in this region -- linear approximation -- and regression result is more adequate.

The advantage of the hybrid approach can be demonstrated with a simple one-dimensional example in Fig.~\ref{fig:hybr_image_b}. Consider a non-linear function $t(x) = x sin(x) + x$ (red dashed line) and a linearized function $h(x) = ax + b$ (green solid line - baseline approximation), where coefficients $(a,b)$ are fitted according to the training data corresponding to the non-linear toy function $t$. Fig.~\ref{fig:hybr_image_a} presents the dependence learned by GPR (blue solid line) from the full non-linear function, while in Fig.~\ref{fig:hybr_image_b}) the hybrid approach of the summation of the linearized function and the additive GP function learned on the residuals between $t(x)$ and $h(x)$ is shown. It can be seen from Fig.~\ref{fig:hybr_image_a} that when training data does not cover the function domain well enough, the GP function will tend toward the prior mean equal to zero. In the case of a hybrid approach, the function will tend to the values of the linearized function. Therefore, the worst-case output results will not be equal to zero, but to the linearized model outputs. Thus, a more robust model is proposed, in which the efficiency will not be so strongly violated in the case that the training set is not fully presented. In the case of the hybrid GP chance-constrained AC-OPF, the worst-case output results will be equal to those obtained with the approximated DC CC-OPF.

The hybrid framework combines a linear DC approximated physical model and an additive GP-part learned on the residuals between a non-linear AC and a linear DC power flow model. Thus, the hybrid approach to the AC power flow equation is modeled as follows:
\begin{equation}\label{eq:hyb_1}
y^{out} = z^{out}(x^{in}_i) + (g^{out}(x^{in}_i) + \omega_i), ~~~~~ i=1,...,m_s
\end{equation}
where $z^{out}$ is a known linear part and $g^{out}$ is an additive GP-term that describes unknown non-linearities of the system that need to be learned from historical data. According to~\ref{eq:hyb_1}, a residual output $r^{out}_i$ is equal to:
\begin{equation}\label{eq:hyb_2}
r^{out}_i = y^{out} - z^{out}(x^{in}_i) = g^{out}(x^{in}_i) + \omega_i, ~~~~~ i=1,...,m_s
\end{equation}

Even though the hybrid approach is a more robust method, it may still face computational complexity issues due to the GP kernel matrix. To address this challenge, we introduce a sparse GP approximation to enhance computational efficiency while maintaining the robustness and accuracy of the method.

\section{Hybrid approach} \label{ch5_hyb_approach}

As previously discussed in Section~\ref{ch4_approach}, utilizing Gaussian processes to model power flow equations with uncertainty offers a more tractable solution. Unlike the full approach, our proposed hybrid approach replaces the power flow balance equation with a combination of a linear approximation  and an additive GPR model learned on the residuals:
\begin{subequations}\label{hyb_balance}
\begin{align}
    \mu = z^{out}(\mu_{x^{in}_*}) + \mu(x^{in}_*) ~~~~~~~~~~\\
    \sigma^2 = diag(\Sigma_{z^{out}}(\Sigma_{x^{in}_*})) + diag(\Sigma(x^{in}_*))
\end{align}
\end{subequations}
where $z^{out}(\mu_{x^{in}_*}) = [v_{dc}, A\mu^T_{x^{in}_*} + b]$; $v_{dc} = 1~p.u.$ is constant DC output voltage, $A$ is linear coefficient matrix and $b$ is intercept vector; $diag(\Sigma_{z^{out}}(\Sigma_{x^{in}_*}))$ and $diag(\Sigma(x^{in}_*))$ are column vectors of diagonal matrix elements where 
$$\Sigma_{z^{out}} = \begin{bmatrix} 0 & 0\\ 0 & A \Sigma_{x^{in}_*} A^T \end{bmatrix}$$

Applying the GP-balance equation \ref{hyb_balance}, an analytical reformulation of chance-constraints with uncertainty margins   Eq.~\ref{eq:obj}-\ref{eq:pr_u_1} derived in Section~\ref{ch4_approach}, the hybrid GP CC-OPF problem is:
\begin{subequations}\label{eq:hyb_approach}
\begin{align}
 &\min_{p_g, \alpha, \mu, \sigma} \sum_{i\in \mathcal{G}} \{ c_{2,i}(p_{g,i}^{2} + tr(\Sigma_{\omega})\alpha_{i}^{2}) + c_{1,i}p_{g,i} + c_{0,i}\} \label{eq:Gp-cost_hyb}\\
 &\text{s.t.~} \sum_{i\in \mathcal{G}} \alpha_i = 1, ~ \alpha_i \geq 0 \\
 &~~~~~ \sum_{i\in \mathcal{G}} p_{g,i} = \sum_{i\in \mathcal{L}} p_{l_i} - \sum_{j\in \mathcal{L}} p_{rs_j} \\
 &~~~~~~  \mu_i = z^{out}(\mu_{x^{in}_*}) + \mu(x^{in}_*)  \label{eq:27d}\\ 
 &~~~~~~ \sigma^2_i = diag(\Sigma_{z^{out}}(\Sigma_{x^{in}_*})) + diag(\Sigma(x^{in}_*))  \label{eq:27e}\\
 &~~~~~~ y_i^{out_{min}} + \lambda_{i} \leq \mu_{i} \leq y_z^{out_{max}} - \lambda_{i}\\
 &~~~~~~ p_{g,i}^{min} + \lambda_{p_{g,i}} \leq p_{g,i} \leq p_{g,i}^{max} - \lambda_{p_{g,i}} 
\end{align}
\end{subequations}
where $\lambda_{i} = r_{i} \sqrt{\sigma^2_{i}}$ for $i\in n_y$ and $\lambda_{p_{g,i}} = r_{p_{g,i}} \alpha_i\, \sqrt{tr(\Sigma_{\omega})}$ are uncertainty margins. 

Similar to the full approach, the non-linear optimization framework CasADi is used to solve the hybrid GP CC-OPF formulation in Eq.~\ref{eq:hyb_approach} with the available IPOPT solver.

\subsection{Input Approximation as Normal Distributions}
In the GPR model, for the given deterministic inputs, the predicted outputs have a normal (Gaussian) distribution described with mean and variance (Eq.~\ref{eq:gp_output_1}). Since we consider a stochastic problem, the inputs are also distributed. Evaluating the outputs of a GPR from an input distribution is generally intractable and the resulting outputs' distribution is not normal. For certain assumptions of the input distribution, some strict over-approximations of the resulting distributions exist \cite{koller2018learning}, but they are computationally demanding and very conservative. Accordingly, we consider computationally cheap and a practical approximation where the control and uncertain inputs are approximated as jointly Gaussian distributed:
\begin{equation}\label{eq:19}
    x^{in}_* \sim \mathcal{N}(\mu_{x^{in}_*}, \Sigma_{x^{in}_*}) = \mathcal{N} \left(\begin{bmatrix}
 p_g\\
 p_{\omega}
 \end{bmatrix}, 
 \begin{bmatrix}
 \Sigma_g & \Sigma_{g\omega}\\\
 \Sigma_{g\omega}^T & \Sigma_{\omega}
 \end{bmatrix}\right)\\
\end{equation}
where $p_g$, $p_{\omega} = [p^T_l, p^T_{rs}]^T$ are vectors of optimized control variables and forecasted operating points; $\Sigma_g \in \mathbb{R}^{n_{\mathcal{G}} \mathsf{x} n_{\mathcal{G}}}$, $\Sigma_{\omega} \in \mathbb{R}^{n_{\mathcal{L}} \mathsf{x} n_{\mathcal{L}}}$ and $\Sigma_{g\omega} \in \mathbb{R}^{n_{\mathcal{G}} \mathsf{x} n_{\mathcal{L}}}$ are input covariance sub-matrices. 

The covariance matrices $\Sigma_{\omega}$, $\Sigma_g$ and $\Sigma_{g \omega}$ are equal to the equations~\ref{cov_1},~\ref{cov_2}, and~\ref{cov_3}, respectively.
\section{Experimental Setup and Results} \label{ch5_results}

The simulations in this chapter are performed on an Intel Core i7-5500U CPU @ 2.40GHz and 8GB of RAM.  The python \textit{hybrid-gp}\footnote{\url{https://github.com/mile888/hybrid_gp}} framework was developed to simulate the investigation and \textit{pandapower} \cite{thurner2018pandapower} package was used to validate the results.

\subsection{Case Study}
The performance and scalability of the proposed hybrid GP CC-OPF approach are evaluated on 9, 39 and 118 bus IEEE test systems. All systems are provided with \texttt{pandapower}~\cite{thurner2018pandapower}.

\textit{IEEE 9-bus Test Systems} consists of 3 generators and 3 loads on the high-voltage level of $V_{n} = 345kV$. We introduce two renewable source generators at buses 3 and 5 with a total forecast power output of 80 MW, which is approximately 25.4\% of the total active power demand. We assume that the reactive to real power ratio is 0.3. All loads and renewable sources are uncertain.

\textit{IEEE 39-bus Test Systems} consists of 10 generators and 21 loads with $V_{n} = 345kV$. Renewable source generators are placed at buses 1, 11, 14, 21, 23, and 28 with a total forecast power output of 1260 MW, which is approximately 20.2\% of the total active power demand. The assumed power ratio is 0.3. Also, we consider that all loads and renewable sources are assumed to be uncertain.

\textit{IEEE 118-bus Test Systems} consists of 54 generators and 99 loads with $V_{n} = 345kV$. The six renewable source generators are placed at buses 10, 27, 47, 51, 78, and 92 with a total forecast power output of 1260 MW, which is approximately 29.7\% of the total active power demand. All loads and RES are considered as uncertain elements with an assumed reactive to real power ratio of $0.3$.

The forecast errors are modeled as zero mean, multivariate Gaussian random variables with a standard deviation corresponding to 15\% of the forecasted loads ($\sigma_l = 0.15\, p_l$) and 30\% of the forecasted renewable sources ($\sigma_{rs} = 0.3\, p_{rs}$). The acceptable violation probabilities $\epsilon$ are set to $\epsilon_{p_g}=0.1\%$ and $\epsilon = 2.5\%$. We assume a very small percentage of violations ($\epsilon_{p_g}$) according to \cite{abbaspourtorbati2015swiss}, since generation constraints are physically impossible to violate. However, output constraints are soft constraints where some violations can be tolerated ($\epsilon_y$) if the magnitude and duration are not too large, or removed through additional control actions such as generation re-dispatch~\cite{roald2017chance}.

\subsection{Model Performance}
The hybrid GPR model is trained on residuals between datasets collected from full AC-OPF and DC-OPF approximation. The synthetic data of the  IEEE 9-bus system consists of 8 inputs features and 15 outputs. The IEEE39 model has 37 inputs and 74 outputs, while the GPR model of the IEEE 118-bus system has 159 inputs and 291 outputs. We generated 75, 200 and 400 random training samples for IEEE 9, 39 and 118-bus systems and considered a different number of inducing points of 10, 30, and 50 for both full and hybrid GPR models. We use IPOPT solver to solve both hybrid and full GP CC-OPF and find a feasible solution with a convergence tolerance of $\epsilon_{tol} = 10^{-5}$. 

The results in Table~\ref{table:table_hybr_1} indicate that the hybrid GPR model with TA1 outperforms both RMSE and CPU time with respect to the full GPR model with TA1 in \textbf{Chapter~\ref{ch:4}}. Similarly, sparse approximation allows to significant reduce the time complexity for both hybrid and full approaches. The power of the sparse approximation is presented for all cases. Decreasing the number of inducing points reduces the computation time in all test cases (IEEE9, IEEE39 and IEEE118) and approaches (full and hybrid), but increases RMSE. Accordingly, we can find a trade-off between the computation time and the accuracy of the faster and more robust hybrid GP approach. To determine the level of error of the mean output values in prediction, the RMSE of the output data from the analytical DC approximated CC-OPF was considered as the worst-case error. For the given same inputs, the RMSE of the CC DC-OPF is RMSE = $6.65e^{-2}~p.u.$ of IEEE9, RMSE = $3.04e^{-2}~p.u.$ of IEEE39, and RMSE = $5.84e^{-1}~p.u.$ of IEEE118. In other words, applying the GPR model with a few samples leads to at least 6 times more accurate approximation than the DC baseline. 

Therefore, our proposed hybrid GP approach in both cases gives a smaller error than the well-known DC approximation. The comparison of the results are summarized in Table~\ref{table:table_hybr_1}.
\begin{table}[htbp]
\centering
 \begin{tabular}{c|c|c|c|c|c|c}
 \hline
 {\bf Test Case} &
 \multicolumn{2}{c}{\bf IEEE 9} \vline & \multicolumn{2}{c}{\bf IEEE 39} \vline &
 \multicolumn{2}{c}{\bf IEEE 118} \\
 \hline
 Parameters & RMSE & Time & 
 RMSE & Time & RMSE & Time\\
 & $10^{-2}$ p.u. & sec. & $10^{-2}$ p.u. & sec. & $10^{-2}$ p.u. & sec.\\
 \hline
 hybrid TA1 & \textbf{0.30} & 0.24 & \textbf{0.53} & 10.28 & \textbf{21.4} & 162.4\\
 \hline
 hybrid-sparse 50 TA1 & 0.56 & 0.10 & 1.12 & 2.93 & 23.2 & 29.3 \\
 \hline
 hybrid-sparse 30 TA1 & 0.79 & 0.06 & 1.22 & 1.23 & 31.4 & 14.5 \\
 \hline
 hybrid-sparse 10 TA1 & 1.22 & \textbf{0.02} & 1.41 & \textbf{0.51} & 39.9 & \textbf{7.4}\\
 \hline
 full TA1 & 0.63 & 0.35 & 1.37 & 19.4 & 83.7 & 212.1 \\
 \hline
 full-sparse 50 TA1 & 0.97 & 0.17 & 2.69 & 4.12 & 87.8 & 41.1\\
 \hline
 full-sparse 30 TA1 & 1.67 & 0.13 & 2.80 & 1.42 & 93.1 & 24.1 \\
 \hline
 full-sparse 10 TA1 & 2.43 & 0.03 & 3.25 & \textbf{0.51} & 99.7 & 10.2\\
 \hline 
 \end{tabular}
 \caption[Results of the full and hybrid GP CC-OPF approaches]{GP CC-OPF approximation method results for IEEE 9, IEEE 39 and IEEE 118 bus systems.} 
 \label{table:table_hybr_1}
\end{table}

To support readers in better understanding the results presented in Table~\ref{table:table_hybr_1}, we provide graphical illustrations for all systems in Fig.~\ref{fig:IEEE9_RMS_CPU},~\ref{fig:IEEE39_RMS_CPU} and~\ref{fig:IEEE118_RMS_CPU} .

\begin{figure}[!t]
 \centering
 \subfloat[\centering RMSE IEEE9]{{\includegraphics[width=9.4cm]{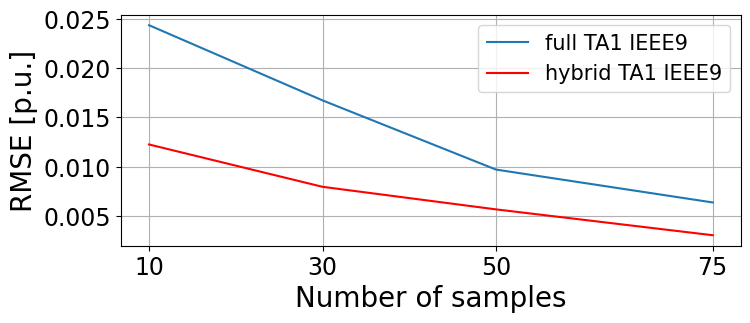} }}%
 \qquad
 \subfloat[\centering CPU time IEEE9]{{\includegraphics[width=9.4cm]{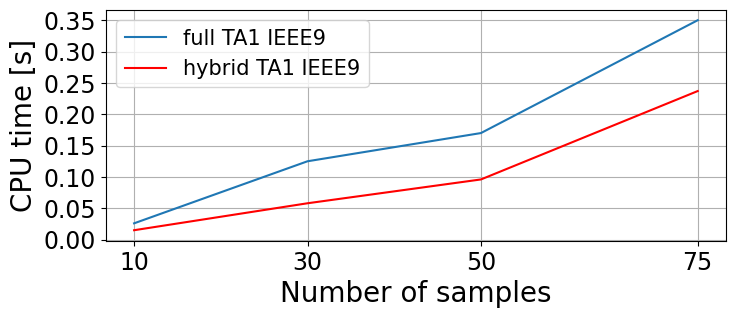} }}%
 \caption[IEEE9 - RMSE and CPU time results]{Ilustration of the RMSE and CPU time results for the IEEE9 bus system.}%
 \label{fig:IEEE9_RMS_CPU}%
\end{figure}

\begin{figure}[!t]
 \centering
 \subfloat[\centering RMSE IEEE39]{{\includegraphics[width=9.4cm]{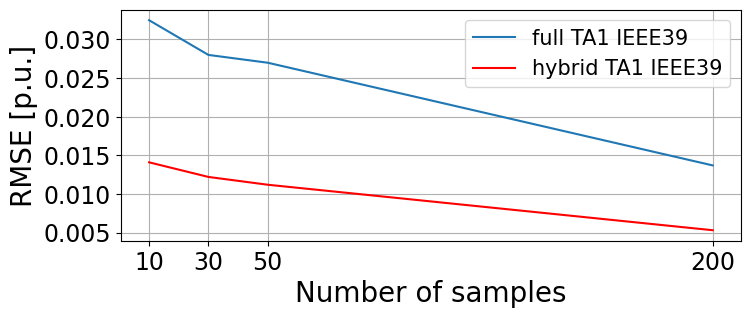} }}%
 \qquad
 \subfloat[\centering CPU time IEEE39]{{\includegraphics[width=9.4cm]{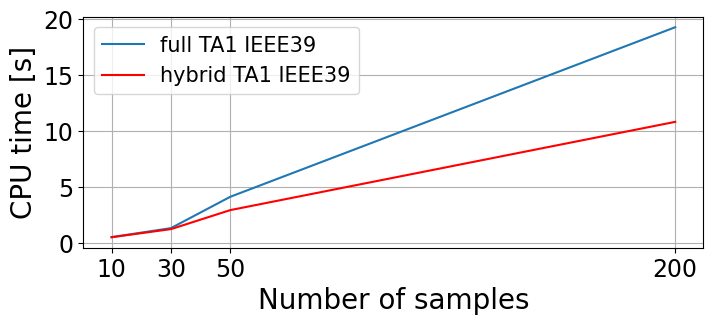} }}%
 \caption[IEEE39 - RMSE and CPU time results]{Ilustration of the RMSE and CPU time results for the IEEE39 bus system.}%
 \label{fig:IEEE39_RMS_CPU}%
\end{figure}

\begin{figure}[!t]
 \centering
 \subfloat[\centering RMSE IEEE118]{{\includegraphics[width=9.4cm]{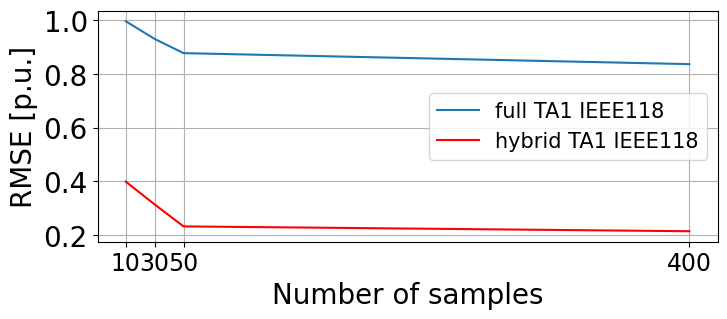} }}%
 \qquad
 \subfloat[\centering CPU time IEEE118]{{\includegraphics[width=9.4cm]{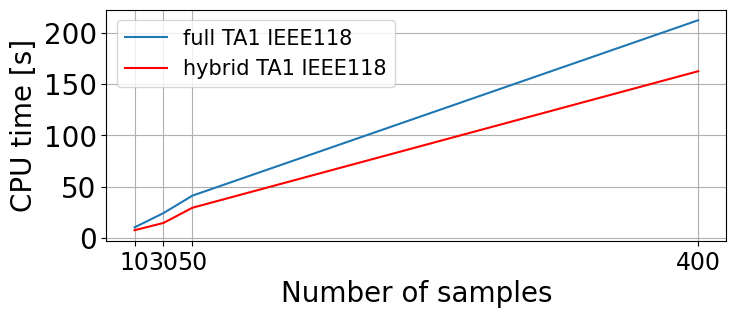} }}%
 \caption[IEEE118 - RMSE and CPU time results]{Ilustration of the RMSE and CPU time results for the IEEE118 bus system.}%
 \label{fig:IEEE118_RMS_CPU}%
\end{figure}

\subsection{Comparison of GP CC-OPF to Sample-Based Methods}~\label{comp_res}
In this section, we compare optimal objective function value, computational time, and reliability of the proposed hybrid and full GP CC-OPF approaches with Scenario Approximation (SA) CC-OPF \cite{mezghani2020stochastic}. In all cases, we use the first-order Taylor approximation (TA1) as a function of linearization. The empirical constraint violation is computed using 1000 Monte-Carlo samples that follow (Gaussian) uncertainty distribution. 

We compare our approach to two baselines. (A) Deterministic AC-OPF problem for each uncertainty realization and taking a corresponding quantile in the output distribution. (B) A deterministic AC-OPF for the mean load and RES, with fixed and equal participation factor $\alpha^*$. Notice that (A) is unrealistic as it solves AC-OPF after uncertainty realization (full-resource), while (B) does not optimize the participation factor $\alpha$. Thus (A) gives the upper (though unrealistic) bound and (B) gives the lower bound on the feasible solution value. Numerical evaluation of the bounds, given in Table~\ref{table:table_hybr_2}, shows that the hybrid-sparse GP modeling provides a flexible scheme that can improve computational effort without sacrificing accuracy.

\begin{table}[!t]
\centering
\begin{tabular}{l|l|l|l}
 \hline\hline
 System &
 \multicolumn{3}{c}{IEEE 9} \\\cline{2-4}
 \hline
 Parameters & Cost [\$] & Infeas. Prob. [\%] & Time (s)\\
 \hline
 A (full recourse)& $4.056\cdot 10^3$ & - & 0.86\\
 \hline
 B (base-case)& $3.467 \cdot 10^3$ & 9.76 & 0.86 \\
 \hline
 full GP CC-OPF & $4.039 \cdot 10^3$ & 0.44 & 0.35\\
 \hline
 \textbf{hybrid GP CC-OPF} & $4.024\cdot 10^3$ & 0.80 & 0.12 \\
 \hline
 20 CC-OPF & $3.481 \cdot 10^3$ & 7.52 & 4.4\\
 \hline
 50 CC-OPF & $3.836 \cdot 10^3$ & 4.92 & 12.1\\
 \hline
 100 CC-OPF & $3.985 \cdot 10^3$ & 1.24 & 22.8\\
 \hline\hline
 \end{tabular}
 \begin{tabular}{l|l|l|l}
 \hline\hline
 System &
 \multicolumn{3}{c}{IEEE 39}\\\cline{2-4}
 \hline
 Parameters & Cost [\$] & Infeas. Prob. [\%] & Time (s)\\
 \hline
 A (full recourse) & $7.869\cdot 10^6$ & - & 0.93\\
 \hline
 B (base case)& $7.533 \cdot 10^6$ & 35.36 & 0.93 \\
 \hline
 full GP CC-OPF & $7.752\cdot 10^6$ & 2.36 & 19.38 \\
 \hline
 \textbf{hybrid GP CC-OPF} & $7.752\cdot 10^6$ & 2.48 & 10.82 \\
 \hline
 50 CC-OPF & $7.642\cdot 10^6$ & 14.20 & 96.4 \\
 \hline
 100 CC-OPF & $7.696\cdot 10^6$ & 8.24 & 184.7 \\
 \hline
 200 CC-OPF& $7.813\cdot 10^6$ & 0.16 & 505.0 \\
 \hline\hline
 \end{tabular}
 \begin{tabular}{l|l|l|l}
 \hline\hline
 System &
 \multicolumn{3}{c}{IEEE 118}\\\cline{2-4}
 \hline
 Parameters & Cost [\$] & Infeas. Prob. [\%] & Time (s)\\
 \hline
 A (full recourse) & $26.91\cdot 10^6$ & - & 1.59\\
 \hline
 B (base case)& $26.39 \cdot 10^6$ & 49.97 & 1.59 \\
 \hline
 full GP CC-OPF & $26.91\cdot 10^6$ & 2.47 & 212.1 \\
  \hline
 \textbf{hybrid GP CC-OPF} & $26.91\cdot 10^6$ & 2.49 & 162.4 \\
 \hline
 100 CC-OPF & $26.52\cdot 10^6$ & 28.12 & 683.3 \\
 \hline
 200 CC-OPF & $26.71\cdot 10^6$ & 19.62 & 1856.5 \\
 \hline
 500 CC-OPF& $26.90\cdot 10^6$ & 4.78 & 39537.1 \\
 \hline\hline
 \end{tabular}
 \caption[Cost functions and probability of violations]{Cost function values and probability of violation of a constraint at a solution}
 \label{table:table_hybr_2}
\end{table}

Results in Table~\ref{table:table_hybr_2} show that the proposed hybrid GP CC-OPF model outperforms sample-based chance constraints reformulation in the context of computational complexity. In all system cases, the proposed GP method needs an enormously lower CPU time to find a solution. Moreover, the hybrid approach is faster than GP CC-OPF based on full AC-PF from \textbf{Chapter~\ref{ch:4}}.

\subsection{Interpretation of results}

In this section, we will provide a visual interpretation of the results obtained from the previous section~\ref{comp_res}. We will consider all three parameters: cost, security (infeasible probability), and time, as shown in Table~\ref{table:table_hybr_2}. These parameters will be graded on a scale from 0 to 5, where 0 represents the worst case and 5 represents the best.

Fig.~\ref{fig:vis_compar} presents a visual comparison of all approaches in terms of the highlighted parameters. In each case (IEEE9, IEEE39, and IEEE118), the red horizontal lines represent the trade-off between security and cost. It is evident that both the hybrid and full approaches are capable of finding a suitable trade-off between security and cost. However, the hybrid approach exhibits faster performance.

\begin{figure}[!t]
 \centering
 \subfloat[\centering IEEE9]{{\includegraphics[width=11.8cm]{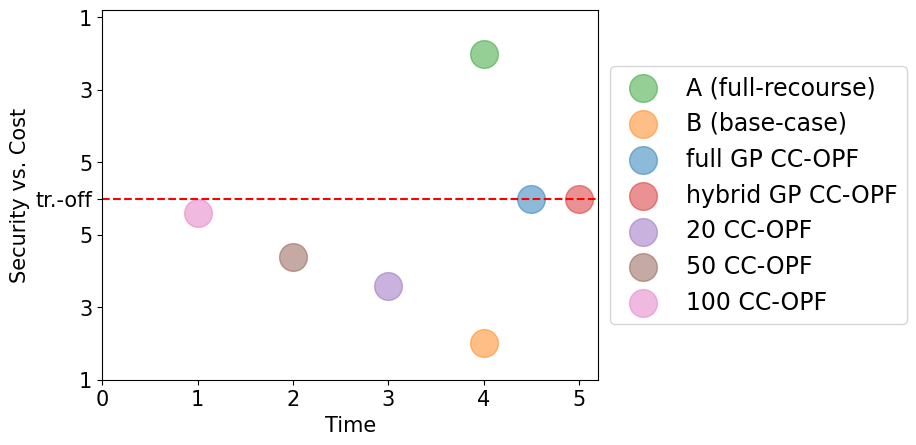} }}%
 \qquad
 \subfloat[\centering IEEE39]{{\includegraphics[width=11.8cm]{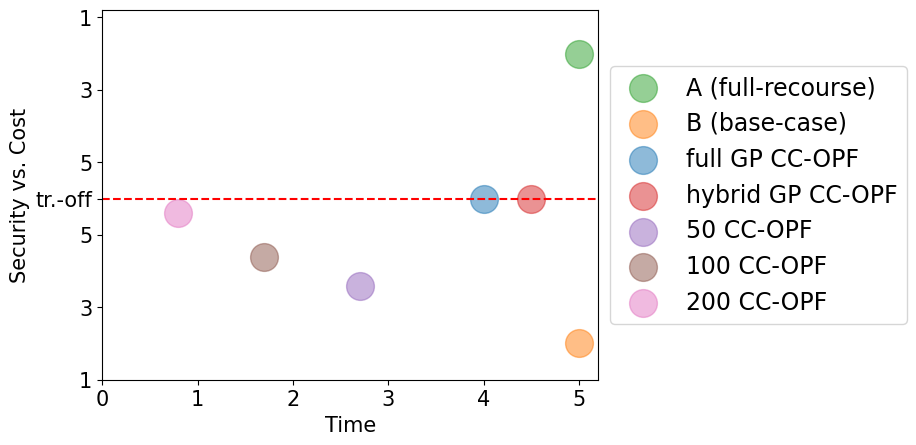} }}%
 \qquad
 \subfloat[\centering IEEE118]{{\includegraphics[width=11.8cm]{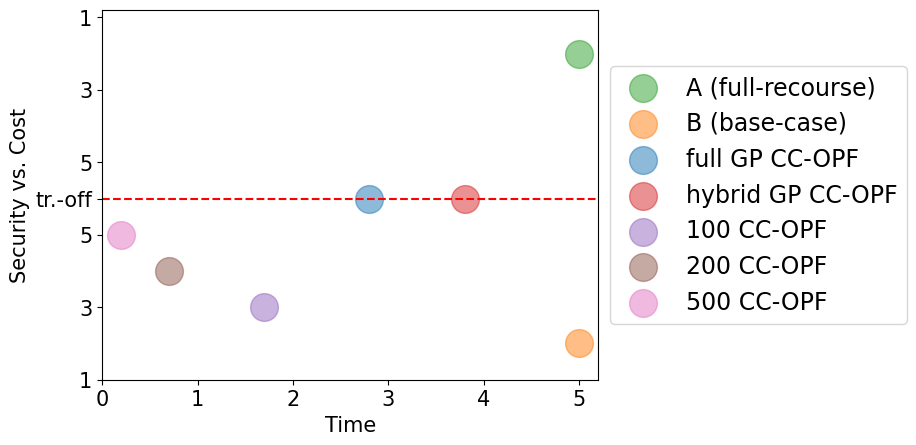} }}%
 \caption[Visual interpretation of the comparison of approaches]{Illustration of cost, security, and time comparison approaches for solving stochastic AC-OPF.}%
 \label{fig:vis_compar}%
\end{figure}

Furthermore, as the system size increases and the approaches become more scalable, the time required for both the hybrid and full approaches also increases compared to approaches A (full-resources) and B (base-case). Despite this increase in time, the trade-off between security and cost is maintained.

\subsection{Comparison of Uncertainty Margins}
Fig. \ref{fig:3} compares the uncertainty margins \cite{schmidli2016stochastic} calculated with the hybrid and full GP CC-OPF (red and blue), the Monte Carlo simulation of AC-OPF for base-case (B) (yellow and green) and the Monte Carlo simulation for CC-OPF solution (brown and dark salmon) with 100 (IEEE9), 200 (IEEE39) and 500 (IEEE118) scenarios. The bar plots represent an example of voltage, reactive power and apparent power flow uncertainty margins for each system. 

The three standard deviations (3 std) intervals for uncertainty margins are analytically derived using the estimated GP-variance (\ref{eq:27e}) around the estimated GP-mean (\ref{eq:27d}). The uncertainty margins from Monte Carlo are asymmetrical due to MC-based distributions owing to non-linear AC-PF. Thus, Monte Carlo uncertainty margins determine upper ($1-\epsilon$) = 99.73\% and lower ($\epsilon$) = 0.27\% quantiles (equal to 3 std) of the output distribution around the AC-PF solution $y(x)$, denoted by $y^{out}_{1-\epsilon}$ and $y^{out}_{\epsilon}$. Thus, the constraint tightenings are calculated by:
\begin{subequations}\label{eq:30}
\begin{align}
    \lambda^{upper} = y^{out}_{1-\epsilon} - y^{out}(x^{in}) \\
     \lambda^{lower} = y^{out}(x^{in}) - y_{\epsilon}^{out}
\end{align}
\end{subequations}

Compared to the voltage margins, the power margins are much larger in absolute terms. The GP CC-OPF margins of both approaches are symmetric, leading to larger or smaller margins compared with the upper and lower empirical quantiles obtained from the Monte Carlo AC-OPF and CC-OPF simulations in (\ref{eq:30}). This indicates that the GP uncertainty margins are well approximated. 

\begin{figure}[!t]
 \centering
 \subfloat[\centering IEEE9]{{\includegraphics[width=9.9cm]{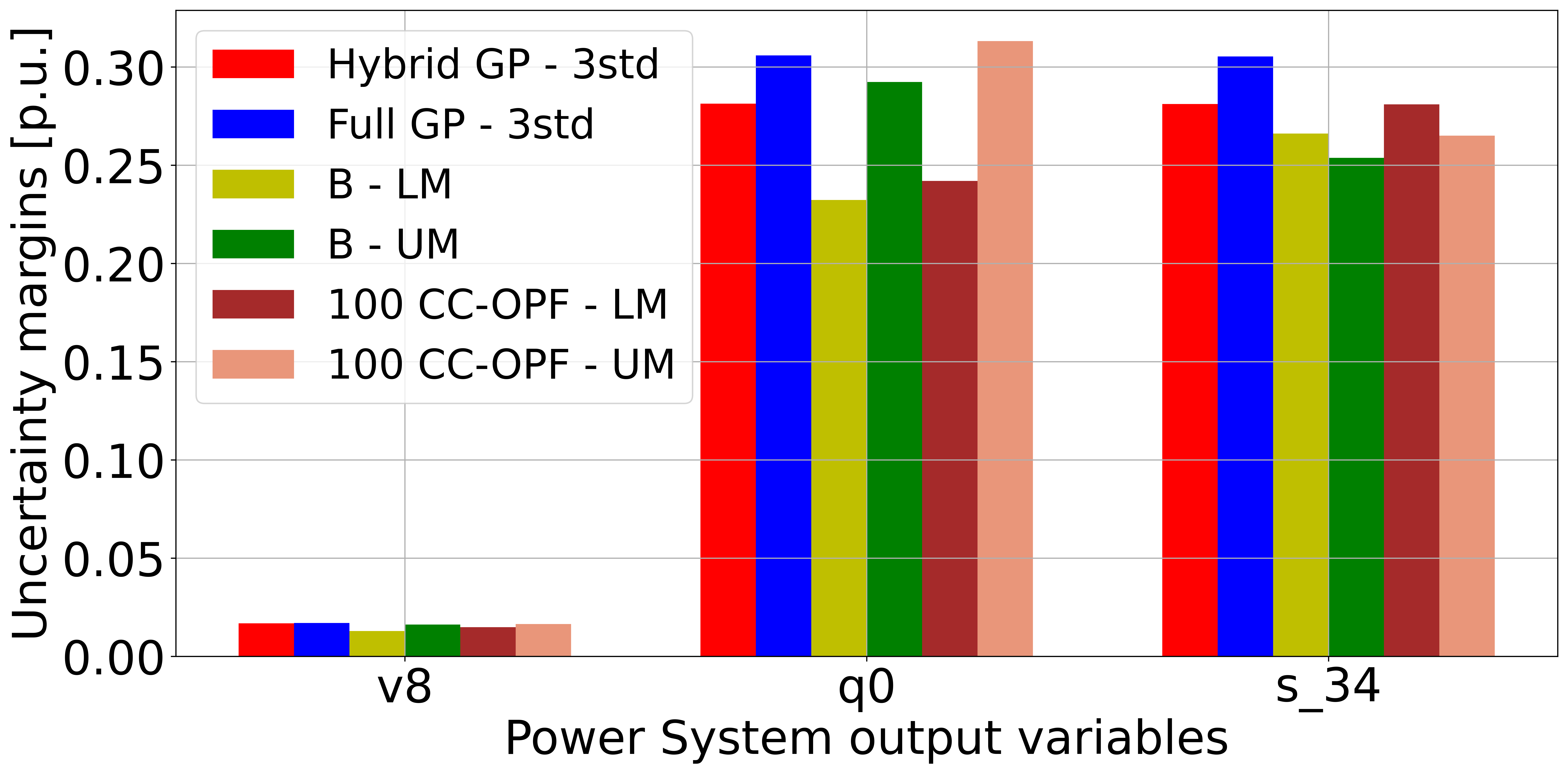} }}%
 \qquad
 \subfloat[\centering IEEE39]{{\includegraphics[width=9.9cm]{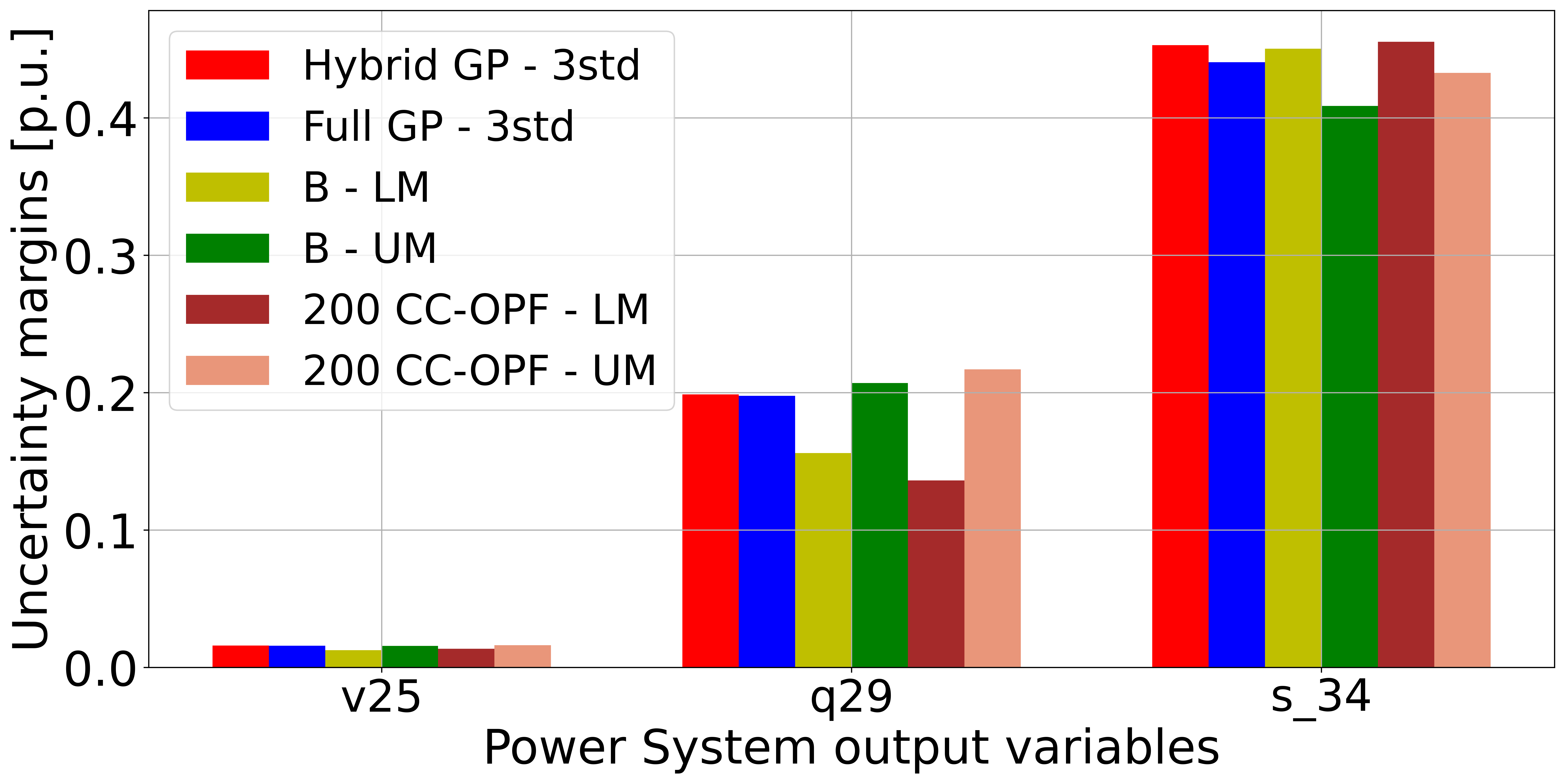} }}%
 \qquad
 \subfloat[\centering IEEE118]{{\includegraphics[width=9.9cm]{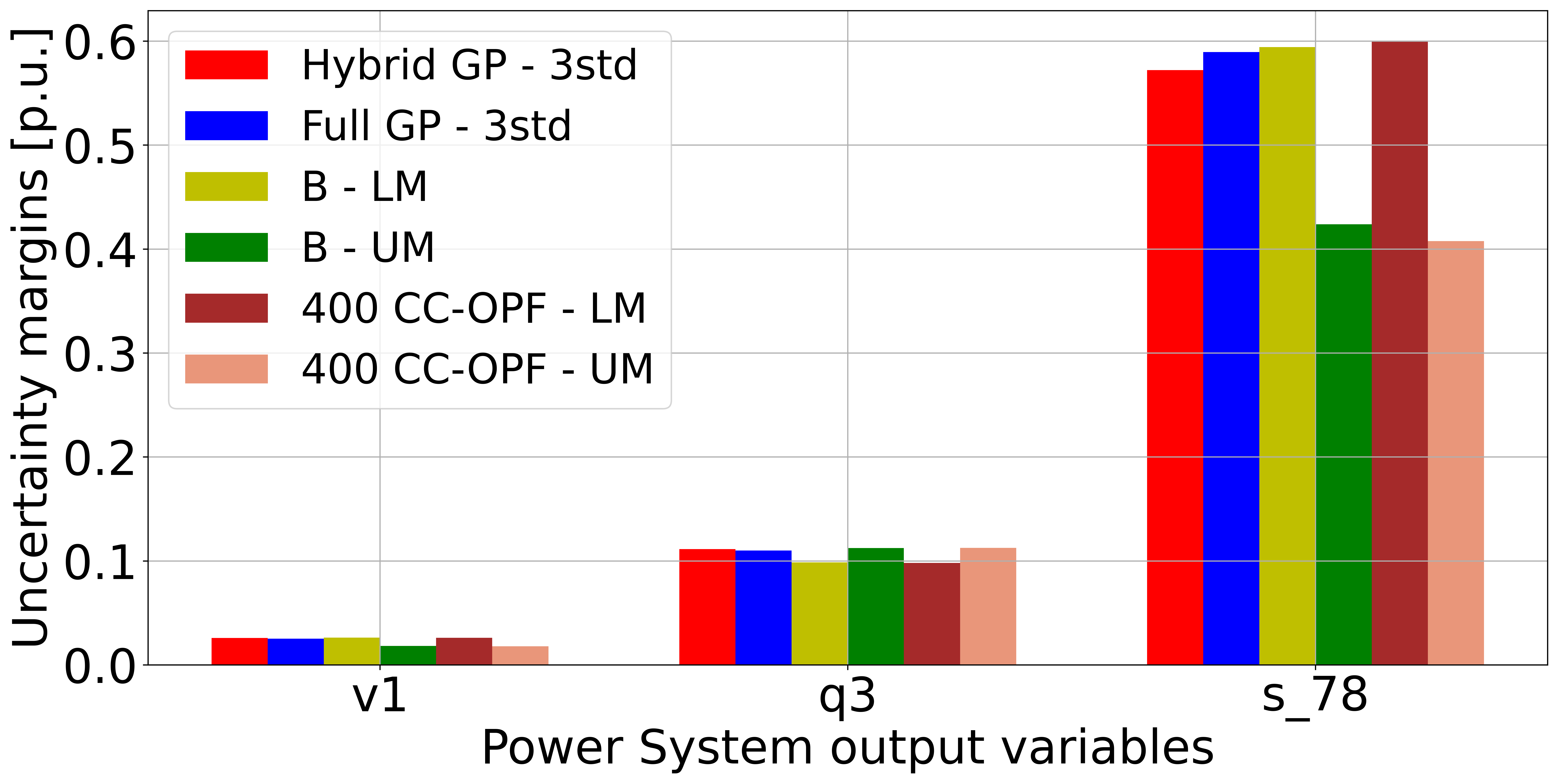} }}%
 \caption[Comparation of uncertainty margins]{Output uncertainty margins $\lambda$, for example, voltage, reactive and apparent power constraints. (LM - lower margin 0.27\% interval; UM - upper margin 99.73\% interval; B - deterministic AC-OPF for the mean load and RES, with fixed and equal participation factor $\alpha^{*}$).}%
 \label{fig:3}%
\end{figure}

\subsection{Analysis in the Case of Distribution Gaps}
In this section, we analyze the presence of gaps in distributions in the IEEE9 case. The purpose of the experiment is to test the robustness of the ML algorithm in case of corrupted data. The gaps in distributions may appear in case of an information attack, or poor data collection. A simple example of that was presented in Fig.~\ref{fig:hybr_image} -- the interval from, approximately, $7.5$ to $17$ has a lack of input data coverage which leads to poor performance of full GP. However, the hybrid GP, presented in this thesis handles such intervals better. We expect to observe a similar result in the case of power grids as well.

First, let us specify the experiment setup. The experiment consists of three major steps: corrupting the data, training on the corrupted data and testing the predictions of models. 
We compare two models: full GP from \textbf{Chapter~\ref{ch:4}} and the proposed hybrid GR model. The corruption step is conducted as follows: after generating the training dataset, we throw out a part of data, i.e., samples of pairs $(x^{in}_i, y^{out}_i)$, such that it would result in crossing out an interval of some buses' input data. Examples of such corruption are provided in Fig.~\ref{fig:ieee9_RMSE_corrupt}. 
Next, both full and hybrid GP are trained on the corrupted dataset. Finally, the RMSE metric is being evaluated on both regression models on the crossed-out regions.

We conduct the outlined experiment on case IEEE9, with the dataset generated as in Section~\ref{ch2_dataset}. The corruption of data is conducted on three buses separately, i.e., in each of the three experiments only one bus is a target for corruption. The load fluctuation data is crossed out at buses 4, 6 and 8. In each case, 15 MW interval has been thrown out. The histograms of load data are presented in Figure \ref{fig:ieee9_RMSE_corrupt}. The others' distributions have not experienced any significant change. Next, we trained ordinary and hybrid GPR on these data. The evaluated RMSE metrics are presented in Table \ref{table:ieee9_RMSE_corrupt}.

\begin{figure*}[th]
 \centering
 \subfloat{{\includegraphics[width=0.3\linewidth]{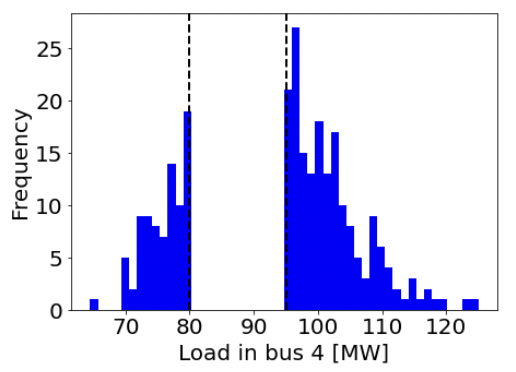} }}%
 \subfloat{{\includegraphics[width=0.3\linewidth]{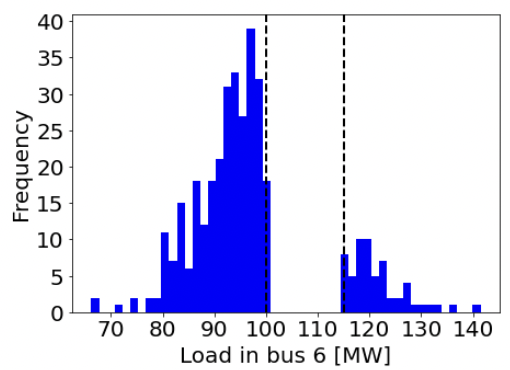} }}%
 \subfloat{{\includegraphics[width=0.3\linewidth]{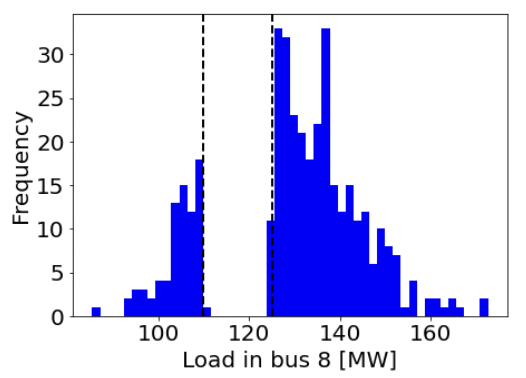} }}%
 \caption{An example of misspecified input, where a 15 MW interval around the mean bus load value is inaccessible. The latter may happen as a result of cyberattacks or meter failures. The proposed algorithm demonstrates stability over significant adversarial changes in input data.}
 \label{fig:ieee9_RMSE_corrupt}
\end{figure*}

In each case, the hybrid GP approach shows from one to two orders of magnitude lower RMSE than the full GP approach. On a closer look, one can observe that the difference in RMSE magnitude is larger when the number of thrown-out samples is larger, in other words, the larger probability mass is being crossed out. Specifically, the vicinity of expectation, the interval with the highest density ($55 \%$), has been excluded in the experiment with bus 4. This led to the difference of $2$ magnitude order in RMSE of ordinary and hybrid GPR. Conversely, for experiments with loads 6 and 8, the difference in magnitude order is $1$, since the distribution mass is lower -- $38~\%$ and $33.5~\%$, respectively. The results are summarized in Table~\ref{table:ieee9_RMSE_corrupt}. 

\begin{table}[ht]
\centering
    \begin{tabular}{l|l|l|l}
        \hline
        Experiment & RMSE GPR & RMSE Hybrid GPR & $\%$ dropped \\ \hline
        Load 4 & $3.65 \cdot 10^{-1}$ & $\boldsymbol {9.11 \cdot 10^{-3}}$ & $55.0\%$  \\ \hline
        Load 6 & $3.89 \cdot 10^{-1}$ & $\boldsymbol {1.16 \cdot 10^{-2}}$ & $38.0\%$ \\ \hline
        Load 8 & $9.25 \cdot 10^{-2}$ & $\boldsymbol {8.34 \cdot 10^{-3}}$ & $33.5\%$ \\ \hline
        \end{tabular}
    \caption[RMSE comparison in the case of missing data]{Summary of RMSE comparison in the case of missing data. The last column specifies the portion of data distribution of the corresponding load that has been dropped.}
    \label{table:ieee9_RMSE_corrupt}
\end{table}

In summary, we studied the robustness of the proposed novel hybrid GPR method for CC AC-OPF against previously proposed full GPR for CC AC-OPF on an example of a missing part of the data domain. One can conclude that the proposed hybrid is more robust since it provides smaller RMSE even in the regions of the domain that have not been covered in the training data.

Another approach that may improve the results stands for introducing ranking regularization when learning the Gaussian processes~\cite{sidana2017representation,sidana2021user}. The latter allows to prioritize the error minimization across certain most critical lines and/or most uncertain outputs.

\section{Discussion}
To evaluate the optimization performance and uncertainty margins of the proposed hybrid approach, different stochastic approaches are considered. These approaches include the well-known sample-based approaches and the analytical state-of-the-art GP-based approach proposed in \textbf{Chapter~\ref{ch:4}}. This Chapter represents an improvement of the full GP approach that is faster and more robust.

The advantage of the GP approach is that it is based on Gaussian processes and thus enables to avoid of local sub-minima during optimization and realization of chance constraints in an easy and cheap way. The hybrid approach is faster since initial optimizing points are better defined with the help of a nominal linear model. Robustness is reflected in the fact that in the case where the historical data is poorly distributed in the space due to the system’s technical nature, the proposed model will be equal to the linearly approximated DC model. Thus, the possible prediction error will be significantly reduced. 

Although the proposed approach reduces the computational time, still it will be increased with system scale. This problem can be easily overcome by applying sparse GPR and dimensionality reduction techniques while maintaining a similar solution. Also, this framework can be easily extended to multi-period extensions of CC-OPF, where forecasts and errors may be correlated.
\section{Conclusion} \label{ch5_conclusion}
This Chapter discusses the use of Gaussian process regression (GPR) to learn non-linearities in stochastic OPF for improved performance. By combining the additive GPR with a linear power system model, we can improve solution accuracy and reduce computational complexity. This approach also helps to avoid spurious local minimums, making the AC-OPF problem more tractable. However, since GPR is not very scalable, a sparsity trick is employed to significantly reduce complexity. Accordingly, we propose a hybrid GPR approach that shows higher robustness to missing and corrupted data than the classical full GPR. Through simulation examples, we demonstrate how the proposed formulations provide secure control with improved performance for numerous IEEE test cases, with low optimization time. In particular, our approach provides a trade-off between the system's security, optimal cost, and time complexity.
 \chapter{GP CC-OPF Software}\label{ch:6}
\chaptermark{Software}

\section{Introduction}
In this chapter, a brief overview is presented on the software that has been created from scratch to facilitate the realization of the contributions made in this thesis. The code for this software is accessible through GitHub~\href{https://github.com/mile888/gp_cc-opf} {https://github.com/mile888/full}, \href{https://github.com/mile888/hybrid_gp}{https://github.com/mile888/hybrid} and Codeocean platforms~\href{https://codeocean.com/capsule/9532852/tree/v1}{https://codeocean.com/}. In addition to this, a paper has been published that describes this code in \textit{Software Impact} journal~\cite{mitrovic2023gp}.

\section{User guide}
The software implements the GP CC-OPF  approach, which is depicted in Fig.~\ref{fig:software_1}. This approach has been developed within a Python environment and relies on various libraries such as NumPy, SciPy, Pandas, and CasADi. NumPy is utilized for efficient matrix and vector operations, which are used for constructing the proposed algorithm. SciPy is employed for optimizing the GP model, while Pandas played a crucial role in generating and preparing synthetic datasets. CasADi, on the other hand, serves as a solver for non-linear optimization problems and is used in our case to model and realize the CC-OPF approach. Code details are presented below.

\begin{figure}[!t]
\centering
\vspace*{-10pt}
\includegraphics[width=0.95\textwidth]{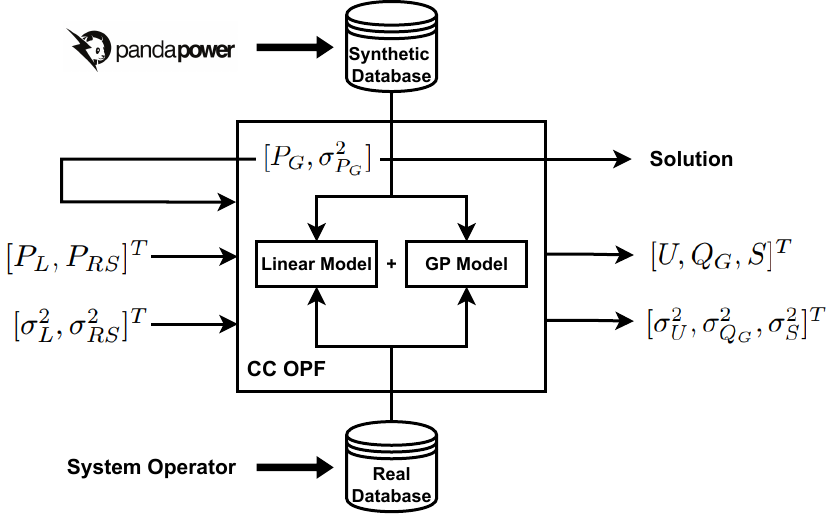}
\caption[General scheme of GP CC-OPF approach]{General scheme of GP CC-OPF approach realized in the software.}
\label{fig:software_1}
\end{figure}

\subsection{Data preparation}
We investigate two software approaches: the full GP CC-OPF and the hybrid GP CC-OPF. The key distinction between these approaches lies in their training methodologies.

The full approach is trained on the complete AC-PF  balance equations, while the hybrid approach consists of a linear DC  part and a residual part. The linear part can be implemented in two ways: using physical linear DC-PF equations or adopting a data-driven way. In this study, we employ the data-driven way, where data from the DC-PF model is generated to learn the linear function. The linear model, illustrated in the Fig.~\ref{fig:software_1}, utilizes ordinary least squares Linear Regression provided by the scikit-learn library~\cite{scikit-learn}. Additionally, the data for the residual part is generated by calculating the residuals between the AC-PF and DC-PF sample results on which the GP is trained.

Since all simulations are performed on synthetic datasets, we have developed a program for generating these datasets. We utilize the pandapower framework~\cite{thurner2018pandapower} to simulate power systems and generate the corresponding synthetic datasets. While the ultimate goal is for this software to be applicable to real-world data, we currently evaluate its performance using synthetic data. In both cases and approaches, the GP model is trained on standardized data.

The folder named \textit{<IEEE\{x\} - pandapower>} for full and \textit{<IEEE\{x\}\_hybrid - pandapower>} for hybrid approach provide the implementation details on how to simulate static power system flows and generate synthetic datasets using the pandapower framework.

\subsection{GP CC-OPF setup}
The GP CC-OPF approach combines conventional chance-constrained optimization with an integrated data-driven machine learning model, as illustrated in Fig.~\ref{fig:software_1}. The software implementation comprises several Python modules that interact with each other:
\begin{itemize}
    \item \textbf{gp\_functions.py}: This module creates the machine learning model using the SEard (Squared Exponential with Automatic Relevance Determination) kernel and various approximations such as TA1, TA2, and EM.
    \item \textbf{gp\_optimization.py}: The gp\_optimization.py module defines the optimization structure for the GP model.
    \item \textbf{gp\_regression.py}: This module integrates both gp\_functions.py and gp\_optimization.py to train the GP model.
    \item \textbf{gp\_ccopf.py} (for the full approach): The gp\_ccopf.py module incorporates the trained GP model for the full GP CC-OPF approach.
    \item \textbf{gp\_ccopf\_hybrid.py} (for the full approach): The gp\_ccopf\_hybrid.py module incorporates the trained GP model for the hybrid GP CC-OPF approach.
    \item \textbf{linear\_model.py} (for the hybrid approach): In this module, a Linear Regression model is created to generate the linear DC part within the hybrid approach.
\end{itemize}

These modules work together to implement the full GP CC-OPF approach and its hybrid variant. These modules work together to implement the full GP CC-OPF approach and its hybrid variant. A schematic representation of the module interaction is shown in Fig.~\ref{fig:software_2}.

\begin{figure}[h]
\centering
\vspace*{-10pt}
\includegraphics[width=0.95\textwidth]{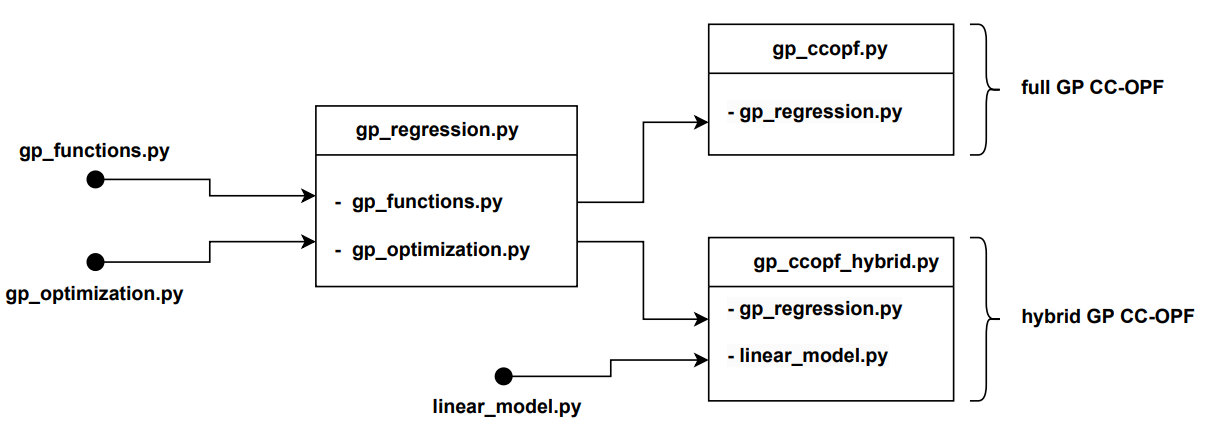}
\caption[General scheme of GP CC-OPF approach]{General scheme of GP CC-OPF approach realized in the software.}
\label{fig:software_2}
\end{figure}

The gp\_functions.py module creates the GP machine learning model while gp\_optimization.py provides the optimization structure of the GP model. In both approaches gp\_regression.py combines them for training the GP model. The resulting model is then integrated into gp\_ccopf.py for the full approach as well as gp\_ccopf\_hybrid.py for the hybrid approach. Additionally, the linear\_model.py module is included in gp\_ccopf\_hybrid.py module and handles the creation of the Linear Regression model for the linear DC part.

\subsection{Outputs of the tool}
The GP CC-OPF approach, implemented in the developed tool, enables users to obtain optimal solutions for controllable generators while minimizing the cost. Additionally, the tool provides valuable information such as output variables and their associated propagated standard deviations. These variables include voltage magnitudes at non-controllable buses, reactive power at controllable buses, and apparent power flow in the grid lines. By examining these outputs, users can assess the feasibility and accuracy of the obtained solution.

In summary, the developed tool utilizing the GP CC-OPF approach not only delivers optimal solutions for controllable generators but also provides crucial information regarding the system's variables and their propagated standard deviations. This allows users to monitor the feasibility and accuracy of the obtained solution effectively.

\section{Conclusion}
Currently, the industry lacks developed software that effectively solves stochastic OPF problems. System operators predominantly rely on deterministic OPF solutions that do not account for generation uncertainties. In the scientific community, although considerable progress has been made in stochastic OPF, the algorithms proposed in the literature are often inaccessible to the public. This presents challenges for researchers in terms of comparing their methods with existing work and necessitates significant effort to re-implement existing algorithms. Consequently, the availability of open-source code can significantly impact both industry and academia.

In future versions, our plan is to enhance the tool by introducing a user-friendly interface that is intuitive to use. We firmly believe that the proposed data-driven approach, presented in the form of public code, will have a positive impact on future research in stochastic OPF. Furthermore, we anticipate that it will serve as a stimulus for engineers to adopt these methods in the industry.

 \chapter{Conclusion and Future Perspectives}\label{ch:conclusion}
\chaptermark{Conclusion}

\section{Concluding remarks}
This thesis addresses the challenge of solving stochastic OPF using a data-driven chance-constrained approach. To achieve this, we use a Gaussian process regression (GPR) model that enables us to propagate uncertainties and satisfy the analytical reformulation of CC-OPF. The first significant contribution of the thesis is to replace the full AC-PF balance equations with GPR and incorporate them into the CC-OPF problem. This approach leads to significant improvements in both computational complexity and the accuracy of the solution. Furthermore, this approach facilitates the calculation of uncertainty margins in stochastic AC-OPF, achieving less than a 2.5\% deviation in chance constraint optimization. A key advantage of this approach is the use of the Squared Exponential covariance function within the Gaussian Process model. This allows us to approximate smoother functions, consequently avoiding local and sub-local minima, leading to faster and more robust solutions.

However, this approach can be affected by the quality of the dataset, which can impact the solution's robustness. To address this issue, we propose a more robust solution that is at worst-case equal to the linearly approximated DC model. We achieve this by proposing a hybrid approach that combines a linear DC model with a GP-based model learned on the residuals between AC-PF and DC-PF. This approach overcomes the drawbacks of the full GP approach and provides a more robust and efficient solution. To further improve the computational complexity and find a trade-off between complexity and accuracy, we use a sparse GP technique. This technique enables us to significantly reduce computational complexity while maintaining a high level of accuracy. Through the implementation of this hybrid and sparse approach, we can enhance efficiency, achieving a minimum 2x speedup and up to 6x more accurate solutions when compared to state-of-the-art approaches.

Overall, our approach offers a promising solution for addressing stochastic OPF problems by combining data-driven strategies with chance-constrained methodologies and leveraging machine learning techniques. Additionally, it is worth noting that neither of the proposed GP approaches necessitates system parameters or topology information. Consequently, the implementation of these methods can be seamlessly executed in real-time, with the overall efficiency contingent upon the quality of the dataset.

In summary, this thesis has yielded significant research outcomes such as:
\begin{enumerate}
\item developed a data-driven approach that analytically calculates uncertainty margins in stochastic AC-OPF with less than 2.5\% deviation in chance constraints optimization;
\item local and sub-local minima could be avoided by using the Squared Exponential covariance function of the Gaussian Process model;
\item implementing a hybrid and sparse approach can improve efficiency with a minimum speedup of 2 times and up to 6 times more accurate solutions compared to state-of-the-art approaches;
\item information about system parameters and topologies is not required.
\end{enumerate}

It is worth noting that the first contribution to this dissertation, assuming a full GP CC-OF approach, is published in the \textit{International Journal of Electrical Power \& Energy Systems (IJEPES)}~\cite{mile1}. The proposed hybrid GP CC-OF approach is presented and published at the \textit{IEEE Belgrade PowerTech 2023} conference~\cite{mile2}. The software developed to implement the approaches presented in this thesis is explained and published in the \textit{Software Impact} journal~\cite{mitrovic2023gp}.

\section{Future perspectives}
The current work is based on synthetic data. Although many ML solutions and applications have been proposed, they have not yet been widely implemented in real-world settings and are often only tested on open-source data. To evaluate the effectiveness of proposed software solutions, it would be valuable to test them on real power system data generated by Transmission System Operators (TSOs).

If the proposed software solution works well on real-world data, it has a high probability of becoming a primary data-driven solution for solving stochastic Optimal Power Flow (OPF) problems globally.

 \addcontentsline{toc}{section}{\bibname}

\begin{singlespace}
\nocite*
\bibliographystyle{apalike}
\bibliography{main}

\end{singlespace}

\end{document}